\documentclass[12pt,a4paper]{report}
\usepackage{geometry}
\usepackage{setspace}
\usepackage{titlesec}
\usepackage{tocloft}
\usepackage{amsmath,amssymb,amsfonts,amsthm}
\usepackage{algorithm}
\usepackage{makecell}
\usepackage{booktabs}
\usepackage{graphicx}
\usepackage{multirow}
\usepackage{footmisc}
\usepackage{subcaption}
\usepackage{caption}
\usepackage{adjustbox}
\usepackage{tabularx} 
\usepackage{enumitem}
\usepackage{pifont}
\usepackage{tcolorbox}
\usepackage{tikz}
\usetikzlibrary{backgrounds}
\usetikzlibrary{calc}
\usetikzlibrary{positioning}
\usetikzlibrary{shapes.geometric, arrows.meta}
\usetikzlibrary{decorations.pathreplacing}
\usetikzlibrary{shapes.misc}
\usetikzlibrary{patterns.meta}
\usepackage[noend]{algpseudocode}
\newtheorem{theorem}{Theorem}[chapter]
\newtheorem{assumption}{Assumption}[chapter]

\usepackage[numbers]{natbib}
\theoremstyle{definition}
\newtheorem{definition}{Definition}[chapter]
\newtheorem{corollary}{Corollary}[chapter]
\newtheorem*{hypothesis}{Hypothesis}
\newtheorem*{condition}{Condition}
\newtheorem{remark}{Remark}[chapter]
\titleformat{\chapter}[hang]{\Large\bfseries}{Chapter\ \thechapter.}{1em}{}
\titleformat{\section}[hang]{\large\bfseries}{\thesection}{1em}{}

\begin{document}
\newpage
\thispagestyle{empty}

\begin{center}
% --- top line (small) ---
{\bf\large A Dissertation for the Degree of Doctor of Philosophy in Mathematics}

\vspace{2.2cm}

% --- title (large, auto-wrap within a centered box) ---
{\LARGE\bfseries
\parbox{1.\textwidth}{\centering
On the Problem of Consistent Anomalies\\[0.3em]
in Zero-Shot Anomaly Detection}
}

\vspace{2.2cm}

% --- affiliation block ---
{\bf\large
Department of Mathematics\\
Graduate School\\
Chungnam National University
}

\vspace{1.8cm}

% --- author ---
{\bf\large By}\\[0.4cm]
{\Large\bfseries Tai Le-Gia}

\vspace{1.8cm}

% --- advisor ---
{\bf\Large Advisor: Prof. JaeHyun Ahn}

\vfill

% --- date at bottom ---
{\bf\large February, 2026}
\end{center}
%%%%%%%%%%%%%%%%%%%%%%%%%%%%%%%%%%%%%%%%%%%%%%%%%%%%%%%%%%%%%%%%%%%%%%
%%  Inside Cover page  %%  modification : title of paper			%%
%%					   %%				  advisor professor name	%%
%%					   %%				  your major				%%
%%					   %%				  send date of judge paper  %%
%%					   %%				  your name					%%
%%%%%%%%%%%%%%%%%%%%%%%%%%%%%%%%%%%%%%%%%%%%%%%%%%%%%%%%%%%%%%%%%%%%%%
\newpage
\thispagestyle{empty}

% --- Title (auto-wrap across multiple lines) ---
\begin{center}
    
{\LARGE\bfseries
\parbox{1.\textwidth}{\centering
On the Problem of Consistent Anomalies\\[0.3em]
in Zero-Shot Anomaly Detection}
}
\vspace*{2.5cm}

% Advisor
\centerline {\bf\Large Advisor : Prof. JaeHyun Ahn} \vspace*{2.5cm}

% Submission lines
\centerline {\bf\large Submitted to the Graduate School} \medskip
\centerline {\bf\large in Partial Fulfillment of the Requirements} \medskip
\centerline {\bf\large for the Degree of} \medskip
\centerline {\bf\large Doctor of Philosophy in Mathematics} \vspace*{2cm}

% Date
\centerline {\bf\Large October, 2025} \vspace*{2cm}

% Affiliation
\centerline {\bf Department of Mathematics}\medskip
\centerline {\bf Graduate School}\medskip
\centerline {\bf\large Chungnam National University, Korea} \vspace*{1.5cm}

% Author
\centerline {\bf By} \bigskip
\centerline {\bf\Large Tai Le-Gia}
\end{center}
\newpage
\thispagestyle{empty}
\begin{center}
{\bf To Approve the Submitted Dissertation}\\[0.5em]
{\bf for the Degree of Doctor of Philosophy in Mathematics}
\end{center}

\vspace*{1cm}

\begin{center}
{\bf By \bf Tai Le-Gia}
\end{center}

\vspace*{1cm}

\begin{center}
{\Large\bfseries
\parbox{0.9\textwidth}{\centering
On the Problem of Consistent Anomalies\\[0.3em]
in Zero-Shot Anomaly Detection}
}
\end{center}

\vspace*{1.5cm}

\begin{center}
{\bf December 2025}
\end{center}

\vspace*{1.0cm}
\begin{center}
{\noindent\bf Committee Chair}{\quad \_\_\_\_\_\_\_\_\_\_\_\_\_\_\_\_\_\_\_\_\_\_\_\_\_\_\_\_\_\_\_\_\_\_\_\_\_\_\_\_\_\_\_\_\_\_\_}\vspace*{1cm}\\
{\noindent\bf Committee\qquad}\quad \quad \_\_\_\_\_\_\_\_\_\_\_\_\_\_\_\_\_\_\_\_\_\_\_\_\_\_\_\_\_\_\_\_\_\_\_\_\_\_\_\_\_\_\_\_\_\_\_\vspace*{1cm}\\
{\noindent\bf Committee\qquad}\quad \quad \_\_\_\_\_\_\_\_\_\_\_\_\_\_\_\_\_\_\_\_\_\_\_\_\_\_\_\_\_\_\_\_\_\_\_\_\_\_\_\_\_\_\_\_\_\_\_\vspace*{1cm}\\
{\noindent\bf Committee\qquad}\quad \quad \_\_\_\_\_\_\_\_\_\_\_\_\_\_\_\_\_\_\_\_\_\_\_\_\_\_\_\_\_\_\_\_\_\_\_\_\_\_\_\_\_\_\_\_\_\_\_\vspace*{1cm}\\
{\noindent\bf Committee\qquad}\quad \quad \_\_\_\_\_\_\_\_\_\_\_\_\_\_\_\_\_\_\_\_\_\_\_\_\_\_\_\_\_\_\_\_\_\_\_\_\_\_\_\_\_\_\_\_\_\_\_\vspace*{1cm}\\
\end{center}

%\vspace*{1cm}

\begin{center}
{\large\bf Graduate School}\\[0.5cm]
{\large\bf Chungnam National University, Korea}
\end{center}
\newpage
\setcounter{page}{0}
\thispagestyle{empty}
\tableofcontents

%%%%%%%%%%%%%%%%%%%%%%%%%%%%%%%%%%%%%%%%%%%%%%%%%%%%%%%%%
%%  Body pages  %%  Start to section of Introduction.  %%
%%				%%	Now, make out your paper.		   %%
%%%%%%%%%%%%%%%%%%%%%%%%%%%%%%%%%%%%%%%%%%%%%%%%%%%%%%%%%

% ------------------------------------------------------
\chapter*{Preface}
\addcontentsline{toc}{chapter}{Introduction}

This preface serves as a guide for this dissertation. It summarizes the content of each chapter and outlines the main contributions. Almost all the content presented herein has been published or is currently under submission, as of the time of writing, in collaboration with my supervisors.

Anomaly detection, also known as outlier or novelty detection, is a classical yet active research field in statistics and machine learning, with origins dating back to the 1960s~\cite{grubbs1969procedures}. It aims to identify rare events or samples that deviate significantly from the majority of the data. As defined by~\citet{hawkins1980identification}:
\vspace{2mm}
\begin{quote}
``An outlier is an observation which deviates so much from the other observations as to arouse suspicions that it was generated by a different mechanism.''
\end{quote}
\vspace{2mm}

Anomaly detection has found applications across a wide range of domains, including computer vision~\cite{Liu_2024_IndustrialAnomalySurvey}, medical imaging~\cite{fernando2021deep}, and finance~\cite{zamanzadeh2024deep}. This thesis focuses on anomaly detection in industrial and medical imaging, where anomalies appear as defects, tumors, or structural irregularities that must be localized with high precision.

Early research in anomaly detection relied heavily on statistical modeling and distance-based algorithms. Classical approaches such as Gaussian Mixture Models (GMMs)~\citep{yang2009outlier} and $k$-Nearest Neighbors (kNN)~\cite{10.5555/3086742} modeled normality under assumptions about data distributions. These techniques performed well for low-dimensional or structured datasets but struggled with high-dimensional inputs such as images, where the curse of dimensionality and the lack of discriminative features limited their effectiveness.

The deep learning era marked a fundamental shift in the field~\cite{pang2021deep}. Neural networks enabled the extraction of low-dimensional yet semantically rich features from complex data. Autoencoders, variational autoencoders, and generative adversarial networks became popular for reconstructing normal data and detecting deviations through reconstruction errors~\cite{schlegl2019f,PINAYA2022102475,ddim}. These methods achieved remarkable success in visual and medical domains but typically required extensive training data and suffered from strong domain dependency.

In industrial inspection, the demand for high accuracy with minimal supervision led to the development of standardized datasets such as MVTec AD~\cite{mvtec} and Visa~\cite{visa}. These benchmarks inspired a new wave of training-free approaches that operate directly in feature space. PatchCore~\cite{patchcore} and related methods~\citep{padim, anomalydino} demonstrated that pretrained models could serve as powerful feature extractors, enabling anomaly detection by measuring patch-level similarity between test and reference samples. Such methods successfully bypassed the need for retraining while maintaining competitive performance.

Recent advances in foundation models have reshaped anomaly detection by enabling both few-shot and zero-shot generalization. Large-scale pretrained representations such as CLIP~\cite{CLIP} and DINOv2~\cite{DINOv2} provide highly transferable visual features learned from vast and diverse datasets. These advances have supported the emergence of two major settings in anomaly detection. In few-shot detection, models uses only a small number of labeled normal samples. In zero-shot detection, no labels are required. Batch-based zero-shot methods leverage the abundance of normal patterns within large unlabeled batches to identify anomalies. In contrast, text-based zero-shot methods built upon vision–language models (VLMs)~\cite{WinCLIP,fort2021exploring} align visual patterns with semantic concepts, enabling interpretable and prompt-driven anomaly reasoning. Together, these developments have greatly reduced the need for costly supervision and expanded the applicability of anomaly detection across diverse domains.

Despite these outstanding advancements, zero-shot anomaly detection remains an ill-posed and under-explored problem, with many realistic challenges that have not been systematically addressed. In this thesis, we introduce the problem of \emph{consistent anomalies} in zero-shot anomaly detection, where similar defects recur across images. We also investigate the extension of zero-shot anomaly detection to \emph{3D medical imaging} by utilizing pre-trained 2D models. Finally, we present exploratory results that encourage the combination of batch-based and text-based paradigms, taking a step toward unified zero-shot anomaly detection frameworks.

In Chapter~\ref{chap:background}, we review the foundational concepts and related works, categorizing existing zero-shot methods into text-based and batch-based paradigms while highlighting their underlying assumptions and limitations.

In Chapter~\ref{chap:empirical}, we introduce the problem of consistent anomalies and present two empirical phenomena: the \emph{Similarity Scaling Phenomenon}, which describes the power-law decay in similarity growth rates for normal elements, and the \emph{Neighbor-Burnout Phenomenon}, which characterizes deviations caused by consistent anomalies.

In Chapter~\ref{chap:algorithm}, we build upon these phenomena from Chapter~\ref{chap:empirical} and construct the \emph{CoDeGraph}-a graph-based zero-shot method designed to address the consistent-anomaly problem.

In Chapter~\ref{chap:theory-scaling}, we establish the theoretical foundation for the Similarity Scaling Phenomenon using Extreme Value Theory (EVT) and geometric analysis on locally regular manifolds, demonstrating how heavy-tailed similarity distributions arise under relatively loose conditions.

In Chapter~\ref{chap:3d-extension}, we extend zero-shot batch-based methods to 3D MRI volumes by developing a training-free volumetric pipeline based on multi-axis slicing, pooling, and fusion of 2D patch tokens, ensuring spatial coherence in anomaly localization across slices.

Finally, in Chapter~\ref{chap:bridge-batch-text}, we take the first step toward bridging batch-based and text-based zero-shot methods, laying the groundwork for unified foundation-model-driven anomaly detection.

% ------------------------------------------------------
\chapter{Background and Related Work}
\label{chap:background}
This chapter establishes the foundation of the thesis by defining the anomaly detection problem and reviewing key methodologies. Its purpose is to clarify terminology, outline major learning paradigms, and summarize recent advances that motivate the subsequent chapters.

Section~\ref{chap:background}.1 introduces the formal definition and taxonomy of anomaly detection.
Section~\ref{chap:background}.2 reviews the main learning paradigms, including full-shot, few-shot, and zero-shot settings.
Section~\ref{chap:background}.3 focuses on zero-shot anomaly detection, presenting \emph{text-based} and \emph{batch-based} approaches and highlighting their complementary strengths and limitations.
\section{Anomaly Detection: Taxonomy and Problem Definition}

Anomaly Detection (AD) is the task of identifying observations that are unlikely under the probability distribution of normal data.  
Let $(\mathcal{X}, \mathcal{F})$ be a measurable space, and let $p_N$ denote the (unknown) probability distribution that generates \emph{normal} samples on $\mathcal{X}$.  
Given an observation $x \in \mathcal{X}$, the ultimate goal of AD is to learn an \emph{indicator function}
\[
    G(x): \mathcal{X} \rightarrow \{0, 1\},
\]
where $G(x) = 0$ indicates that $x$ was generated by $p_N$ (normal), and $G(x) = 1$ indicates that $x$ is anomalous, i.e., not drawn from $p_N$.

In essence, the anomaly detection problem is an \emph{ill-posed} problem due to the fact that  
normal samples are not fully observable, while anomalous samples are typically rare, heterogeneous, or entirely absent during training.  
Hence, AD is often formulated as a \emph{one-class classification} problem, where the challenge is to infer the boundary of normality solely from available normal samples.  
In practice, the objective is to construct an \emph{anomaly score function} \( a : \mathcal{X} \rightarrow \mathbb{R}, \) with the philosophy that it assigns low scores to likely normal samples and high scores to potential anomalies. The final decision is given by a threshold $\tau$, which can be selected using normal samples or a validation dataset if it is available:
\begin{equation}
    G(x) =
    \begin{cases}
        0, & \text{if } a(x) < \tau, \\
        1, & \text{if } a(x) \ge \tau.
    \end{cases}
    \label{for:anomaly_threshold}
\end{equation}

\subsubsection*{Anomaly Classification and Segmentation}

Researches in vision anomaly detection can broadly be divided into two branches.  
The first focuses on distinguishing \emph{semantically different} samples—for instance, when the normal class consists of images of dogs, while cats, cars, or any other objects are considered anomalous.  
This setting is often referred to as \emph{out-of-distribution (OOD) detection}, as the anomalies belong to distributions that are semantically distinct from the normal data.

The second branch, which is the main focus of this thesis, deals with cases where normal and anomalous samples are \emph{semantically identical}.  
Here, anomalies arise not from class differences but from local and logic irregularities within the same semantic category—for example, defects in industrial images or tumors in medical scans.  
The aim is not only to identify which samples are anomalous but also to \emph{localize} the anomalous regions.

Throughout this book, we adopt a unified framework based on the concepts of \emph{collections} and \emph{elements}.  
Each collection $X = \{x_1, \dots, x_n\}$ corresponds to an individual sample (e.g., an image or a 3D MRI volume), and each element $x_i$ represents a local spatial or structural unit (e.g., a patch or a 3D-patch).  
Anomaly classification (AC) determines whether a collection $X$ contains any anomalous elements, while anomaly segmentation (AS) aims to identify the anomalous elements within $X$.

Recent advances in computer vision have been driven by \emph{foundation models} such as CLIP \cite{CLIP}, Dinov2 \cite{DINOv2} and GPT4 \cite{gpt4}, which are trained on massive datasets. These models yield rich, transferable representations that can be adapted (e.g., fine-tuned) and generalize remarkably well across downstream tasks, often outperforming models trained from scratch~\cite{gan2024erasing}. Anomaly detection has also greatly benefited from this paradigm: current state-of-the-art (SOTA) methods rely on pre-trained encoders to obtain robust and semantically rich feature embeddings, which serve as the foundation for detecting anomalies. Most contemporary approaches to AC and AS, especially in \emph{few-shot} and \emph{zero-shot} settings, follow a general two-stage architecture:

\begin{enumerate}
    \item \textbf{Feature extraction:} A pre-trained encoder $f$ processes an entire collection $X$ and produces a set of element-level features,
    \[
        f(X) = (\mathbf{z}_1, \dots, \mathbf{z}_p), \qquad \mathbf{z}_i \in \mathbb{R}^d,
    \]
    where $p$ is the number of elements (e.g., patches) and $d$ is the feature dimension.  
    Each feature $\mathbf{z}_i$ encodes the local structural and semantic information of its corresponding region within the collection.

    \item \textbf{Anomaly scoring:} An anomaly score function
    \[
        a : \mathbb{R}^d \rightarrow \mathbb{R},
    \]
    assigns an anomaly value to each element-level feature $\mathbf{z}_i$, such that $a(\mathbf{z}_i)$ is low for normal regions and high for anomalous ones.  
Following the literature \citep{patchcore,MuSc,cflow}, distance-based or distribution-based methods typically compute the collection-level anomaly score $A(X)$ as the maximum (or top-k average) of the element-level scores:
$$A(X) = \max_i a(\mathbf{z}_i).$$
In contrast, text-based approaches often derive the anomaly score directly from a global representation, such as the special \texttt{[CLS]} token, which aggregates semantic information across all elements in $X$.
\end{enumerate}
%Throughout this thesis, we may interchangeably refer to \emph{element-level} features as \emph{patch-level} (for 2D images) or \emph{voxel-level} (for 3D volumes), and to \emph{collection-level} scores as \emph{image-level} or \emph{volume-level} scores, depending on the modality.

\section{Learning Paradigms: Full-shot and Few-shot}
Anomaly detection methods are broadly categorized into full-shot, few-shot, and zero-shot paradigms according to the amount of normal data available for training.

\subsection*{Full-shot learning}
In the full-shot setting, a large corpus of normal training images from the target domain is assumed to be available. Two main branches are commonly recognized:

\begin{itemize}
    \item \textbf{Feature-based methods:} These methods extract element-level features from normal images to characterize the normal feature distribution and subsequently identify abnormal regions.
    
        \begin{enumerate}
        \item Memory-bank methods \citep{patchcore, padim} construct a database of representative normal features and compute anomaly scores based on nearest-neighbor distances or local density estimation.
        
        \item Teacher--student methods \citep{efficientAD, ast} train a student network to mimic the representations learned by a fixed teacher network solely on normal samples. The disagreements between the teacher and student responses are used as anomaly scores.
        
        \item Distribution learning methods~\citep{msflow, cflow} employ probabilistic density estimation, typically via normalizing flows or other generative models, to learn explicit likelihood maps of normal features.
    \end{enumerate}
    
    \item \textbf{Reconstruction-based methods:} These approaches are founded on the assumption that models trained exclusively on normal data can reconstruct normal regions accurately but fail to reconstruct abnormal ones. Typically, a neural network is trained on normal samples to reconstruct themselves using autoencoders \citep{efficientAD,cai2023dual} or diffusion models \citep{ddim}. Anomalies are then detected by comparing input data with their reconstructions using pixel-wise discrepancies.
\end{itemize}

\subsection*{Few-shot learning}
Few-shot methods assume access to only a small number of normal samples (typically four to eight) from the target domain. The goal is to generalize the notion of normality from such limited supervision. To compensate for the scarcity of data, few-shot approaches aim to enrich the available feature representations. One common strategy applies data augmentations that diversify the feature manifold and improve generalization under limited samples \citep{anomalydino,GraphCore}. Another strategy leverages text-based modalities, in which semantic prompts serve as auxiliary supervision to guide the model’s interpretation of normal features~\citep{incrtl,lv2025oneforall}. More recently, AnomalyGPT~\citep{anomalygpt} has combined these two directions by generating synthetic anomalies alongside textual descriptions, thereby enriching both the visual and semantic feature spaces and further enhancing few-shot generalization.

\section{Zero-shot Anomaly Detection}
Zero-shot anomaly detection seeks to identify abnormal regions without accessing any target-domain data during training. Current zero-shot anomaly detection methods fall into two main categories: text-based and batch-based approaches. Both paradigms adhere to the strict constraint that the model receives no training on the target dataset, yet they differ in the prior knowledge they exploit and the operational scenarios they target. Before exploring these paradigms in detail, we first introduce the Vision Transformer (ViT) and Vision-Language Models, which serve as the foundational backbones for both batch-based and text-based methods.

\subsection{Vision Transformer and Vision--Language Models}
\begin{figure}[t!]
 \centering
  \includegraphics[width=0.95\textwidth]{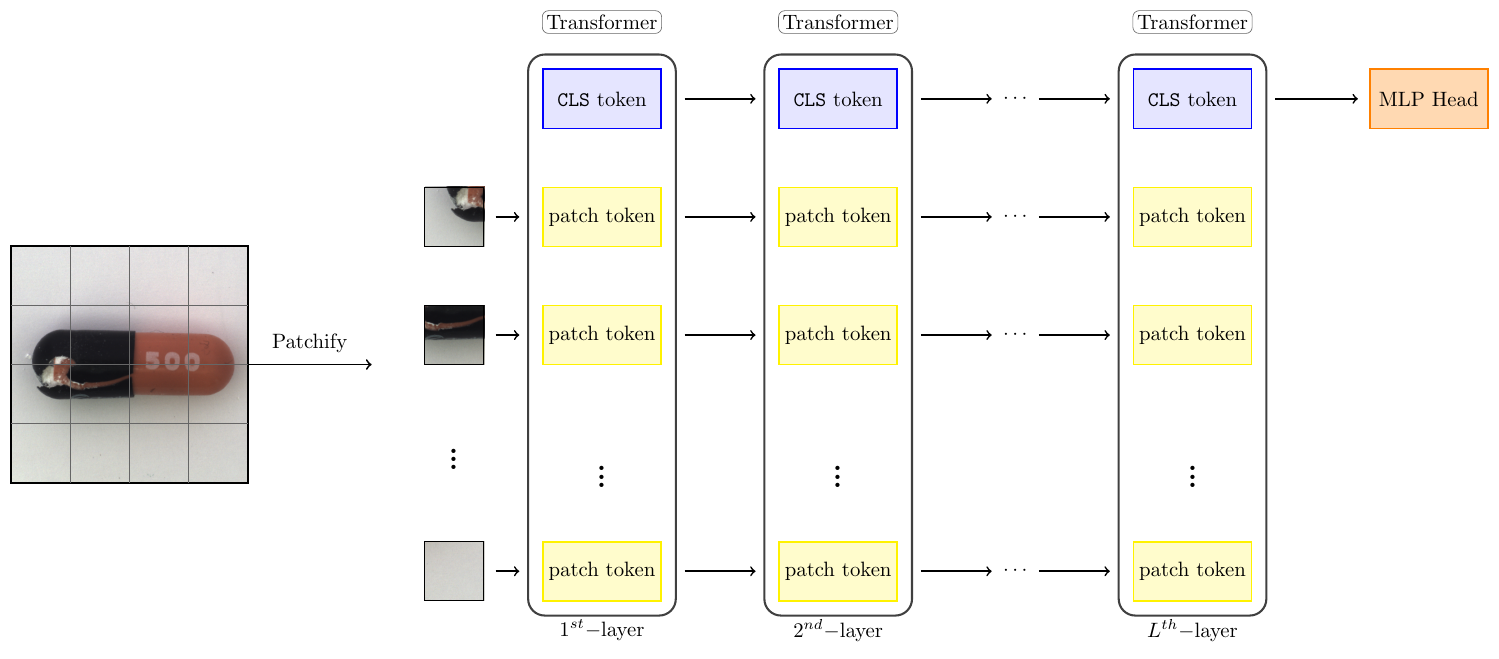}
  \caption{Vision Transformer architecture.}
  \label{fig:vit_arch}
\end{figure}
\subsubsection{Vision Transformer}
\label{sec:vit}
The Vision Transformer~\citep{ViT} adapts the Transformer architecture~\citep{vaswani2017attention}, originally designed for sequential text data, for image understanding tasks. Unlike conventional CNN architectures~\citep{krizhevsky2012imagenet} that rely on local receptive fields and hierarchical structure, ViT treats an image as a sequence of tokens. Given an image $\mathbf{x} \in \mathbb{R}^{H \times W \times C}$, the image is divided into non-overlapping patches of size $P \times P$, each flattened into a vector and combined as
\[
\mathbf{x}_p = [\mathbf{x}_p^1, \mathbf{x}_p^2, \dots, \mathbf{x}_p^N] \in \mathbb{R}^{N \times (P^2 \cdot C)},
\]
where $N = HW / P^2$ is the number of patches per image. Each patch is linearly projected into a $D$-dimensional feature space using a learnable projection matrix $\mathbf{E} \in \mathbb{R}^{(P^2 \cdot C) \times D}$ to produce the so-called \emph{patch embeddings}, $\mathbf{z}_p^i = \mathbf{x}_p^i \mathbf{E} \in \mathbb{R}^{D}$. To retain positional information, ViT adds 1D learnable positional embeddings $\mathbf{E}_{\text{pos}} \in \mathbb{R}^{(N+1) \times D}$. Furthermore, a special learnable \texttt{[CLS]} token $\mathbf{x}_{\text{cls}} \in \mathbb{R}^D$ is prepended to the sequence, serving as a global representative of the entire image. This \emph{patchification} process can be summarized as follows:
\begin{align}
    \mathbf{z}_0 &= 
    \big[ \mathbf{x}_{\text{cls}}; \mathbf{x}_p^1\mathbf{E}, \mathbf{x}_p^2\mathbf{E}, \ldots, \mathbf{x}_p^N\mathbf{E} \big] 
    + \mathbf{E}_{\text{pos}},
    && \mathbf{E} \in \mathbb{R}^{(P^2 \cdot C) \times D}, \;
    \mathbf{E}_{\text{pos}} \in \mathbb{R}^{(N+1) \times D}.
\end{align}

The sequence $\mathbf{z}_0$ is then processed sequentially through $L$ Transformer blocks using Eq.~\ref{msa} and Eq.~\ref{mlp}). Each block alternates between a multi-head self-attention (MSA) layer and a two-layer perceptron (MLP) with a GELU activation:
\begin{align}
    \mathbf{z}'_{\ell} &= \mathrm{MSA}(\mathrm{LN}(\mathbf{z}_{\ell-1})) + \mathbf{z}_{\ell-1}, &&\ell = 1, \dots, L. \label{msa}
\\
    \mathbf{z}_{\ell} &= \mathrm{MLP}(\mathrm{LN}(\mathbf{z}'_{\ell})) + \mathbf{z}'_{\ell}, && \ell = 1, \dots, L. \label{mlp}
\end{align}
where $\mathrm{LN}(\cdot)$ denotes layer normalization. Layer normalization \citep{ba2016layer} is applied before each block, and residual connections are added after each block to return $\mathbf{z}_{\ell} = \left[\mathbf{z}_{\ell}^{0}; \mathbf{z}_{\ell}^{1}, \ldots, \mathbf{z}_{\ell}^{N} \right]$. At the final layer, the output of the \texttt{[CLS]} token, $\mathbf{z}_L^{0}$, represents the entire image (collection-level feature), while the remaining patch tokens $\mathbf{z}_L^{i}$, $i = 1, \dots, N$, encode local structural and semantic details (element-level features).

During pre-training, ViT is optimized for image classification using the final \texttt{[CLS]} token. A linear classification head $h(\cdot)$ (typically an MLP or a single linear layer) is attached to the last \texttt{[CLS]} output to predict the probability distribution over $K$ classes:
\[
\hat{\mathbf{y}} = \mathrm{Softmax}\big(h(\mathbf{z}_L^{0})\big), \qquad
\mathcal{L}_{\text{cls}} = - \sum_{k=1}^{K} y_k \log \hat{y}_k,
\]
where $\mathbf{y}$ is the one-hot encoded ground-truth label.  
Because gradients flow primarily through $\mathbf{z}_L^{0}$, the \texttt{[CLS]} token learns to aggregate contextual information from all patch tokens via self-attention. As a result, it captures a compact yet semantically rich summary of the entire image, while each patch token retains localized feature information. The diagram from ViT architecture is given in Fig.~\ref{fig:vit_arch}.

Once pre-trained, the ViT backbone can be reused as a general-purpose feature extractor. The patch tokens $\{\mathbf{z}_l^{i}\}$ provide dense spatial descriptors suitable for fine-grained tasks such as segmentation or anomaly localization, whereas the \texttt{[CLS]} token serves as a global embedding for image-level tasks such as classification or retrieval.  

Throughout this thesis, these two token types play a central role: patch tokens serve as the fundamental units for element-level anomaly scoring, while the \texttt{[CLS]} token provides a compact global representation for collection-level decisions.
\subsubsection{Vision--Language Models}

Vision--Language Models (VLMs) extend the Transformer architecture to jointly learn representations from both visual and textual modalities. A representative model in this family is the \emph{Contrastive Language--Image Pre-training (CLIP)} model~\citep{CLIP}, which aligns images and texts in a shared multimodal embedding space through large-scale self-supervised learning.

A VLM consists of two encoders: a \emph{vision encoder} $f_v$ and a \emph{text encoder} $f_t$.  
The vision encoder $f_v$ is typically a Vision Transformer that processes an image $\mathbf{x}_v \in \mathbb{R}^{H \times W \times C}$ into a sequence of patch embeddings and a global \texttt{[CLS]} token, as described in the previous subsection:
\[
    f_v(\mathbf{x}_v) = \mathbf{z}_{L} = \big[\,\mathbf{z}^v_{\text{cls}},\, \mathbf{z}^v_1, \ldots, \mathbf{z}^v_N\,\big], \qquad
    \mathbf{z}^v_{\text{cls}}, \mathbf{z}^v_i \in \mathbb{R}^{D}.
\]
Here, $\mathbf{z}^v_{\text{cls}}$ and $\mathbf{z}^v_i$ denote the \texttt{[CLS]} token and patch tokens, respectively.

The text encoder $f_t$, implemented as a Transformer language model, encodes a text description (prompt) $\mathbf{x}_t = [w_1, \ldots, w_L]$, which is bracketed with \texttt{[SOS]} and \texttt{[EOS]} tokens\footnote{\texttt{[SOS]} and \texttt{[EOS]} stand for \emph{start of sequence} and \emph{end of sequence}, respectively. They mark the beginning and end of a text input in Transformer-based language models.}.  
The activation of the \texttt{[EOS]} token at the final layer is used to represent the sentence-level feature, serving a role analogous to the \texttt{[CLS]} token in ViT:
\[
    f_t(\mathbf{x}_t) = \big[\,\mathbf{z}^t_{1}, \ldots, \mathbf{z}^t_{L},\, \mathbf{z}^t_{\text{eos}}\,\big], \qquad
    \mathbf{z}^t_{\text{eos}} \in \mathbb{R}^{D}.
\]

To compare the two modalities, both $\mathbf{z}^v_{\text{cls}}$ and $\mathbf{z}^t_{\text{eos}}$ are projected into a shared multimodal embedding space of dimension $h$ through separate linear projection heads:
\[
    \mathbf{u}_v = W_v\, \mathbf{z}^v_{\text{cls}}, \qquad 
    \mathbf{u}_t = W_t\, \mathbf{z}^t_{\text{eos}}, 
    \qquad W_v, W_t \in \mathbb{R}^{h \times D}.
\]
These projected embeddings $\mathbf{u}_v, \mathbf{u}_t \in \mathbb{R}^{h}$ are normalized to unit length and used as the final multimodal representations for contrastive alignment.

CLIP is trained on large-scale image--text pairs $\{(\mathbf{x}_v^i, \mathbf{x}_t^i)\}_{i=1}^N$ collected from the web.  
The objective is to maximize agreement between matching image--text pairs while minimizing similarity between mismatched ones.  
This is achieved via a \emph{symmetric contrastive loss} that includes both image-to-text and text-to-image directions:
\begin{align}
\mathcal{L}_{\text{CLIP}} &= 
-\frac{1}{2} \sum_{i=1}^{N} 
\Bigg[
\log
\frac{\exp\big( \langle \mathbf{u}_v^i, \mathbf{u}_t^i \rangle / \tau \big)}
{\sum_{j=1}^{N} \exp\big( \langle \mathbf{u}_v^i, \mathbf{u}_t^j \rangle / \tau \big)}
+
\log
\frac{\exp\big( \langle \mathbf{u}_t^i, \mathbf{u}_v^i \rangle / \tau \big)}
{\sum_{j=1}^{N} \exp\big( \langle \mathbf{u}_t^i, \mathbf{u}_v^j \rangle / \tau \big)}
\Bigg],
\end{align}
where $\langle \cdot,\cdot \rangle$ denotes cosine similarity and $\tau$ is a learnable temperature parameter.  
Through this loss, the vision encoder learns to produce global \texttt{[CLS]} embeddings that semantically align with text embeddings, while maintaining discriminability across unrelated image--text pairs.

After pre-training, the learned encoders $f_v$ and $f_t$ enable zero-shot classification without additional fine-tuning.  
Given a set of names $\mathcal{C} = \{c_1, \dots, c_K\}$, textual prompts such as ``a photo of a [class]'' are encoded to obtain $\mathbf{u}_t^k = f_t(c_k)$, and the visual embedding $\mathbf{u}_v = f_v(\mathbf{x})$ of a test image is compared against all class embeddings:
\[
    \hat{y} = \arg\max_{k} \langle \mathbf{u}_v, \mathbf{u}_t^k \rangle.
\]
This open-vocabulary inference provides a flexible semantic interface for specifying high-level concepts, which can be readily adapted to downstream tasks such as zero-shot classification \citep{CLIP} or semantic image-retrieval \cite{zaigrajew2025interpreting}. In this thesis, this capability will later be extended to the task of \emph{zero-shot anomaly detection}, which adapts the same vision--language alignment principle to distinguish between normal and abnormal regions without requiring target-domain training data.
\subsection{Text-Based Methods for Zero-Shot Anomaly Detection}

Text-based approaches to zero-shot anomaly detection build directly upon VLMs such as CLIP, leveraging the semantic alignment between visual and textual embeddings.  
The core philosophy, first introduced by WinCLIP~\citep{WinCLIP}, is to formulate anomaly detection as a two-class classification problem within the multimodal embedding space.  
Specifically, a pair of text tokens are constructed: $\mathbf{u}_t^{n}$ representing the \emph{normal} state, and $\mathbf{u}_t^{a}$ representing the \emph{abnormal} state.  
Given an image $X$ with visual embeddings $\{\mathbf{u}_v\}$ and the corresponding text embeddings $\{\mathbf{u}_t^{n}, \mathbf{u}_t^{a}\}$, the anomaly score for each patch is computed via a softmax comparison:
\[
    s_a =
    \frac{\exp(\langle \mathbf{u}_v, \mathbf{u}_t^{a} \rangle / \tau)}
    {\exp(\langle \mathbf{u}_v, \mathbf{u}_t^{a} \rangle / \tau) + \exp(\langle \mathbf{u}_v, \mathbf{u}_t^{n} \rangle / \tau)},
\]
where $\tau$ is the temperature hyperparameter. This probabilistic formulation quantifies how strongly each patch aligns with the notion of “abnormality” relative to “normality," which CLIP implicitly learns through large-scale pretraining.

Subsequent works have extended this framework to improve localization accuracy and cross-domain generalization.  
APRIL-GAN~\citep{APRIL-GAN} enhances anomaly segmentation by learning a linear projection that maps intermediate patch tokens $\mathbf{z}_{\ell < L}$ (in Eq.~\eqref{msa}) into CLIP’s multimodal embedding space, enabling direct comparison between fine-grained visual representations at low layers and text embeddings.  
This projection is trained using auxiliary datasets with pixel-level anomaly annotations that are distinct from the target objects (e.g., train on MVTec and test on Visa).  
AnomalyCLIP~\citep{AnomalyCLIP} suggests that anomalies in different domains might share the same semantic meaning and proposes an \emph{object-agnostic} prompt design, replacing class-specific nouns with a generic placeholder such as [object].
This abstraction reduces dependence on specific categories and improves robustness in open-world settings, though it still relies on annotated datasets for fine-tuning text prompts.  
In general, most zero-shot methods in this category either employ sophisticated prompt-engineering strategies or incorporate learned projection modules to strengthen alignment between textual and visual features.

Despite their conceptual elegance, text-based methods remain constrained to \emph{visually interpretable anomalies}—such as scratches, dents, or surface defects—that exhibit clear semantic cues.  
They struggle with logic or contextual anomalies, where abnormality arises from relational inconsistencies or dataset-specific artifacts that lack explicit visual markers.  
These limitations motivate the development of alternative zero-shot approaches that do not rely on text, such as batch-based methods discussed next.
\subsection{Batch-Based Methods}
\label{sec:batch-based}
Unlike text-based approaches, batch-based zero-shot methods \citep{MuSc, ACR} do not rely on textual supervision.  
Let the test batch be denoted as \(\mathcal{D} = \{ C_1, C_2, \ldots, C_B \}\).  
For clarity, we distinguish two subsets: a \emph{query set} \(\mathcal{Q}\) and a \emph{base set} \(\mathcal{B}\); however, in zero-shot settings, these sets are typically identical, i.e., \(\mathcal{Q} = \mathcal{B} = \mathcal{D}\).  
The goal is to detect anomalous images (collections) and anomalous pixels (elements) in each query image by comparing its patch tokens against those of all other images in the base.

This approach is founded on the \emph{Doppelgänger}\footnote{\emph{Doppelgänger}: German literary term (literally ``double-goer'') denoting a near-identical double or look-alike of a person or thing.} assumption, which states that for most elements (patterns) within an image, there exists at least one visually similar counterpart in other images of the dataset, as presented in Fig~\ref{fig:voronoi}.  
In industrial or medical imaging datasets, normal regions typically exhibit such strong similarity, giving rise to numerous near-identical appearances across samples.  
In contrast, anomalies are rare and often appear as unique, unmatched patterns.  
This \emph{Rarity} property enables zero-shot AC/AS by identifying tokens that fail to find their ``doppelgängers'' in other images.

Under this assumption, MuSc~\citep{MuSc} introduces the \emph{Mutual Scoring Mechanism}.  
Given a query image \(C_i \in \mathcal{Q}\) and its sequence of patch tokens \([\mathbf{z}_i^1, \mathbf{z}_i^2, \ldots, \mathbf{z}_i^N]\), the distance from a patch token \(\mathbf{z}_i^h\) in \(C_i\) to a base image \(C_j \in \mathcal{B}\) \((C_j \neq C_i)\) is defined as the distance to its most similar token in \(C_j\):
\[
  d(\mathbf{z}_i^h, C_j) = \min_{k} \| \mathbf{z}_i^h - \mathbf{z}_j^k \|_2^2.
\]
The set of distances from \(\mathbf{z}_i^h\) to all other base images is then collected and sorted:
\[
  D_{\mathcal{B}}(\mathbf{z}_i^h) = [\, d(\mathbf{z}_i^h, C_{(1)}), \ldots, d(\mathbf{z}_i^h, C_{(B-1)}) \,],
\]
where \(d(\mathbf{z}_i^h, C_{(t)})\) denotes the \(t\)-th smallest distance.  
Tokens that consistently find close matches across many other images are regarded as normal, while those lacking close neighbors are likely anomalous.  
To obtain a stable measure, the $K$ smallest distances are averaged to produce the anomaly score:
\[
  a_{\mathcal{B}}(\mathbf{z}_i^h) = \frac{1}{K} \sum_{t=1}^K d(\mathbf{z}_i^h, C_{(t)}).
\]
This score quantifies how well a token fits into the shared visual space formed by the base images.  
Regions with large \(a_{\mathcal{B}}(\mathbf{z}_i^h)\) values correspond to tokens that fail to find enough counterparts and are thus flagged as anomalous.

The batch-based paradigm offers several key advantages: it is entirely training-free, requires no annotated data for fine-tuning, and can be directly applied to features extracted from visual-only foundation models such as DINOv2~\citep{DINOv2}.  
However, its effectiveness hinges on the validity of the Doppelgänger assumption and the Rarity of anomalous patterns.
\subsection{Comparison of the Two Approaches}
The two zero-shot paradigms offer complementary advantages and limitations:

\begin{itemize}
\item \textbf{Text-Based Methods}
\begin{itemize}
\item \textbf{Advantages:} Enable fast inference and low computational cost, since each image can be processed independently. They also provide semantic understanding through natural language, allowing the use of descriptive prompts to specify states that are visually acceptable yet semantically distinct from normal conditions—such as textual descriptions of dates or physician markers in medical scans.
\item \textbf{Limitations:} Typically require fine-tuning on additional pixel-level annotated datasets to achieve satisfactory text--visual alignment. They generally yield lower anomaly detection performance than full-shot or few-shot methods and are restricted to visually interpretable anomalies, making them unsuitable for logic-related or contextual defects.
\end{itemize}

\item \textbf{Batch-Based Methods}
\begin{itemize}
    \item \textbf{Advantages:} Entirely training-free, requiring only a pre-trained visual backbone without additional fine-tuning or text supervision. They are particularly suitable for visual-only foundation models and demonstrate strong empirical performance—comparable to full-shot methods and superior to few-shot approaches~\citep{MuSc, CoDeGraph}.  
    \item \textbf{Limitations:} Must process multiple images jointly, which increases both computational cost and memory consumption. Performance can degrade when batch sizes are small or when the dataset violates the Doppelgänger assumption—that is, when collections lack sufficient visual similarity to support mutual comparison.
\end{itemize}

\end{itemize}

In summary, text-based methods emphasize efficiency and semantic instruction understanding, while batch-based methods excel in robustness and training-free deployment. Together, these two paradigms define the current landscape of zero-shot anomaly detection and fully data-independent inspection systems.

% ------------------------------------------------------
\chapter{The Problem of Consistent Anomalies}
\label{chap:empirical}
In this chapter, we examine the feature-level behavior of large vision transformers on industrial anomaly datasets. Our aim is to understand why training-free, batch-based zero-shot methods are effective, and under which conditions they fail. To this end, we introduce the concept of consistent anomalies—a class of similar anomalies that recur systematically across multiple images/collections, violating key assumptions of batch-based approaches and leading to misclassification.

The chapter is structured as follows. Section~\ref{sec:basic-definitions} presents the basic assumptions underlying batch-based methods and introduces the problem of consistent anomalies. Section 3.2 presents two key phenomena—Similarity Scaling and Neighbor-Burnout—and provides empirical evidence for them.

\section{Batch-Based Approaches and Pitfalls}
\label{sec:basic-definitions}

\subsection{Setup and notation}
Let the base set $\mathcal{B}=\{C_1,\dots,C_B\}$ be our test dataset. In this thesis, we may refer $C_i$ as \emph{images} or  \emph{collection} interchangeably.  
A pre-trained encoder $f$ produces element-level features
\[
    f(C_j) = (\mathbf{z}_j^{1}, \dots, \mathbf{z}_j^{N}) \in \mathbb{R}^{N\times D},
\]
where $N$ is the number of elements and $D$ the feature dimension.  
We refer to $\mathbf{z}_i^{h}$ as an element (or patch) of the collection $C_i$.  

Let $d(\cdot,\cdot)$ be a metric on $\mathbb{R}^D$ (e.g., Euclidean).  
For $\mathbf{z}_i^{h}\in C_i$, its distance to another \emph{collection} $C_j$ is defined by the smallest distance to its elements:
\begin{equation}
    d(\mathbf{z}_i^{h}, C_j) \;:=\; \min_{k=1,\dots,N} d\big(\mathbf{z}_i^{h}, \mathbf{z}_j^{k}\big),
    \qquad j\neq i .
    \label{eq:dist-to-collection}
\end{equation}
Sorting $\{ d(\mathbf{z}_i^{h}, C_j) : j\neq i\}$ increasingly, we denote the $t$-th order statistic by \( d(\mathbf{z}_i^{h})_{(t)} \) and define the \emph{mutual similarity vector} as
\begin{equation}
    \mathcal{D}_{\mathcal{B}}(\mathbf{z}_i^{h}) \;=\; [\, d(\mathbf{z}_i^{h})_{(1)}, \ldots, d(\mathbf{z}_i^{h})_{(B-1)} \,].
\label{eq:msv}
\end{equation}

Below, we use $\mathbf{z}_n$ for a normal element, $\mathbf{z}_a$ for an anomalous element, and $\mathbf{z}$ for a general element of a collection ${C}$.
We now present the concepts of $\epsilon$-neighbors and $\epsilon$-consistency of a patch $\mathbf{z}$.
\begin{definition}[$\epsilon$-neighbors and $\epsilon$-consistency]
\label{def:eps-neighbor-consistency}
Given $\epsilon>0$ and an element $\mathbf{z}\!\in\!C_i$, define its \emph{$\epsilon$-neighbor set} across the base set $\mathcal{B}$ as
\[
    \mathcal{N}_{\mathcal{B}}(\mathbf{z},\epsilon) 
    \;:=\; 
    \bigl\{\, j\in\{1,\dots,B\}\setminus\{i\} 
    \;\big|\; d(\mathbf{z},C_j) < \epsilon \,\bigr\},
    \qquad 
    |\mathcal{N}_{\mathcal{B}}(\mathbf{z},\epsilon)| \text{ denotes its size}.
\]
When such a relation exists between $\mathbf{z}$ and a collection $C_j$, we refer to it as an \emph{$\epsilon$-consisent link}, or simply $\epsilon$-link.  
An element $\mathbf{z}$ is said to be \emph{$\epsilon$-consistent} if 
$|\mathcal{N}_{\mathcal{B}}(\mathbf{z},\epsilon)| \ge 1$.
\end{definition}
In this book, when the base set $\mathcal{B}$ is clear from context, we may omit it from the notation and simply write such as
$\mathcal{N}(\mathbf{z},\epsilon)$ or $\mathcal{D}(\mathbf{z})$ for brevity.
\subsection{The Problem of Consistent Anomalies}
\begin{figure}[t!]
  \vspace*{-1cm}
  \centering
  \includegraphics[width=0.95\textwidth]{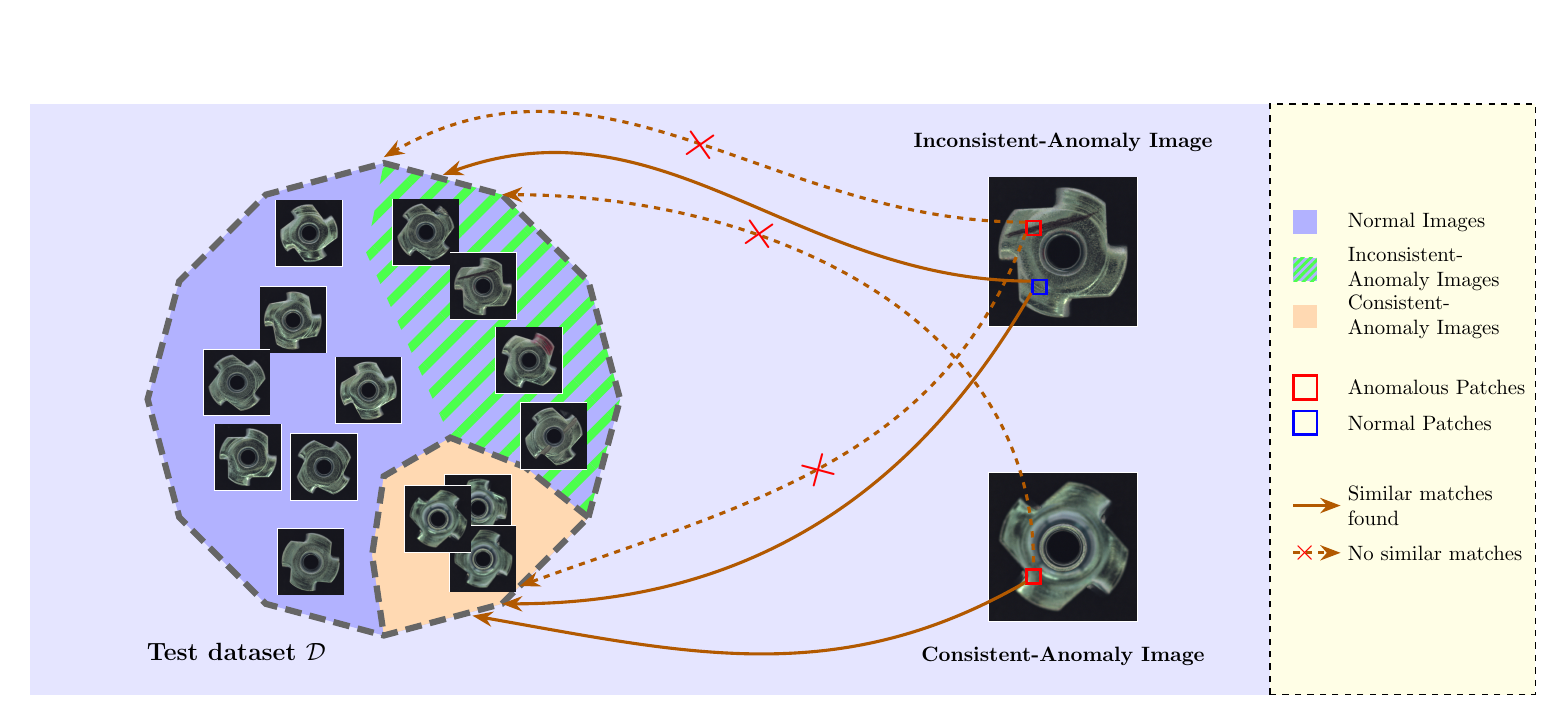}
  \caption{Illustration of zero-shot anomaly detection's consistent-anomaly problem. Industrial images have normal patches (blue squares) that match nearly all test images.   Scratches and other random anomalies have high anomaly scores since they fail to find similar matches across the test set. Defects from consistent-anomaly images (flipped metal nuts) easily find deceptive matches within the images (orange region) sharing the same anomaly pattern (rotate counter-clockwise instead of clockwise).}
  \label{fig:voronoi}
\end{figure}
Before introducing the main contents, we note that the backbone model $f$ is assumed to be \emph{semantically well-trained}: its feature space meaningfully reflects perceptual similarity, allowing us to define a semantic threshold that separates semantically similar from dissimilar elements.

\begin{definition}[Semantic threshold $\epsilon_0$]
\label{def:semantic-threshold}
A radius $\epsilon_0>0$ is fixed as a \emph{semantic threshold}: if $d(\mathbf{x},\mathbf{y})>\epsilon_0$, the pair $(\mathbf{x},\mathbf{y})$ is regarded as \emph{semantically different}.
\end{definition}

\vspace{0.5em}
\noindent
The effectiveness of batch-based zero-shot methods implicitly depends on two assumptions about the base set $\mathcal{B}$.  
The first, the \emph{Doppelgänger} assumption, assumes a decently stable structure between collections: most normal elements can find similar counterparts in other collections.  
The second, the \emph{Rarity} assumption, represents the random and sparse nature of anomalies: anomalous elements are expected to appear irregularly and remain mostly isolated in the feature space.

\begin{assumption}[Doppelgänger]
\label{assump:doppelganger}
There exist $\epsilon \ll \epsilon_0$ and an integer $K$ with $1\!\ll\! K\!<\!B$ such that, for \emph{almost all} normal elements $\mathbf{z}_n$ in $\mathcal{B}$,
\[
    |\mathcal{N}(\mathbf{z}_n,\epsilon)| \;\ge\; K .
\]
That is, each normal element finds many close matches (“doppelgängers”) in other collections.
\end{assumption}

\begin{assumption}[Rarity of anomalies]
\label{assump:rarity}
There exists a small integer $0 \le H \ll B$ such that, for \emph{almost all} anomalous elements $\mathbf{z}_a$ in $\mathcal{B}$,
\[
    |\mathcal{N}(\mathbf{z}_a,\epsilon_0)| \;\le\; H .
\]
Thus anomalies are unique or only weakly matched; they do not accumulate many near-duplicates across the dataset.
\end{assumption}

\vspace{0.8em}
\noindent
For industrial images, it is typical that $H \ll K$. Under these assumptions, MuSc \cite{MuSc} proposed to aggregate the first $K$ entries of $\mathcal{D}(\mathbf{z})$ rather than all $B-1$ values when computing a normality score:
\begin{equation}
    \label{equation:anomaly_score}
    a(\mathbf{z}) \;=\; \frac{1}{K}\sum_{t=1}^{K} d(\mathbf{z})_{(t)},
\end{equation}
where a lower value indicates higher normality.
For a normal element $\mathbf{z}_n$, the Doppelgänger assumption ensures that $a(\mathbf{z}_n)$ remains small and that using only the first $K$ neighbors reduces the influence of distant or noisy matches near the boundary $B-1$. On the other hand, the Rarity assumption inflates the anomaly score of anomalous patches if $K \gg H$.

Now consider an anomalous element $\mathbf{z}_a$ such that $|\mathcal{N}(\mathbf{z}_a,\epsilon)| = H$ with $H<K$.  
Thus its first $H$ distances are less than $\epsilon$, i.e.,
\(
    d(\mathbf{z}_a)_{(i)} < \epsilon
\) for $i \le H$. Then,
\begin{equation}
\label{eq:upper-bound-topK}
K\,a_{\mathcal{B}}(\mathbf{z}_a) \;\le\; H\,\epsilon \;+\; \sum_{t=H+1}^{K} d(\mathbf{z}_a)_{(t)} .
\end{equation}
When $\epsilon$ is small and $H$ is moderately large, the first term on the right-hand side can dominate, pulling $a_{\mathcal{B}}(\mathbf{z}_a)$ downward.  
If, in addition, some of the remaining distances $d(\mathbf{z}_a)_{(t)}$ ($t>H$) are not large—e.g., because multiple anomalous collections share similar structures—the resulting score may become \emph{comparable to or even smaller than} that of normal elements.  
This violates the Rarity assumption and causes significant degradation in top-$K$–based anomaly scoring, as shown in our work~\cite{CoDeGraph}.

In practice, the \emph{Doppelgänger} assumption is often valid because normal samples exhibit stable, repeated structures across collections in a predicted way.  
However, the \emph{Rarity of anomalies} is far less reliable.  
An error in a production line may generate clusters of similar defects, such as paint that is stuck and consequently creates missing print areas.
In such datasets, an anomalous element $\mathbf{z}_a$ may have a large $|\mathcal{N}(\mathbf{z}_a,\epsilon)|$, which keeps the upper bound~\eqref{eq:upper-bound-topK} small and deceptively causes top-$K$ aggregation methods to misclassify anomalies as normal.

\begin{definition}[$\epsilon$-consistent anomaly]
\label{def:eps-consistent-anomaly}
An anomalous element $\mathbf{z}_a$ is called an \emph{$\epsilon$-consistent anomaly}
if $|\mathcal{N}(\mathbf{z}_a,\epsilon)| \ge 1$, and we call the \emph{$\epsilon$-consisent links} between $\mathbf{z}_a$ and its neighbors in $\mathcal{N}(\mathbf{z}, \epsilon)$ as \emph{$\epsilon$-consistent-anomaly links}.
If $|\mathcal{N}(\mathbf{z}_a,\epsilon)| = H$, we call $\mathbf{z}_a$
an \emph{$H$-level $\epsilon$-consistent anomaly}. 
  
\end{definition}

\noindent
\begin{remark}
    Throughout this book, our interest lies in $H$‑level $\epsilon$‑consistent
anomalies with large $H$ and sufficiently small $\epsilon$
(smaller than a typical distance between normal elements). Consequently,
when we speak of consistent anomalies we refer to $H$‑level
$\epsilon$‑consistent anomalies that satisfy those two conditions.
\end{remark}
We now formally introduce the central notion of this chapter and this book—the
problem of consistent anomalies. An illustration for the problem of consistent anomalies is presented in Figure~\ref{fig:voronoi}.

\begin{definition}[Consistent anomaly problem]
\label{def:consistent-anomaly-problem}
A dataset is said to exhibit the \emph{consistent anomaly problem} if it contains $H$-level $\epsilon$-consistent anomalies with small $\epsilon$ (high intra-anomaly similarity) and large $H$ (many repetitions across collections).  
Such datasets violate the Rarity assumption and cause batch-based zero-shot methods to misclassify anomalies as normal.
\end{definition}
\subsection{Empirical Evidence of Consistent Anomalies}
\label{sec:empirical-evidence}
\begin{figure}[t]
\centering

% First row: cable example
\begin{subfigure}{\textwidth}
\centering
\includegraphics[width=0.9\textwidth]{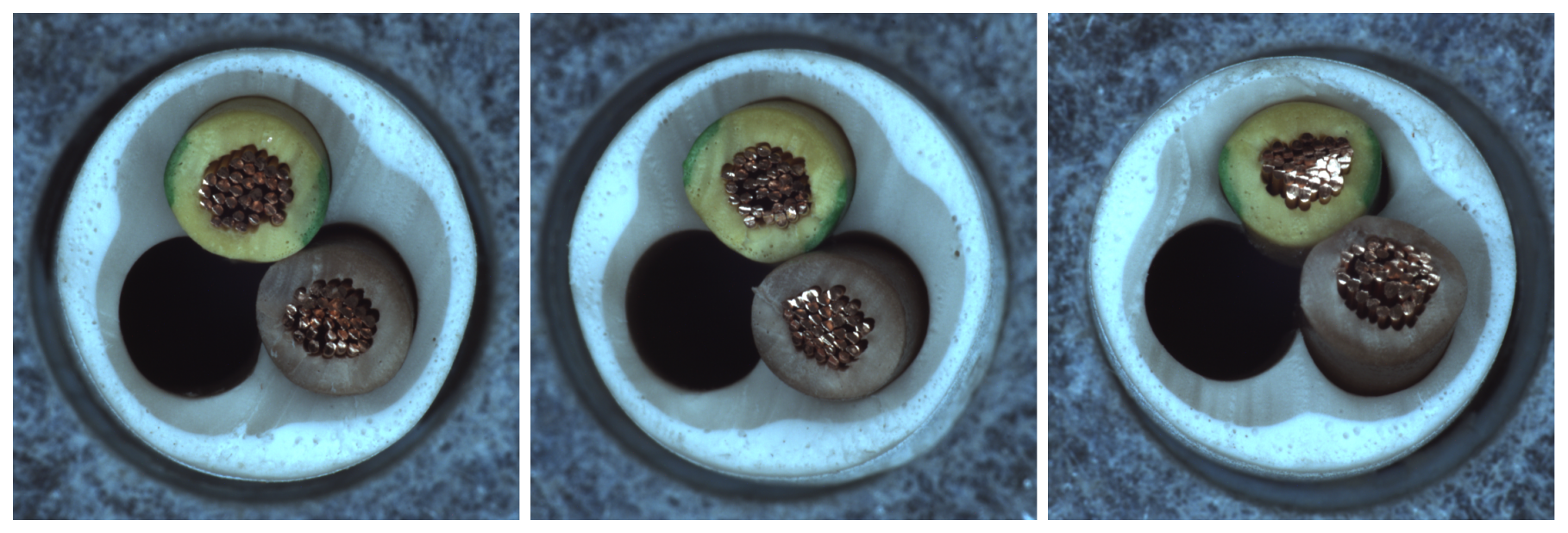}
\caption{Cable (MVTec AD)}
\label{fig:cable-consistent}
\end{subfigure}

\vspace{1em} % vertical space between rows

% Second row: tumor example
\begin{subfigure}{\textwidth}
\centering
\includegraphics[width=0.9\textwidth]{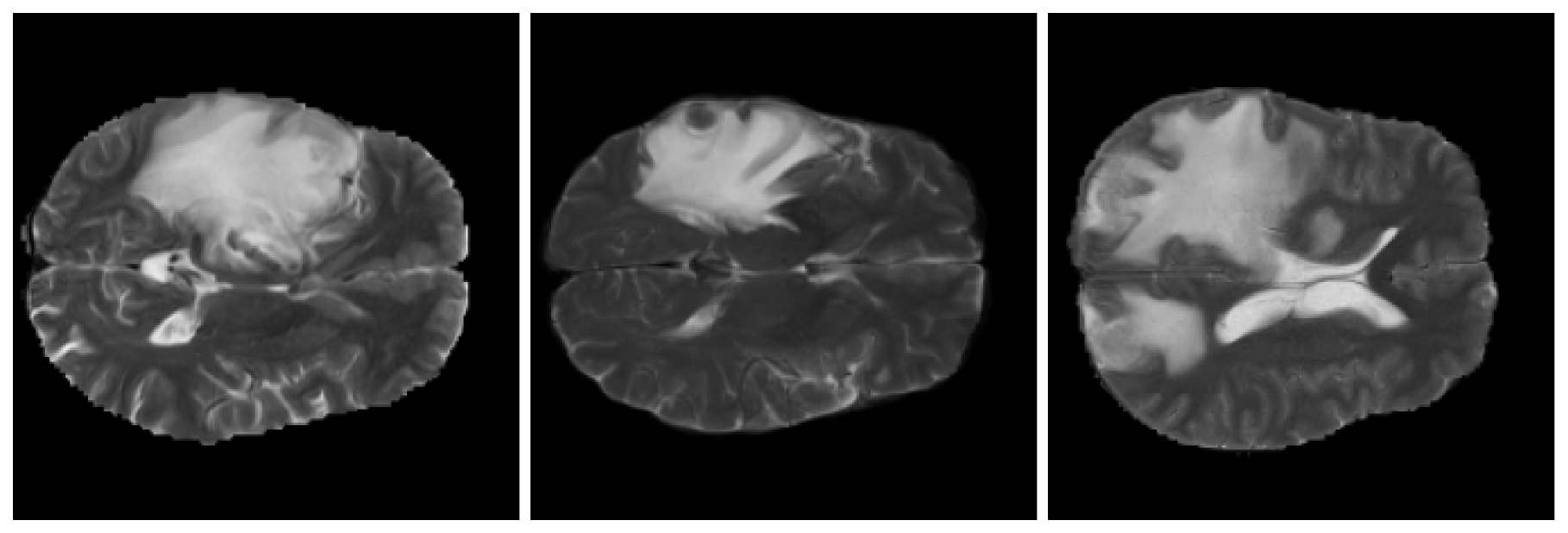}
\caption{Brain MRI (BraTS-2025)}
\label{fig:tumor-consistent}
\end{subfigure}

\caption{Representative examples of consistent anomalies. 
    Top: multiple cable samples from MVTec AD~\cite{mvtec} exhibiting the ``missing components". 
         Bottom: multiple brain MRI from BraTS-2025~\cite{brats} showing similar enhancing tumor regions with high signal intensity on post-contrast T2-weighted images.
         These illustrate that consistent anomalies appear frequently in both industrial and medical domains.}
\label{fig:consistent-anomaly-examples}

\end{figure}
Empirical inspection across several datasets confirms that consistent anomalies are sufficiently common and present a real challenge for zero-shot anomaly classification and segmentation.  
Figure~\ref{fig:consistent-anomaly-examples} illustrates representative examples from industrial and medical domains, including samples from popular benchmarks such as MVTec AD~\citep{mvtec} and BraTS~\citep{brats}.  
These examples demonstrate that multiple test images can share similar anomalous structures, i.e., consistent anomalies.

Consistent anomalies are not limited to two-dimensional image data.  
They have also been observed in other modalitie such as volumetric MRI scans with recurrent tumor morphologies, which we will study in Chapter~\ref{tab:mri3d_results}. 
These observations confirm that the phenomenon extends beyond industrial imagery and potentially exists in any modality where abnormal patterns follow reproducible generation mechanisms.

Quantitatively, the share of images containing consistent anomalies within the test sets we analyzed is about \(5\%\)–\(20\%\), and the size of such anomalies ranges from medium to large regions. 
Although this fraction represents only a minority of the total samples, its presence within a batch is sufficient to alter the statistical landscape, hence degrading the performance of AC/AS methods that are not explicitly designed to handle such anomalies.

It is also evident that text-driven zero-shot methods are unable to reliably detect such cases.  
For instance, in the \texttt{metal\_nut} class, flipped metal nuts are semantically compatible with the textual prompt like ``A photo of a flawless metal nut'' and thus cannot be recognized as anomalies using text embeddings alone~\citep{WinCLIP}.  
Their detection would require either access to ground-truth supervision~\citep{WinCLIP} or an observation from the dataset-level~\cite{CoDeGraph} as described below.

\subsubsection*{Philosophy of consistent anomaly detection}
The batch formulation of batch-based methods offers a natural pathway for zero‑shot detection of consistent anomalies.  
It is founded on a simple yet fundamental principle: patterns that recur systematically across a small set of collections, yet appear far less frequently than other patterns within the batch, should be regarded as anomalous.  
For instance, in the \texttt{metal\_nut} class, flipped metal nuts should be considered anomalies since they occur less than $20\%$ of the samples, whereas other images ($80\%$) share the same normal patterns.  
On the other hand, if an anomalous pattern becomes more frequent than any normal type, there are no clear criterion—statistical or perceptual—to classify it as abnormal, even for human observers.  
Therefore, consistent anomalies can only be identified when they appear as a minority of the samples in the test set.

\begin{assumption}[Minority of consistent anomalies]
\label{assump:limited-consistent}
In the test dataset, the number of samples that share the same consistent anomaly never exceeds the number of normal samples of the same type.
\end{assumption}
The Minority assumption formalizes the philosophical basis described above.  
It guarantees that detecting consistent anomalies is solvable under realistic conditions and remains meaningful for datasets with multiple normal variants.  
For example, in the \texttt{juice\_bottle} class of MVTec LOCO, where normal images correspond to three bottle types (apple, orange, and banana), if consistent‑anomaly images outnumber any of these normal variants, the distinction between normal and abnormal states becomes fundamentally ambiguous.  
\begin{remark}
The existence of consistent anomalies does not contradict the Rarity assumption (Assumption~\ref{assump:rarity}); rather, it generalizes it.  
While Assumption~\ref{assump:rarity} states that almost all anomalies possess at most a few $\epsilon_0$‑neighbors, the present definition allows for a non‑negligible subset of anomalies whose neighbor sets are considerably larger, i.e., $|\mathcal{N}(\mathbf{z}_a,\epsilon_0)| = H$ with $H \gg 1$.  
This generalization reflects a more realistic data regime in which most anomalies remain isolated, but some emerge repeatedly across collections.
\end{remark}
\section{The Two Phenomena of Batch-Based Approaches}
\label{sec:two-phenomena}
The aim of this section is to present empirical evidence for our two phenomena: \emph{Similarity Scaling} and \emph{Neighbor-Burnout}. The Similarity Scaling phenomenon describes the behaviour of the normal token $\mathbf{z}$ across its neighbours, namely how $d(\mathbf{z})_{(i)}$ in the mutual‑similarity vector $\mathcal{D}(\mathbf{z})$ grows with rank $i$. The \textit{Neighbor-Burnout} phenomenon characterizes a statistical deviation from this growth, when consistent anomalies are present. Together, these two phenomena provide an empirical foundation for understanding the consistent-anomaly problem and motivate the method as well as the theoretical
analysis developed later in Chapter~\ref{chap:algorithm} and Chapter~\ref{chap:theory-scaling}.
\subsection{Experimental Setup}
\label{sec:exp-setup}

All experiments presented in this section were conducted under a controlled, training-free configuration. The purpose of this setup is to determine whether the two phenomena can already be observed at the feature level of existing vision transformers, without any additional fine-tuning or supervision.

\paragraph{Encoders}  
Two vision transformer architectures were examined:ViT-L/14@336, trained by OpenAI~\citep{CLIP} using image--text contrastive objectives, and DINOv2-L/14, trained by Meta~\citep{DINOv2} using self-supervised learning. Both models process images at the patch size of $14$. These models share the same ViT architecture but differ in their training supervision, allowing us to test whether the observed phenomena are model-agnostic.
\paragraph{Feature extraction.}  
All images were resized to $518\times518$ before feature extraction. At this resolution, each layer produces $1369$ spatial patch tokens per image. Each vision transformer outputs a sequence of patch embeddings at every layer. For analysis, two representative layers were selected: the $6^{th}$ layer, capturing fine-grained local features, and the $24^{th}$ layer, encoding high-level semantic information.

All patch embeddings were $\ell_2$-normalized prior to computing pairwise distances. Unless otherwise stated, distances are reported in Euclidean distance.

\paragraph{Datasets.}  
Results are presented on selected classes from widely used anomaly detection benchmarks. The MVTec dataset~\citep{mvtec} contains industrial objects with diverse defects, ranging from subtle surface anomalies (e.g., scratches) to structural and logical faults (e.g., missing components). MVTec Loco~\citep{mvtec-loco} emphasizes logical anomalies. For medical imaging, we included 2D slcies of BraTS-METS~\citep{brats}, a 3D MRI dataset of brains with metastatic tumors, and IXI~\citep{ixi}, a dataset of healthy MRI brain scans. Specifically, we selected the largest axial slice from each 3D volume after applying a standard preprocessing pipeline. Following~\citep{brats}, all 3D MRI scans were first registered to the SRI-24 template and then skull-stripping using HD-BET~\cite{hd-bet}. To minimize the presence of zero-value voxels, all MRI scans were cropped to $196\times196\times196$, resized to $224\times224\times224$, and processed with histogram standardization~\citep{nyul2000new}. This resulted totally $905$ images with $711$ normal samples.

This configuration provides a consistent and reproducible framework for evaluating the intrinsic statistical behaviour of patch tokens across modalities. The following subsections present the two empirical phenomena—Similarity Scaling and Neighbor-Burnout—that emerge from this setup.
\subsection{Empirical Evidence of the Similarity Scaling Phenomenon}
\label{sec:similarity-scaling}
\clearpage
\thispagestyle{empty}
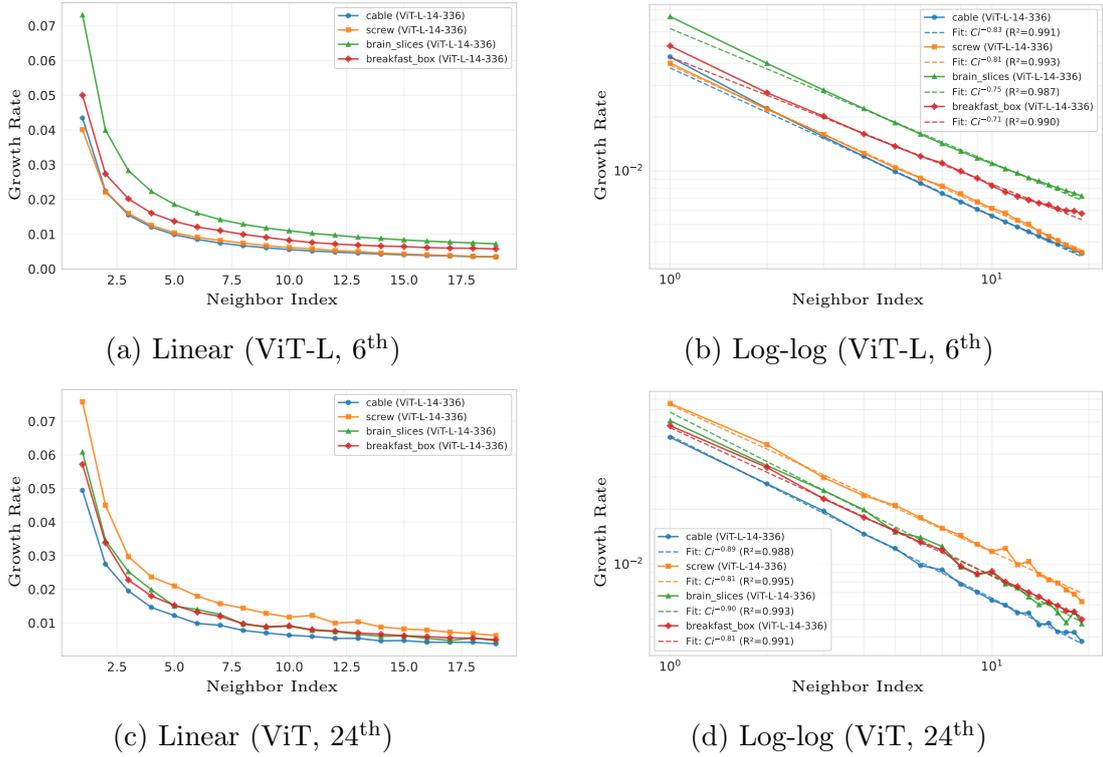
\begin{figure}[t] % or [htbp]
  \centering

  % ====== block 1: ViT ======
  \begin{minipage}{0.98\textwidth}
    \centering
    % row 1
    \begin{subfigure}{0.465\textwidth}
      \centering
      \begin{tikzpicture}
        \node[inner sep=0pt] (img) {\includegraphics[width=.98\textwidth]{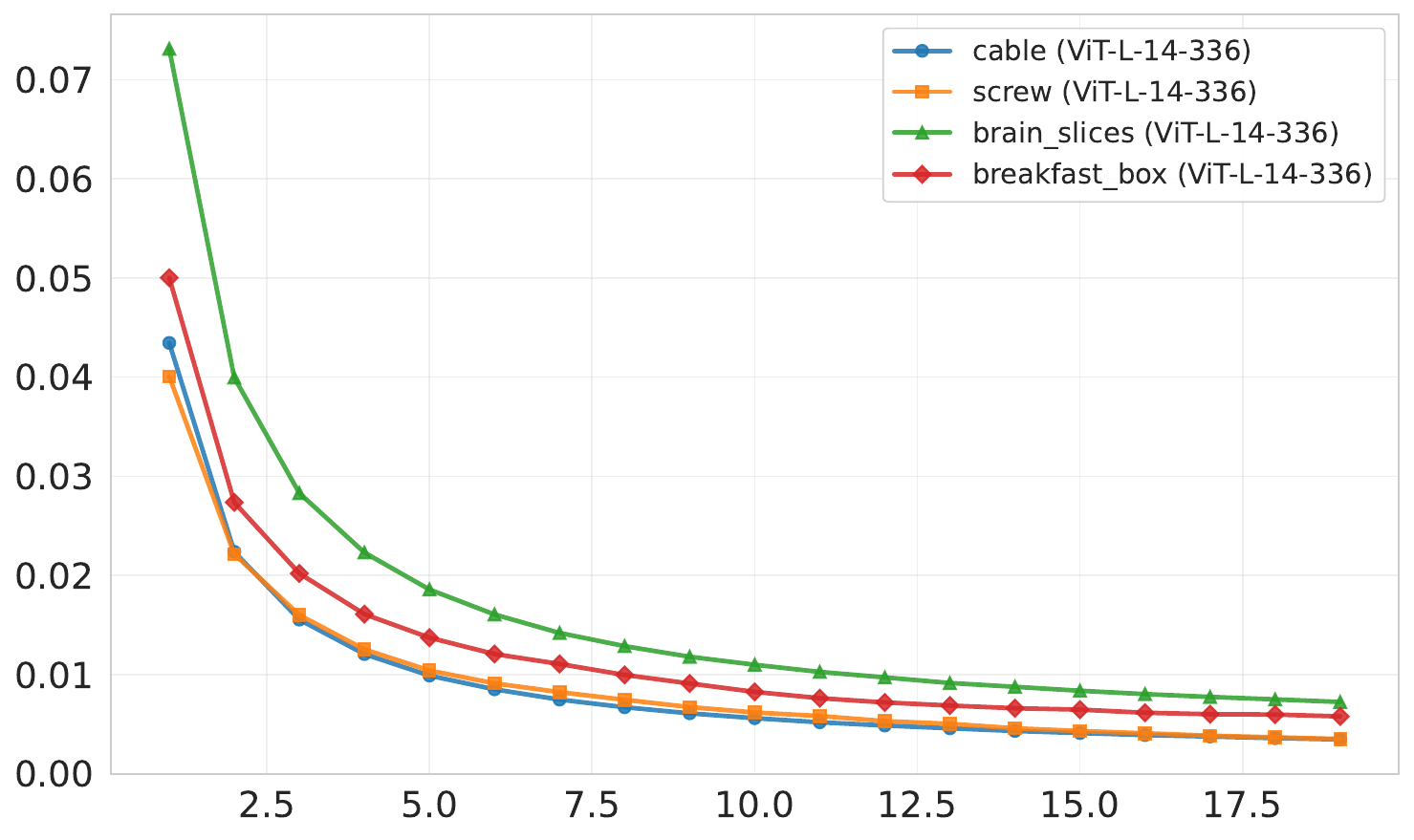}};
        \node[rotate=90, anchor=center] at ([xshift=-0.1cm]img.west) {\tiny Growth Rate};
        \node[anchor=center] at ([yshift=-0.1cm]img.south) {\tiny Neighbor Index};
      \end{tikzpicture}
      \caption{Linear (ViT-L, $6^{\text{th}}$)}
    \end{subfigure}\hfill
    \begin{subfigure}{0.465\textwidth}
      \centering
      \begin{tikzpicture}
        \node[inner sep=0pt] (img) {\includegraphics[width=.98\textwidth]{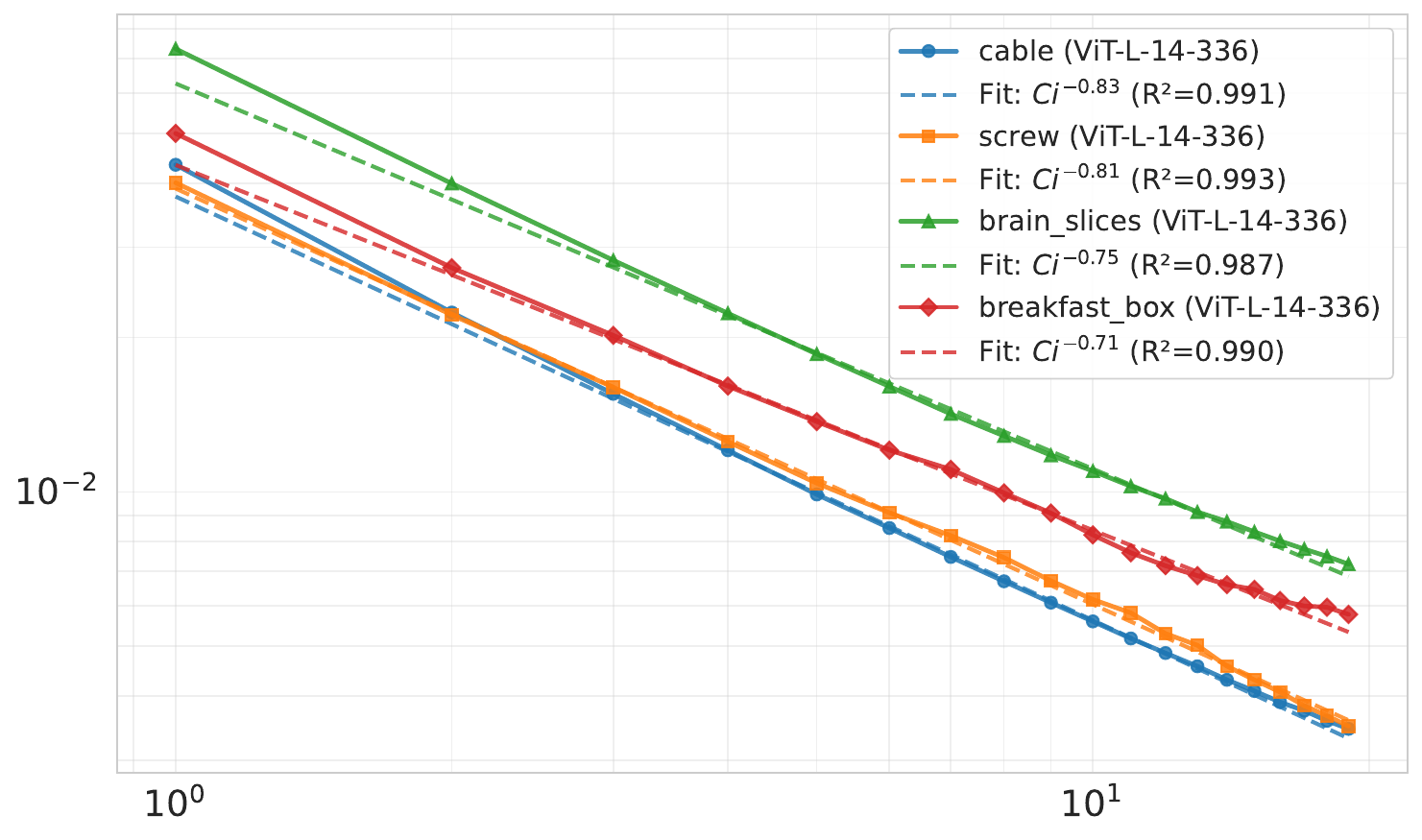}};
        \node[rotate=90, anchor=center] at ([xshift=-0.1cm]img.west) {\tiny Growth Rate};
        \node[anchor=center] at ([yshift=-0.1cm]img.south) {\tiny Neighbor Index};
      \end{tikzpicture}
      \caption{Log-log (ViT-L, $6^{\text{th}}$)}
    \end{subfigure}

    \vspace{0.4em} % was 0.8em

    % row 2
    \begin{subfigure}{0.465\textwidth}
      \centering
      \begin{tikzpicture}
        \node[inner sep=0pt] (img) {\includegraphics[width=.98\textwidth]{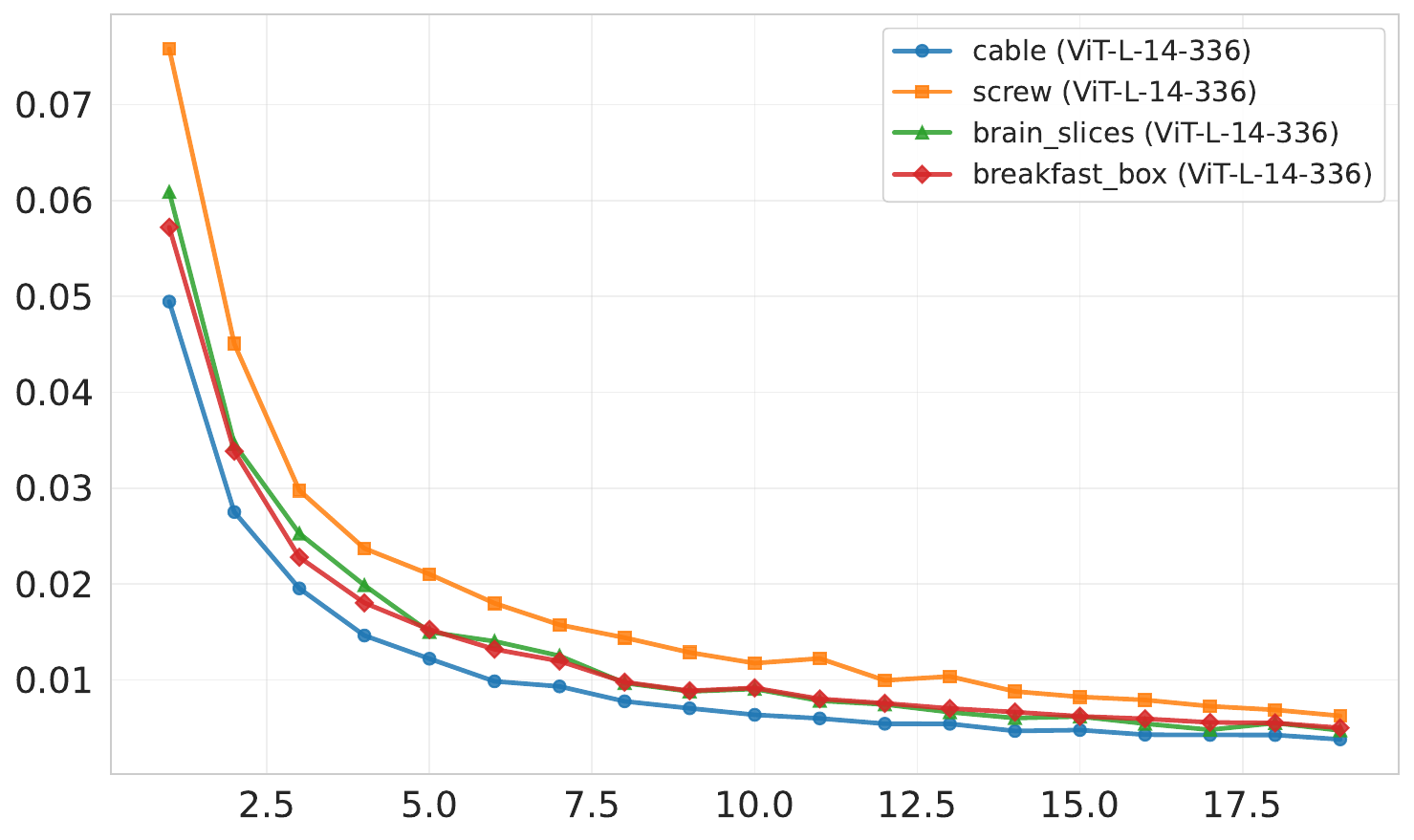}};
        \node[rotate=90, anchor=center] at ([xshift=-0.1cm]img.west) {\tiny Growth Rate};
        \node[anchor=center] at ([yshift=-0.1cm]img.south) {\tiny Neighbor Index};
      \end{tikzpicture}
      \caption{Linear (ViT, $24^{\text{th}}$)}
    \end{subfigure}\hfill
    \begin{subfigure}{0.465\textwidth}
      \centering
      \begin{tikzpicture}
        \node[inner sep=0pt] (img) {\includegraphics[width=.98\textwidth]{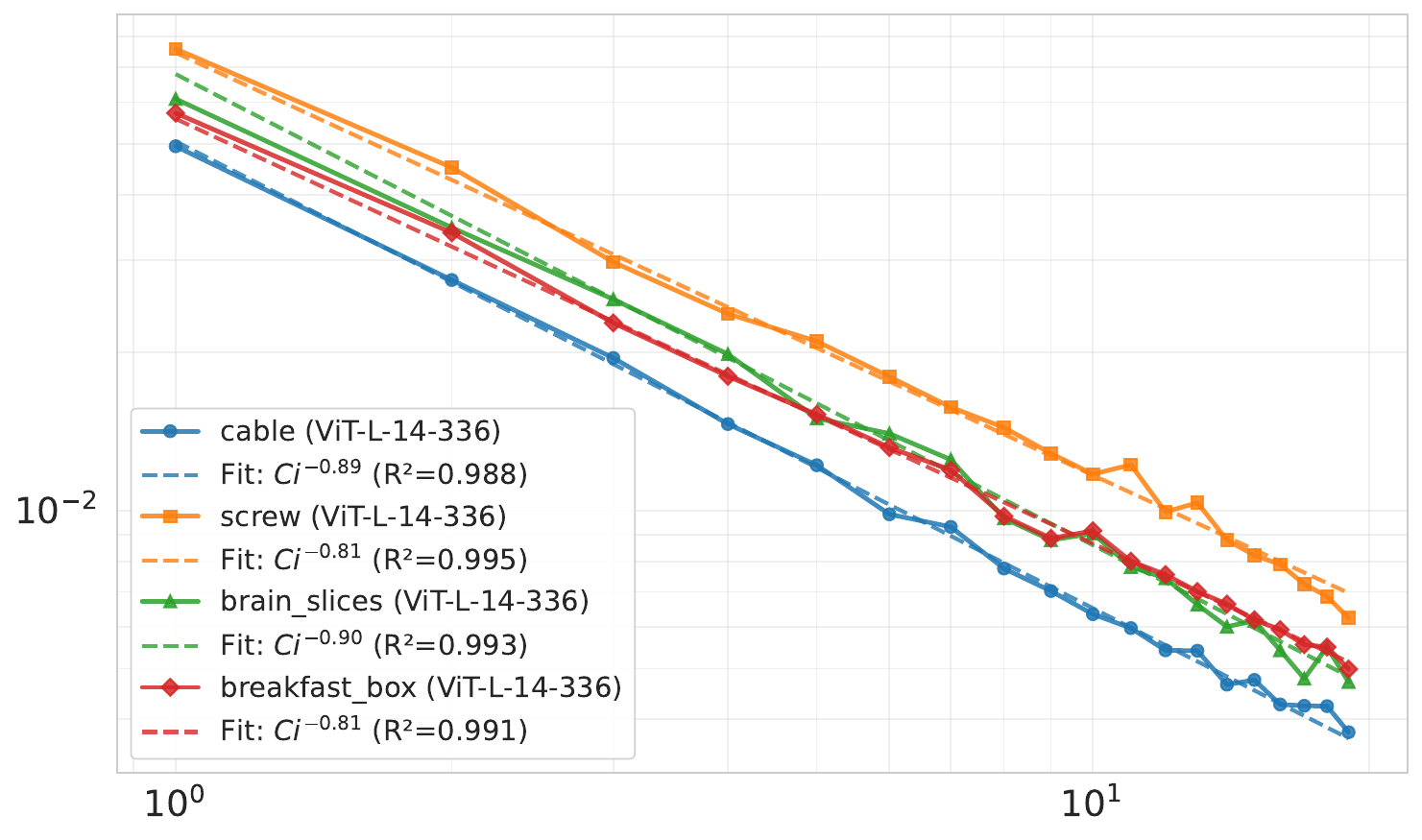}};
        \node[rotate=90, anchor=center] at ([xshift=-0.1cm]img.west) {\tiny Growth Rate};
        \node[anchor=center] at ([yshift=-0.1cm]img.south) {\tiny Neighbor Index};
      \end{tikzpicture}
      \caption{Log-log (ViT, $24^{\text{th}}$)}
    \end{subfigure}

    \caption{Similarity scaling in ViT-L/14@336.}
    \label{fig:sim_scaling_vit_all}
  \end{minipage}
\end{figure}

\begin{figure}[t]
  % ====== block 2: DINOv2 ======
  \begin{minipage}{0.98\textwidth}
    \centering
    % row 1
    \begin{subfigure}{0.465\textwidth}
      \centering
      \begin{tikzpicture}
        \node[inner sep=0pt] (img) {\includegraphics[width=.98\textwidth]{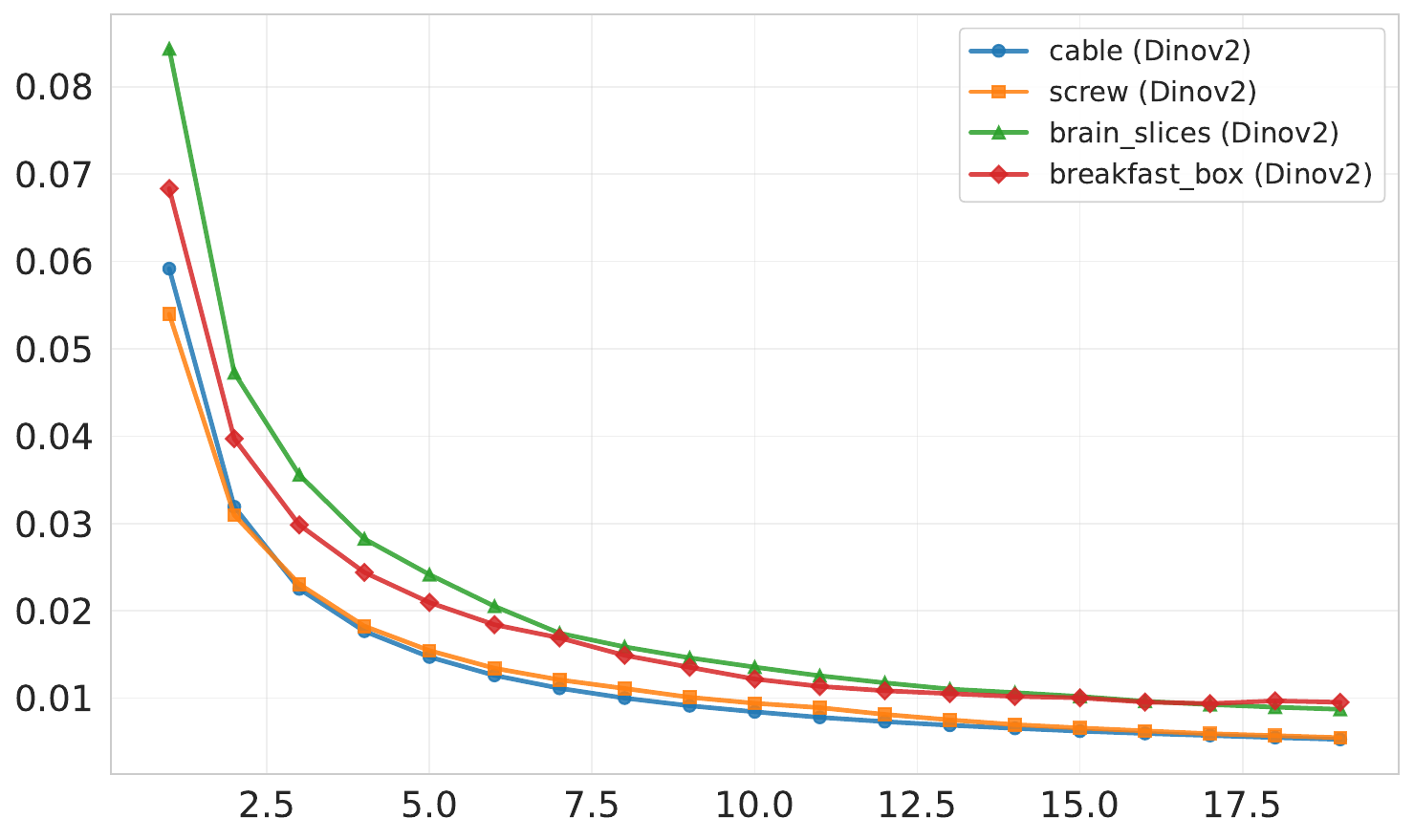}};
        \node[rotate=90, anchor=center] at ([xshift=-0.1cm]img.west) {\tiny Growth Rate};
        \node[anchor=center] at ([yshift=-0.1cm]img.south) {\tiny Neighbor Index};
      \end{tikzpicture}
      \caption{Linear (DINOv2, $6^{\text{th}}$)}
    \end{subfigure}\hfill
    \begin{subfigure}{0.465\textwidth}
      \centering
      \begin{tikzpicture}
        \node[inner sep=0pt] (img) {\includegraphics[width=.98\textwidth]{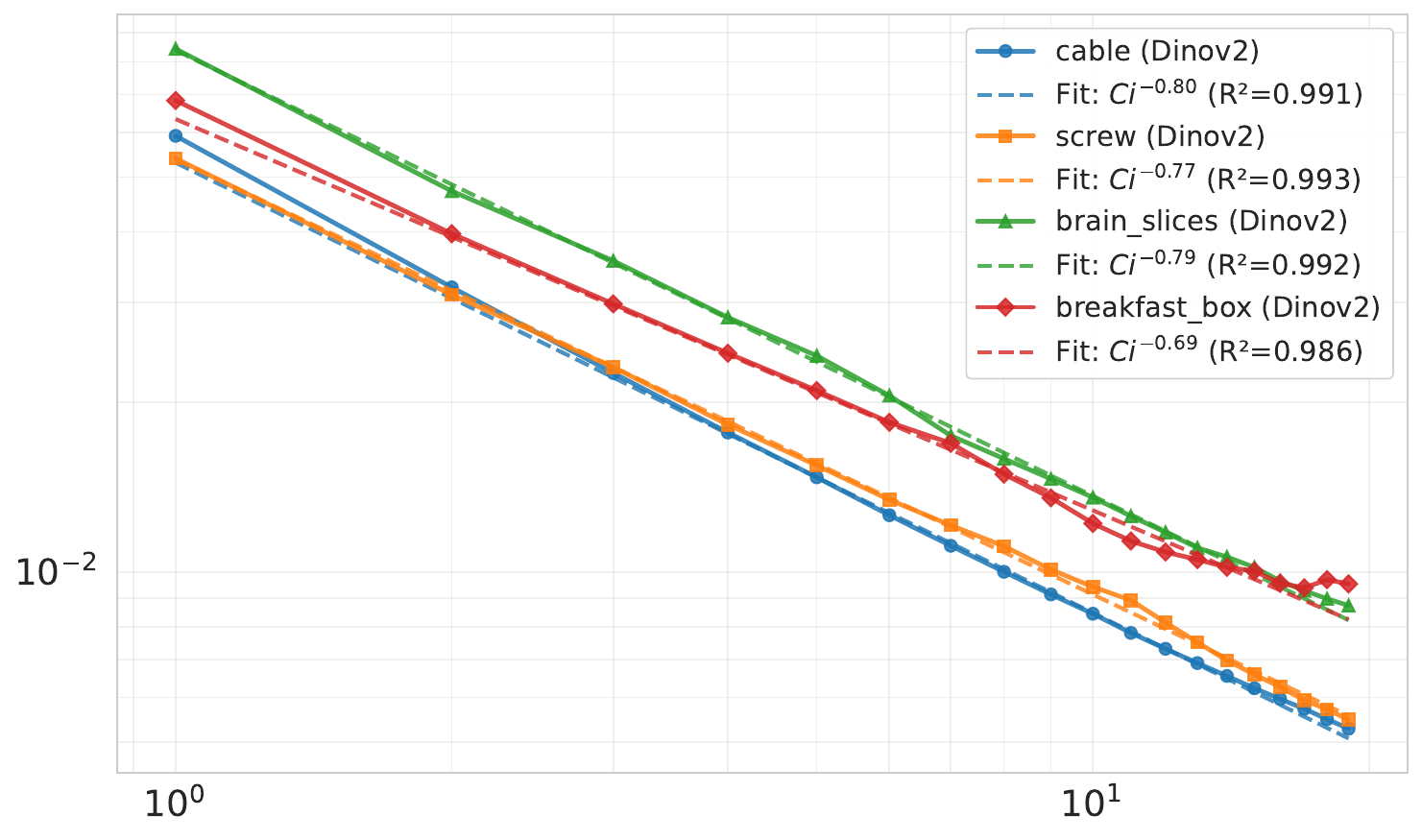}};
        \node[rotate=90, anchor=center] at ([xshift=-0.1cm]img.west) {\tiny Growth Rate};
        \node[anchor=center] at ([yshift=-0.1cm]img.south) {\tiny Neighbor Index};
      \end{tikzpicture}
      \caption{Log-log (DINOv2, $6^{\text{th}}$)}
    \end{subfigure}

    \vspace{0.4em}

    % row 2
    \begin{subfigure}{0.465\textwidth}
      \centering
      \begin{tikzpicture}
        \node[inner sep=0pt] (img) {\includegraphics[width=.98\textwidth]{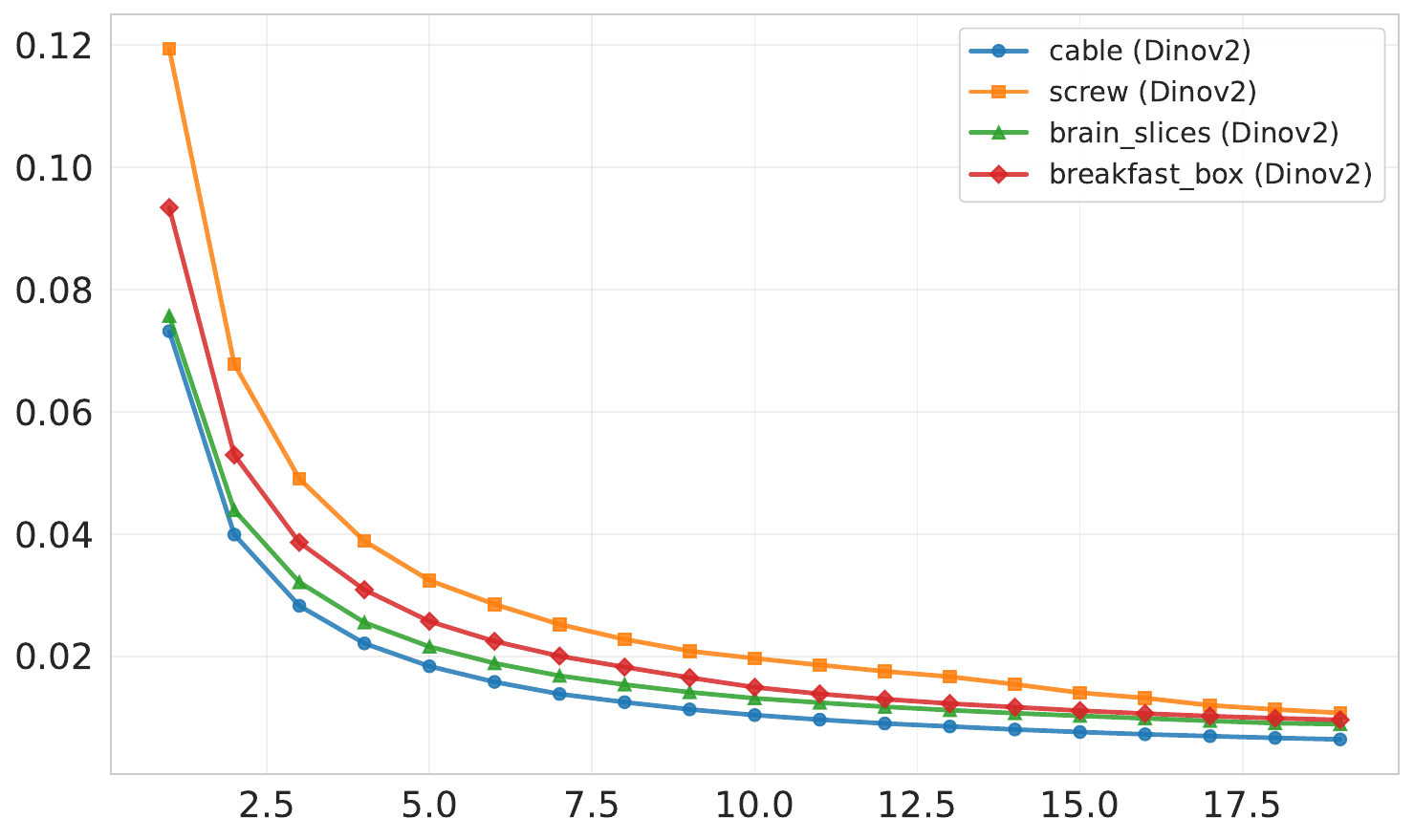}};
        \node[rotate=90, anchor=center] at ([xshift=-0.1cm]img.west) {\tiny Growth Rate};
        \node[anchor=center] at ([yshift=-0.1cm]img.south) {\tiny Neighbor Index};
      \end{tikzpicture}
      \caption{Linear (DINOv2, $24^{\text{th}}$)}
    \end{subfigure}\hfill
    \begin{subfigure}{0.465\textwidth}
      \centering
      \begin{tikzpicture}
        \node[inner sep=0pt] (img) {\includegraphics[width=.98\textwidth]{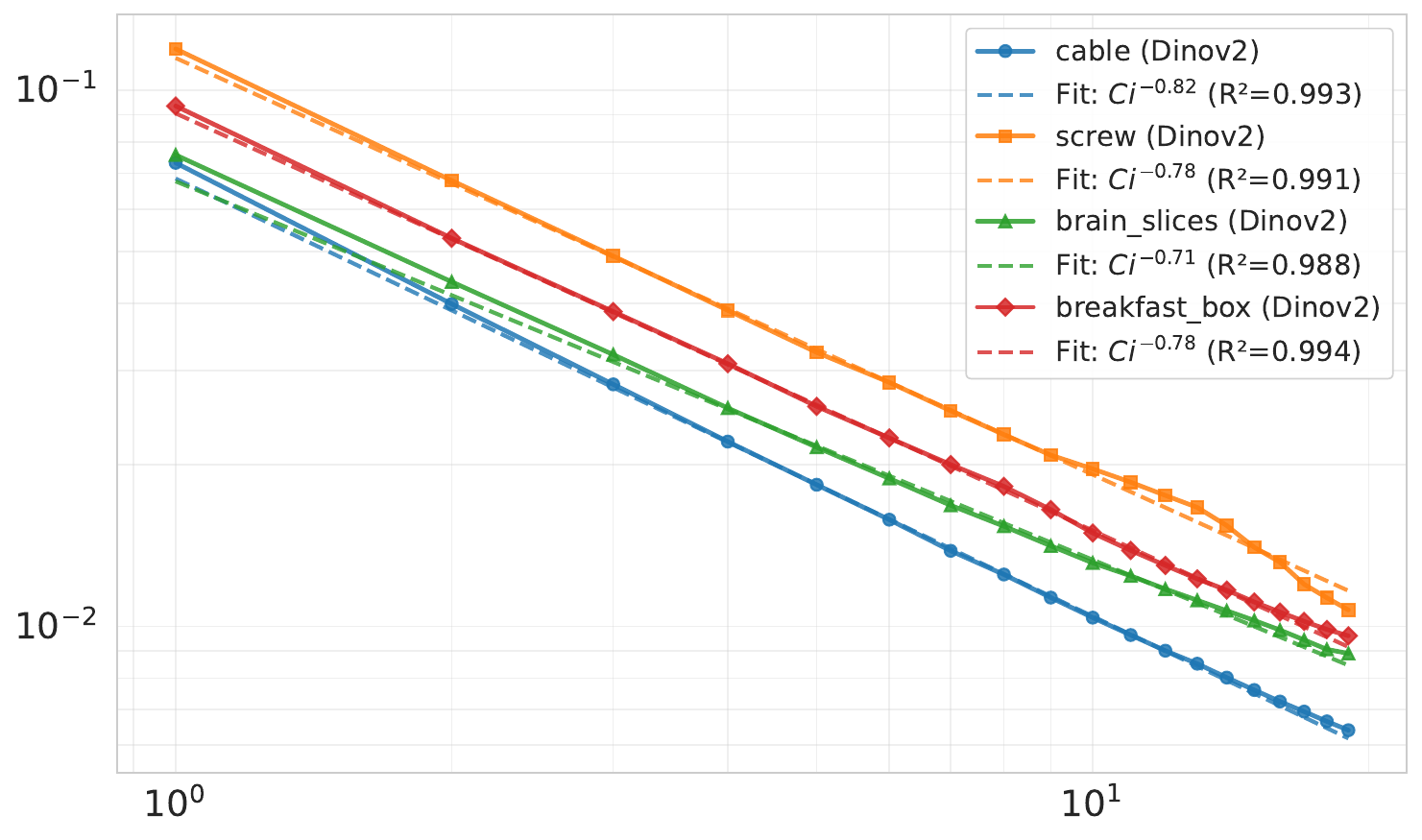}};
        \node[rotate=90, anchor=center] at ([xshift=-0.1cm]img.west) {\tiny Growth Rate};
        \node[anchor=center] at ([yshift=-0.1cm]img.south) {\tiny Neighbor Index};
      \end{tikzpicture}
      \caption{Log-log (DINOv2, $24^{\text{th}}$)}
    \end{subfigure}

    \caption{Similarity scaling in DINOv2.}
    \label{fig:sim_scaling_dino_all}
  \end{minipage}

\end{figure}
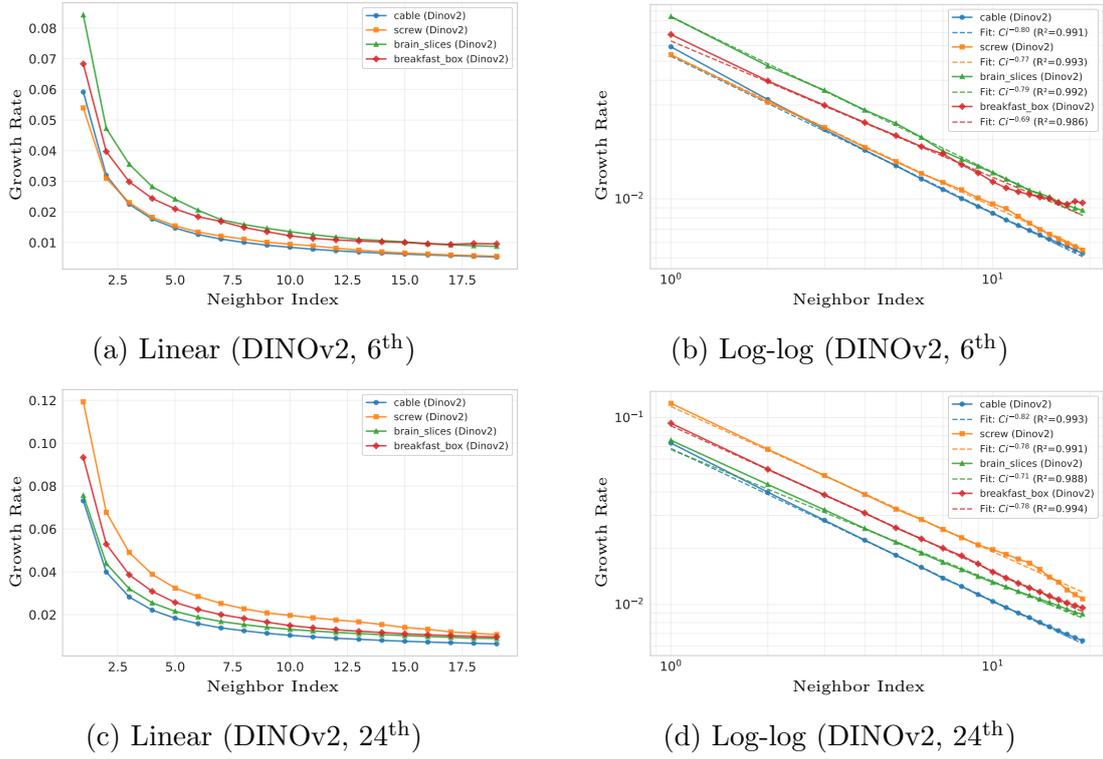

The Similarity Scaling phenomenon summerizes our observations about the way distance of normal patches to nearby collections behave as we move to increasingly distant neighbors. Establishing this empirical baseline is essential, since all subsequent analyses of anomalies—including the neighbor-burnout phenomenon—are formulated as statistical deviations from this law.

Under the Doppelgänger assumption (Assumption~\ref{assump:doppelganger}), most normal elements $\mathbf{z}_n$ possess many close counterparts within the base set $\mathcal{B}$, forming locally dense clusters in the feature space. Consequently, the expectation is that their mutual similarity vectors exhibit smooth and predictable variation. To examine this behavior, we consider the (log) growth rate of mutual similarity vector $\mathcal{D}(\mathbf{z})$  as
\begin{equation}
\label{eq:tau-def}
\tau_i(\mathbf{z}) := \ln \frac{d(\mathbf{z})_{(i+1)}}{d(\mathbf{z})_{(i)}}, \qquad i = 1,2,\ldots,B-2.
\end{equation}
Intuitively, $\tau_i(\mathbf{z})$ measures how rapidly the similarity neighborhood around $\mathbf{z}$ progresses.
If similar counterparts are plentiful, the successive distances grow slowly and $\tau_i$ remains small and stable. If the local neighborhood becomes sparse, $\tau_i$ increases.

Empirically, these log growth rates display an intriguing behavior.  
Across datasets and abstraction levels, $\tau_i$ decays with rank $i$ according to a power law:
\[
\tau_i \propto i^{-\alpha}, \qquad \alpha>0.
\]
A \emph{power law} represents a scale-free relation between two quantities, $y \propto x^{-\alpha}$, whose logarithmic form,
\[
\log y = -\alpha \log x + \text{const},
\]
appears as a straight line on a log–log plot with slope $-\alpha$.  
Such scaling laws are ubiquitous in nature and society—ranging from Zipf’s law \cite{zipf2016human} for word frequencies to the city-size distribution where populations scale inversely with rank \cite{vitanov2015test}. 
In our case, the observed scaling reflects how distances expand with neighbor rank similar to Zipf's law.

\subsubsection*{Empirical Evidence}
To empirically validate the power law of $\tau_i$, we conducted experiments on four representative datasets: MVTec \texttt{cable} (highly aligned and structurally consistent), MVTec \texttt{screw} (pose variation), MVTec LOCO \texttt{breakfast\_box} (logical variation), and brain MRI slices (medical domain). For each dataset, only normal images were used. We computed $\tau_i(\mathbf{z})$ for all patch tokens and averaged across query elements:
\[
\bar{\tau}_i = \mathbb{E}_{\mathbf{z}}[\tau_i(\mathbf{z})].
\]
Plots of $\bar{\tau}_i$ in log-log scale strongly suggest linearity across scales for both low-level (6th-layer) and high-level (24th-layer) features of ViT-L/14-336 (Fig.~\ref{fig:sim_scaling_vit_all}) and Dinov2-L/14 (Fig.~\ref{fig:sim_scaling_dino_all}).  
Ordinary least-squares fits in log–log space yielded exponents $\alpha \approx 0.8$ and coefficients of determination $R^2 \ge 0.98$ across all datasets, confirming that the growth rate follows a consistent power-law decay:
\[
\bar{\tau}_i \propto i^{-\alpha}.
\]
Inspired by this result, we propose the following hypothesis.
\begin{hypothesis}[Similarity Scaling Law]
For normal tokens embedded in the ViT feature space and compared against a sufficiently rich base set, the log growth rate of the mutual similarity vector follows a power-law decay,
\[
\tau_i \propto i^{-\alpha},
\]
and this behavior persists across datasets, modalities, and feature-abstraction levels.
\end{hypothesis}
\subsection{Empirical Evidence of the Neighbor--Burnout Phenomenon}
\label{sec:neighbor-burnout}

This subsection presents the \emph{Neighbor--Burnout} phenomenon, which manifests as distinct spikes in the mutual similarity vector $\mathcal{D}(\mathbf{z}_a)$ of consistent anomalies. Neighbor burnout arises as a direct consequence of the \emph{consistent--anomaly problem}—that is, the violation of the Rarity assumption—combined with the \emph{Similarity Scaling} phenomenon described in Section~\ref{sec:similarity-scaling}.

Consider an anomalous element $\mathbf{z}_a$ that is an $H$-level $\epsilon$-consistent anomaly with $\epsilon \ll \epsilon_0$ and \[|\mathcal{N}(\mathbf{z}_a,\epsilon)| = |\mathcal{N}(\mathbf{z}_a,\epsilon_0)| = H.\] 
This condition creates a \emph{sharp semantic cliff}: the first \(H\) nearest collections contain nearly identical duplicates within the tight semantic radius \(\epsilon\), whereas the \((H\!+\!1)\)-th collection jumps beyond the threshold \(\epsilon_0\). Consequently,
\[
d(\mathbf{z}_a)_{(i)} < \epsilon \text{ for } i \le H,\quad \text{and}\quad d(\mathbf{z}_a)_{(H+1)} \ge \epsilon_0.
\]
Up to index \(H\), the scaling behavior of \(\tau_i(\mathbf{z}_a)\) is indistinguishable from that of normal elements; beyond this index, the jump
\[
    \tau_H(\mathbf{z}_a) = \ln \frac{d(\mathbf{z}_a)_{(H+1)}}{d(\mathbf{z}_a)_{(H)}}
\]
becomes anomalously large due to the ``neighbor-burnout" of similar anomalous neighbors. This local, index-specific violation of the power-law regime is referred to as Neighbor--Burnout phenomenon, and $H$ in this case is refered as the \emph{burnout point} of the consistent anomaly $\mathbf{z}_a$.
\subsubsection*{Empirical Evidence}
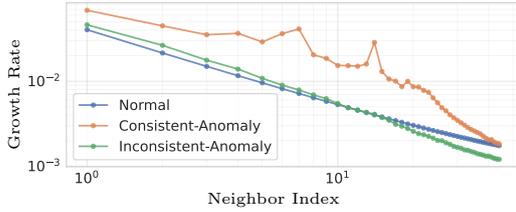
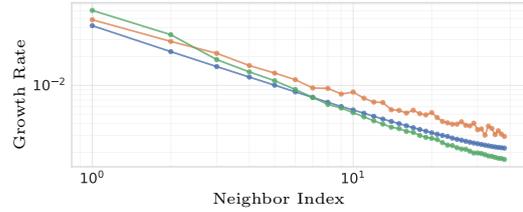
\begin{figure}[t]
  \centering
  \begin{subfigure}{0.45\textwidth}
    \centering
    \begin{tikzpicture}
      \node[inner sep=0pt] (img) {\includegraphics[width=\textwidth]{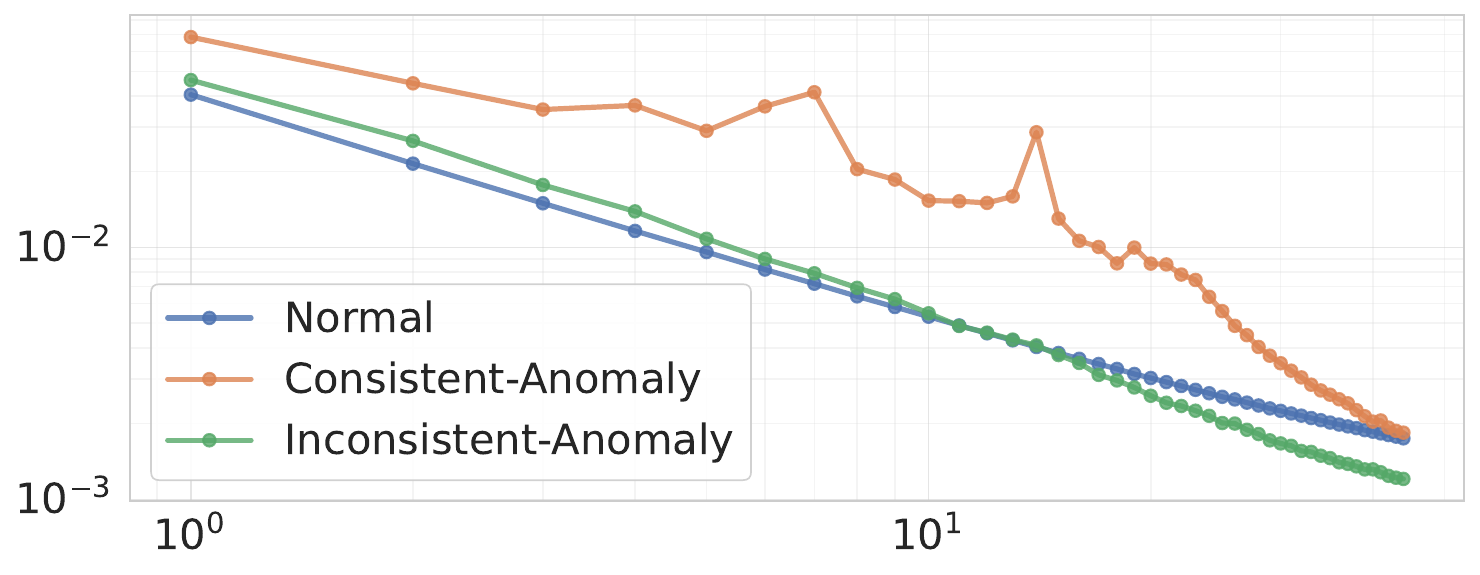}};
      \node[rotate=90, anchor=center] at ([xshift=-0.1cm]img.west) {\tiny Growth Rate};
\node[rotate=0, anchor=center] at ([yshift=-0.1cm]img.south) {\tiny Neighbor Index};
    \end{tikzpicture}
    \caption{\texttt{Cable}: Avg. Growth Rate}
    \label{fig:cable_avg}
  \end{subfigure}
  \hspace{1.1cm}
  %\hfill
\begin{subfigure}{0.45\textwidth}
    \centering
    \begin{tikzpicture}
      \node[inner sep=0pt] (img) {\includegraphics[width=\textwidth]{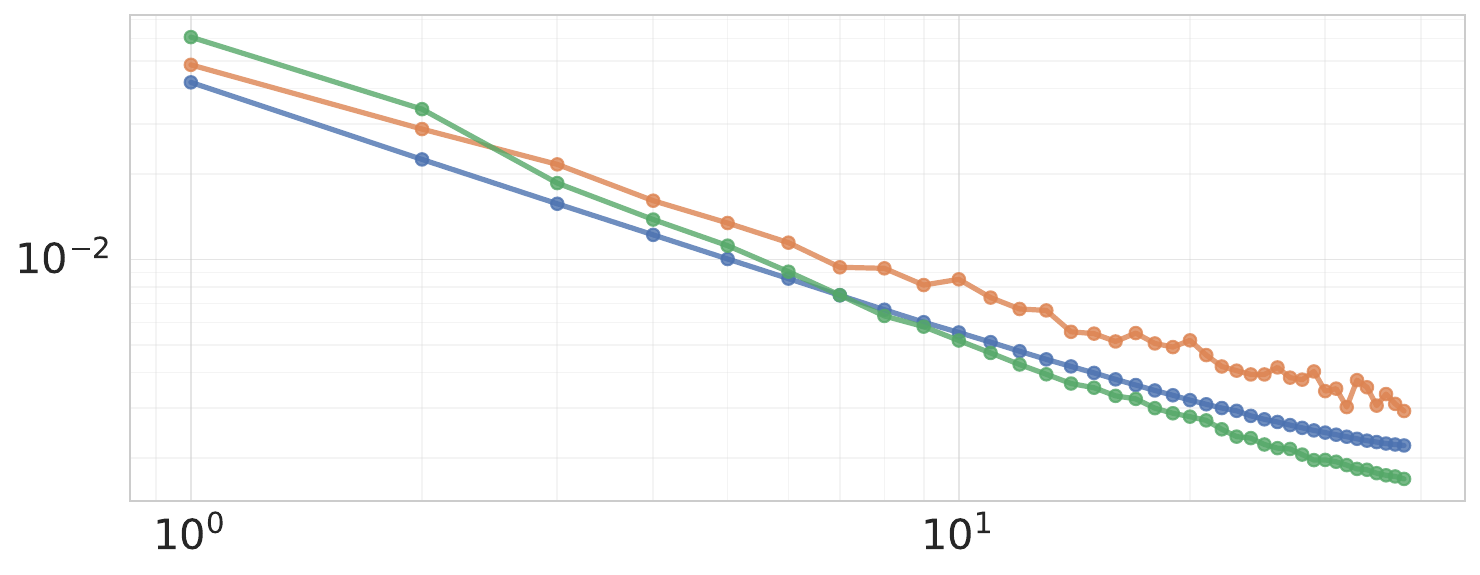}};
      \node[rotate=90, anchor=center] at ([xshift=-0.1cm]img.west) {\tiny Growth Rate};
      \node[rotate=0, anchor=center] at ([yshift=-0.1cm]img.south) {\tiny Neighbor Index};
    \end{tikzpicture}
    \caption{\texttt{Capsule}: Avg. Growth Rate}
    \label{fig:capsule_avg}
  \end{subfigure}
    \caption{Log-log plots of \(\tau^{(i)}(x)\). While normal patches and inconsistent anomalies in the \texttt{Cable} of MVTec AD show power-law decay in the growth rate, consistent anomalies show neighbor-burnout with a sudden rise (orange) in \(\tau^{(i)}(x)\) after exhausting similar matches. All patches in \texttt{Capsule}, which has a minimum presence of consistent anomalies, exhibit power-law decay in the growth.}
  \label{fig:neighbor_burnout}
\end{figure}
To quantify this effect, anomalous patches were grouped using their anomaly scores, defined in Eq.~\eqref{equation:anomaly_score} with the base set $\mathcal{B}$. Anomalous patches are labeled as consistent anomalies if their anomaly score lies below the 80th percentile of normal patch scores, which indicates deceptive matches. The remaining anomalous patches are labeled as inconsistent anomalies.\footnote{This categorization provides a reproducible criterion even for datasets without visually obvious consistent anomalies.} 

The experiments were conducted on both datasets containing repeated anomaly types and those without them. For datasets such as \texttt{cable} from MVTec AD, which exhibit repeated defects, consistent anomalies show a clear deviation from the power-law trend: their mutual similarity vectors exhibit a sudden rise in the averaged growth rate $\tau^{(i)}(x)$ once similar matches are exhausted. In contrast, normal patches and inconsistent anomalies maintain a power-law decay, consistent with the Similarity Scaling phenomenon. Control experiments on datasets without consistent anomalies, such as \texttt{capsule} from MVTec AD, show no such deviation—their growth rates follow smooth power-law decay across all indices. These results confirm that the neighbor--burnout pattern appears exclusively when consistent anomalies are present, and therefore directly reflects the existence of them in the base set $\mathcal{B}$.

The consistent--anomaly problem itself is global, since it arises from the recurrence of similar anomalies across multiple collections. Yet, the neighbor--burnout phenomenon reveals that this global relationship is recorded at the local level within each mutual similarity vector. The spike in $\tau_i(\mathbf{z}_a)$ around $i \approx H$ indicates the exhaustion of a finite pool of near-duplicates, providing a local signature of the consistent anomaly. Moreover, the prefix distances $d(\mathbf{z}_a)_{(i)}$ for $i \le H$ encode inter-collection relationships, showing that the collection containing $\mathbf{z}_a$ potentially shares similar anomalous patches with its first $H$ nearest collections. In this sense, the mutual similarity vector serves as a local record of both local and global information of the dataset $\mathcal{B}$. These observations motivate the method developed in the next chapter.
% ------------------------------------------------------
\chapter{CoDeGraph: Graph-based Framework for Consistent Anomalies}
\label{chap:algorithm}
In this chapter, CoDeGraph, a graph-based framework, is presented to address the problem of consistent anomalies in zero-shot anomaly detection. Portions of this chapter, including experimental setup, results, and figures, are adapted from~\cite{CoDeGraph} with additional analysis, context, and extensions tailored to the broader scope of this thesis.
\medskip

The chapter is organized as follows.  
Section~\ref{sec:motivation-overview} describes the motivation and provides an overview of the CoDeGraph framework.  
Section~\ref{sec:codegraph-framework} introduces the complete method, explaining how consistent anomalies are detected and filtered through a graph-based representation.  
Section~\ref{sec:experiment} presents extensive experiments demonstrating the effectiveness of CoDeGraph on standard industrial datasets, together with analyses of key components and practical considerations.
\section{Motivation and Overview}
\label{sec:motivation-overview}

The consistent anomaly problem, as introduced in Chapter~\ref{chap:empirical}, is not a rare case but a realistic challenge in industrial image anomaly detection, where the same defect repeat across many images due to systematic production errors or recurring environmental factors. This repetition of anomalous patches undermines the effectiveness of batch-based zero-shot methods, such as MuSc~\citep{MuSc}, which rely on the rarity and irregularity of anomalous patterns to boost their scores through top-$K$ aggregation (Eq.~\eqref{equation:anomaly_score}). A robust solution for the consistent anomalies problem is therefore required. 

\begin{figure}[t!]
  \vspace*{-1cm}
  \centering
  \includegraphics[width=0.8\textwidth]{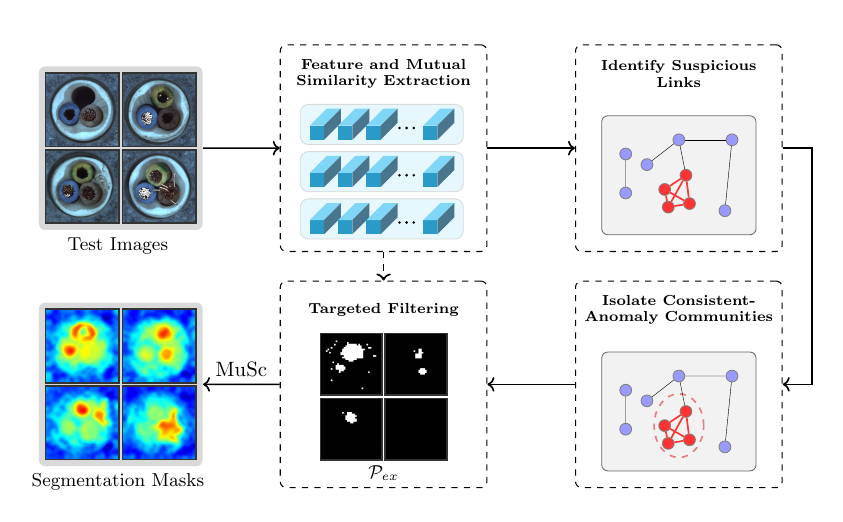}
  \caption{Overview of \emph{CoDeGraph}.}
  \label{fig:overview}
\end{figure}

To address this, our proposed method, Consistent-Anomaly Detection Graph (CoDeGraph), adopts a graph-based approach that leverages the global structure revealed by the neighbor-burnout phenomenon. CoDeGraph is designed as a three-stage pipeline:
\begin{enumerate}
    \item \textbf{Stage 1 --- Local detection of suspicious links:} Identify links potentially connecting consistent anomalous elements by utilizing the neighbor-burnout phenomenon.
    \item \textbf{Stage 2 --- Image-level anomaly similarity graph:} Aggregate the identified links in stage 1 into a weighted collection-level graph, where edge weights reflect the frequency of shared links, forming exceptionally dense communities among collections with repeated anomalies.
    \item \textbf{Stage 3 --- Community-based refinement:} Isolate these exceptionally dense communities as outliers, remove score-dependent elements from the base set $\mathcal{B}$ to get $\mathcal{B}_{\text{refined}}$, and recompute anomaly scores on $\mathcal{B}_{\text{refined}}$.
\end{enumerate}
Our design adheres to three key principles: (i) it is training-free, requiring no fine-tuning on any auxiliary dataset; (ii) it involves minimal intervention, removing only patches clearly associated with dense communities while preserving the normal structure of the base set when consistent anomalies are absent; and (iii) it is backbone- and modality-agnostic, since all steps depend solely on distances and graph structure, making the method applicable to any ViT backbone and any modalities with such properties describe in Chapter~\ref{chap:empirical}. 
The overall diagram of CoDeGraph pipeline is given in Figure~\ref{fig:overview}.
\section{CoDeGraph Framework}
\label{sec:codegraph-framework}
Throughout this section the base set is denoted by $\mathcal{B}=\{C_1,\dots,C_B\}$ as in Chapter~\ref{chap:empirical}, each collection $C_i$ is a set of element-level features $C_i=\{\mathbf{z}_i^1,\dots,\mathbf{z}_i^N\}$, and distances between an element and a collection are defined as in~\eqref{eq:dist-to-collection}. We also retain the log growth rate
\(
    \tau_i(\mathbf{z}) = \ln \dfrac{d(\mathbf{z})_{(i+1)}}{d(\mathbf{z})_{(i)}}.
\)

\subsection{Stage 1 – Local Detection of Suspicious Links}
\label{subsec:stage1-local}

The ultimate goal of CoDeGraph is to construct a graph in which collections that share similar anomalous patterns form dense and distinguishable communities.  
Achieving this requires identifying links that connect two elements whose similarity arises from the repetition of the same anomaly, while avoiding links that merely reflect normal similarity.  
The neighbor-burnout phenomenon provides a clear statistical cue for this distinction.  
For an $H$-level consistent anomaly $\mathbf{z}_a$, the mutual similarity vector initially follows the Similarity Scaling phenomenon $\tau_i(\mathbf{z}_a)\!\propto\! i^{-\alpha}$ but abruptly deviates at the burnout index $H$.  
Distances $d(\mathbf{z}_a)_{(i)}$ for $i<H$ correspond exactly to those “anomaly-to-anomaly’’ relations that persist before burnout—these are the links we seek to capture.  
To isolate them in an unsupervised way, we introduce the concept of an \emph{endurance ratio}.

\paragraph{Endurance ratio.}
A reference rank $\omega$ is fixed beyond the burnout region ($H<\omega$), and for each element $\mathbf{z}$ we define
\begin{equation}
    \label{eq:endurance-ratio-thesis}
    \zeta(\mathbf{z}, C_{(i)}) = \zeta(\mathbf{z})_{(i)}
    \;:=\;
    \frac{d(\mathbf{z})_{(i)}}{d(\mathbf{z})_{(\omega)}},
    \qquad 1 \le i < \omega .
\end{equation}
The endurance ratio measures how long the mutual similarity vector $\mathcal{D}_{\mathcal{B}}(\mathbf{z})$ can \emph{endure} the burnout of its nearest neighbors before reaching the semantic cliff where distances grow sharply.  
A large $\zeta(\mathbf{z})_{(i)}$ indicates that neighbor distances increase gradually and the similarity structure endures smoothly, as expected for normal elements.  
A small $\zeta(\mathbf{z})_{(i)}$ implies that the early neighbors are extremely close in comparison to the distant ones—signaling a short endurance and revealing the presence of near-duplicate anomalies that rapidly exhaust local similarity.

In practice, $\omega$ is chosen as a fraction (typically 30–70\%) of the base-set size $B$ so that it remains beyond typical burnout points $H$ while remaining within the region where the Doppelgänger assumption holds.

\begin{figure}[t]
  \centering
  \begin{subfigure}{0.43\textwidth}
    \centering
    \begin{tikzpicture}
      \node[inner sep=0pt] (img) {\includegraphics[width=\textwidth]{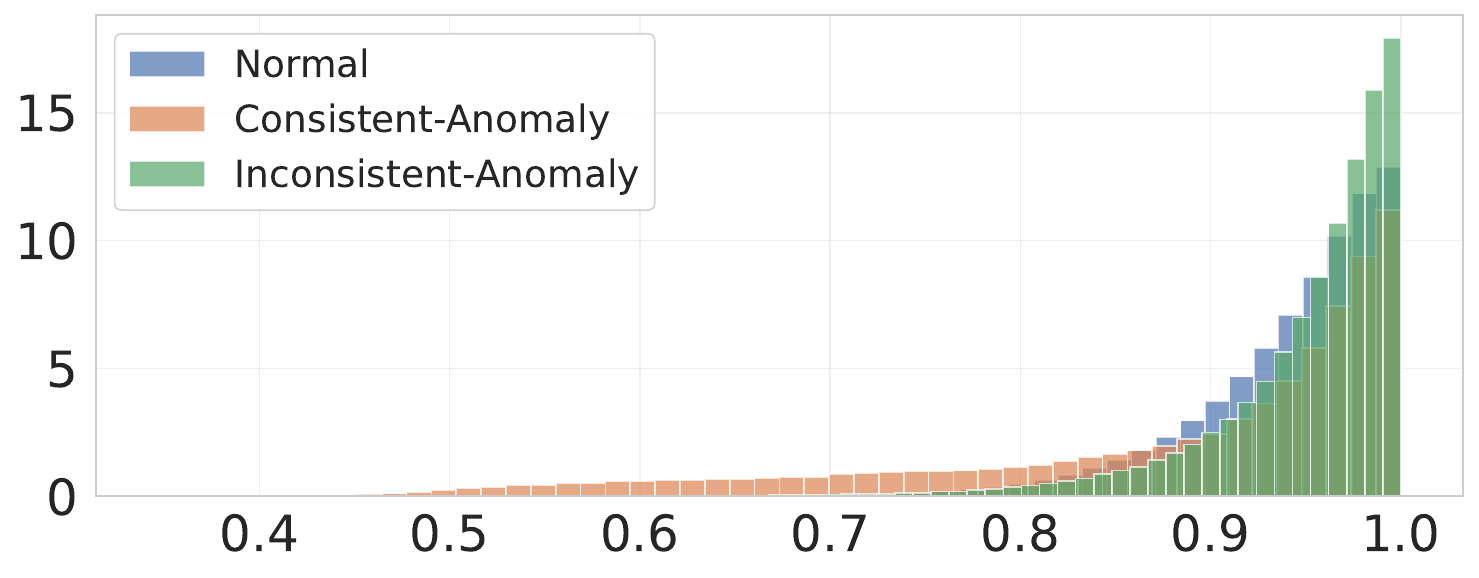}};
      \node[rotate=90, anchor=center] at ([xshift=-0.1cm]img.west) {\tiny Density};
    \end{tikzpicture}
    \caption{Distribution of endurance ratios $\zeta(x, C_{(i)})$}
    \label{fig:hist_endurance_ratio}
  \end{subfigure}
  \hspace{5mm}
  \begin{subfigure}{0.43\textwidth}
    \centering
        \begin{tikzpicture}
      \node[inner sep=0pt] (img) {\includegraphics[width=\textwidth]{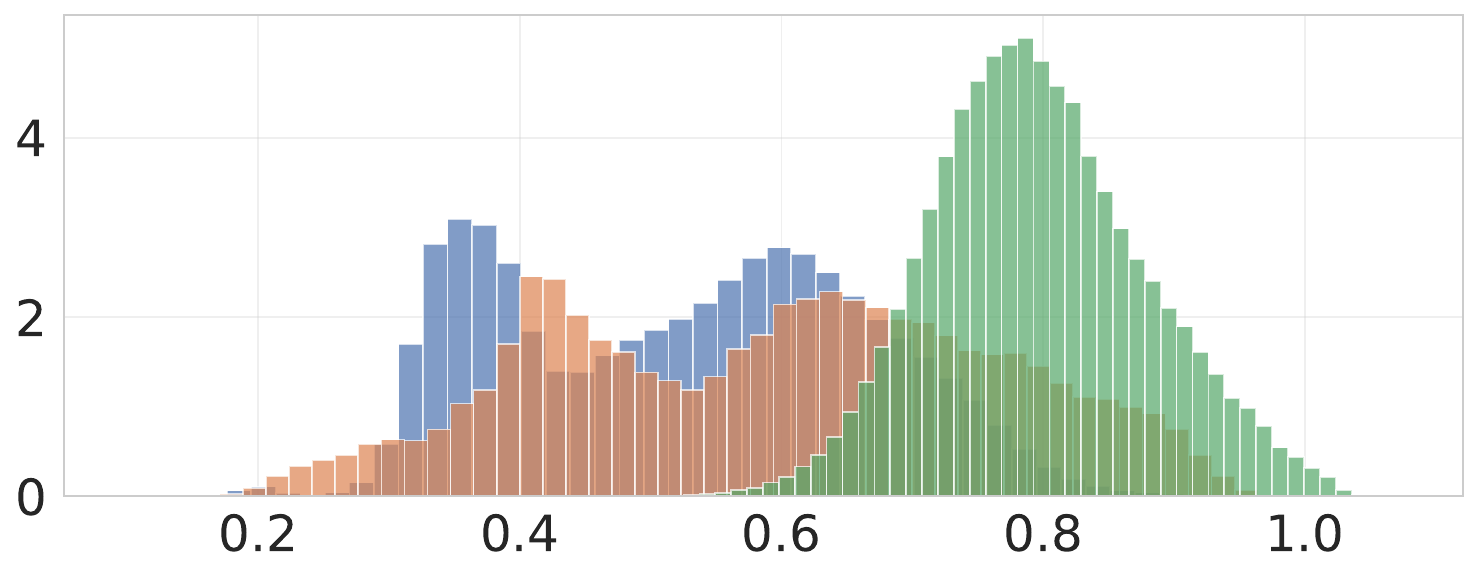}};
      \node[rotate=90, anchor=center] at ([xshift=-0.1cm]img.west) {\tiny Density};
    \end{tikzpicture}
    \caption{Distribution of absolute distances $d(x, C_{(i)})$}
    \label{fig:hist_abs_distance}
  \end{subfigure}
  \caption{Distributions of \( \zeta(x, C_{(i)}) \) and \( d(x, C_{(i)}) \) for \( i<\omega \) across patch types on \texttt{cable}. (a) The endurance ratio provides clear domination at the tail, with consistent-anomaly patches exhibiting significantly lower \( \zeta \), enabling robust identification of suspicious links. (b) Absolute distances \( d(x, C_{i}) \) show overlapping distributions between normal patches and consistent anomalies, making discrimination challenging.}
  \label{fig:distance_comparison}
\end{figure}
\paragraph{Suspicious links.}
Extremely small endurance ratios thus indicate potential consistent-anomaly connections between collections.  
For each element $\mathbf{z}$ and its $i$-th nearest collection $C_{(i)}$, the pair $(\mathbf{z}, C_{(i)})$ is treated as a \emph{link}.  
Let
\[
    L := \bigl\{\, (\mathbf{z}, C_j) \;\big|\; \mathbf{z}\!\in\! C_\ell,\; j\!\ne\!\ell \,\bigr\}
\]
be the set of all possible links in $\mathcal{B}$.  
The subset of \emph{suspicious links}—those most likely connecting repeated anomalies—is then defined as
\begin{equation}
    \label{eq:suspicious-links}
    \mathcal{S}_l
    \;=\;
    \bigl\{\, (\mathbf{z}, C_j)\!\in\! L
        \;\big|\;
        \zeta(\mathbf{z}, C_j)\!\le\!\lambda
    \,\bigr\},
\end{equation}
where $\lambda>0$ is adaptively increased until the graph constructed in Stage 2 reaches the desired coverage.  
Links in $\mathcal{S}_l$ represent deviations from the expected growth of normal data and form the essential building blocks for detecting communities of consistent anomalies in later stages.
\subsection{Stage 2 – Anomaly Similarity Graph Construction}
\label{subsec:stage2-graph}

To reveal the structure formed by suspicous links that connect at element-level, a collection-level graph
\[
    \mathcal{G}=(V,E), \qquad V=\{1,\dots,B\},
\]
is constructed. Each node $i\in V$ represents a collection $C_i$.  
An (undirected) edge $(i,j)$ is created if at least one suspicious link in $\mathcal{S}_l$ connects an element from $C_i$ to $C_j$ (or vice versa).  
Its weight records how many such links exist:
\begin{equation}
    \label{eq:edge-weight}
    w_{ij} \;=\;
        \bigl|\{\, (\mathbf{z}, C_j)\in\mathcal{S}_l \mid \mathbf{z}\in C_i \,\}\bigr|
        + \bigl|\{\, (\mathbf{z}, C_i)\in\mathcal{S}_l \mid \mathbf{z}\in C_j \,\}\bigr| .
\end{equation}
Collections that share the same anomaly type will thus have many cross-links with large $w_{ij}$, while normal images or randomly anomalous ones are linked only sparsely.
\paragraph{Weighted Endurance Ratio.}
While normal patches are generally scattered across the graph $\mathcal{G}$, their abundance can blur the distinctiveness of consistent-anomaly communities.  
This problem is particularly evident in datasets with diverse normal appearances, such as \texttt{breakfast\_box} from MVTec LOCO, where the variety of normal configurations causes many normal patches to exhibit unstable growth rates \( \tau^{(i)} \).  
As a result, these normal patches may inject many small endurance ratios $\zeta$, leading to an excessive number of normal links that weaken the anomaly-specific structure of the graph.  
To alleviate this, CoDeGraph adopts a \emph{weighted endurance ratio} defined as
\begin{equation}
    \label{eq:weighted-endurance-ratio}
    \zeta'(\mathbf{z})_{(i)} \;=\;
        \zeta(\mathbf{z})_{(i)} \, d(\mathbf{z})_{(i)}^{-\alpha}
        \;=\;
        \frac{d(\mathbf{z})_{(i)}^{1-\alpha}}{d(\mathbf{z})_{(\omega)}},
\end{equation}
where $\alpha>0$ is a small weighting parameter.  
The inverse-distance term $d(\mathbf{z})_{(i)}^{-\alpha}$ assigns higher importance to links involving larger absolute distances, which are more likely to arise from anomalous patterns, thus reduce the appearance of normal links in $\mathcal{S}_{\ell}$.   
Furthermore, since removing normal patches from the base set $\mathcal{B}$ tends to degrade MSM performance more severely than excluding anomalous ones, weighting $\zeta$ in this way biases filtration toward anomalous connections—retaining the normal elements within $\mathcal{B}$ and hence maintaining the core strength of MSM.
\paragraph{Coverage-based link selection.}
If $\mathcal{S}_\ell$ in~\eqref{eq:suspicious-links} is chosen too strictly (very small $\lambda$) or too loosely (very large $\lambda$), the resulting graph $\mathcal{G}$ may become either too sparse or too dense for consistent-anomaly communities to stand out.  
To ensure a balanced graph, we do not fix $\lambda$ in advance.  
Instead, we first sort all candidate links $L = \{(\mathbf{z}, C_j)\}$ in ascending order of the weighted endurance ratio $\zeta'(\mathbf{z}, C_j)$ and then \emph{incrementally} add links to the suspicious-link set $\mathcal{S}_\ell$ in small batches.  
After each addition, we evaluate the current graph coverage
\[
    \mathrm{cov}(\mathcal{G}) \;=\;
    \frac{|\{\, i \in V \mid \deg(i) > 0 \,\}|}{B},
\]
where $\deg(i)$ is the degree of node $i$ in the graph built from $\mathcal{S}_\ell$. Links are added until $\mathrm{cov}(\mathcal{G})$ reaches a target value (typically $0.9$–$0.95$ for potential outlier $C_i$). This guarantees that most collections remain represented in the graph, enabling reliable identification of dense communities in the next stage.  
The procedure is summarized in Algorithm~\ref{alg:coverage-selection-thesis}.

At the end of this stage, $\mathcal{G}=(V,E)$ becomes a weighted, nearly connected graph in which:
\begin{itemize}
    \item normal images have low degree and small average edge weight;
    \item images sharing the same consistent anomaly are tightly connected through numerous low-$\zeta'$ links, forming visibly dense subgraphs.
\end{itemize}
This \emph{Anomaly Similarity Graph} serves as the input to Stage~3, where community detection and targeted filtering are performed.

\begin{algorithm}[H]
\caption{Coverage-based Selection of Suspicious Links}
\label{alg:coverage-selection-thesis}
\begin{algorithmic}[1]
\Require Sorted list of candidate links $L = \{(\mathbf{z}, C_j)\}$, target coverage $\tau \in (0,1]$, number of collections $B$
\State Sort $L$ in ascending order of $\zeta'(\mathbf{z}, C_j)$
\State $\mathcal{S}_l \leftarrow \emptyset$ \Comment{selected suspicious links}
\State $k \leftarrow B(B-1)/2$ \Comment{initial batch size}
\Repeat
    \State Add the next $k$ links from $L$ to $\mathcal{S}_l$
    \State Construct temporary graph $\mathcal{G}$ from $\mathcal{S}_l$
    \State $c \leftarrow$ fraction of nodes in $\mathcal{G}$ with degree $\ge 1$
    \If{$c < \tau$}
        \State $k \leftarrow k + B(B-1)/2$
    \EndIf
\Until{$c \ge \tau$ \textbf{or} all links in $L$ are used}
\State \Return $\mathcal{S}_l$
\end{algorithmic}
\end{algorithm}

\subsection{Stage 3 – Community Detection and Filtering}
\label{subsec:stage3-filtering}
The final stage identifies groups of collections that share repeated anomalies and removes
the corresponding elements. Community detection is first applied to the anomaly simi-
larity graph $\mathcal{G}$, followed by a selective filtering process that remove elements that find no
similar counterparts outside of its own communities
\paragraph{Community detection via the Constant Potts Model}
To discover communities in $\mathcal{G}$, CoDeGraph adopts the Leiden algorithm~\citep{leiden} with the Constant Potts Model (CPM) objective~\citep{cpm},
\begin{equation}
    Q_{\mathrm{CPM}} = \sum_{i,j}(A_{ij} - \gamma)\,\delta(\sigma_i,\sigma_j),
\end{equation}
where $A_{ij}=w_{ij}$ is the weighted adjacency matrix, $\delta(\sigma_i,\sigma_j)=1$ if nodes $i$ and $j$ belong to the same community, and $\gamma$ controls the minimum internal density required for a valid community.  
In practice, $\gamma$ is set to the 25th percentile of edge weights to ensure that most intra-community connections among consistent anomalies exceed the resolution threshold.  
Let the detected communities be
\[
    \mathcal{M} = \{M_1, M_2, \dots, M_H\}, \qquad M_h \subseteq V .
\]

\begin{figure}[t]
  \centering
  \begin{minipage}{6.5cm}
    \centering
    \begin{subfigure}{\textwidth}
      \centering
      \begin{tikzpicture}
        \node[inner sep=0pt] (img) {\includegraphics[width=\textwidth]{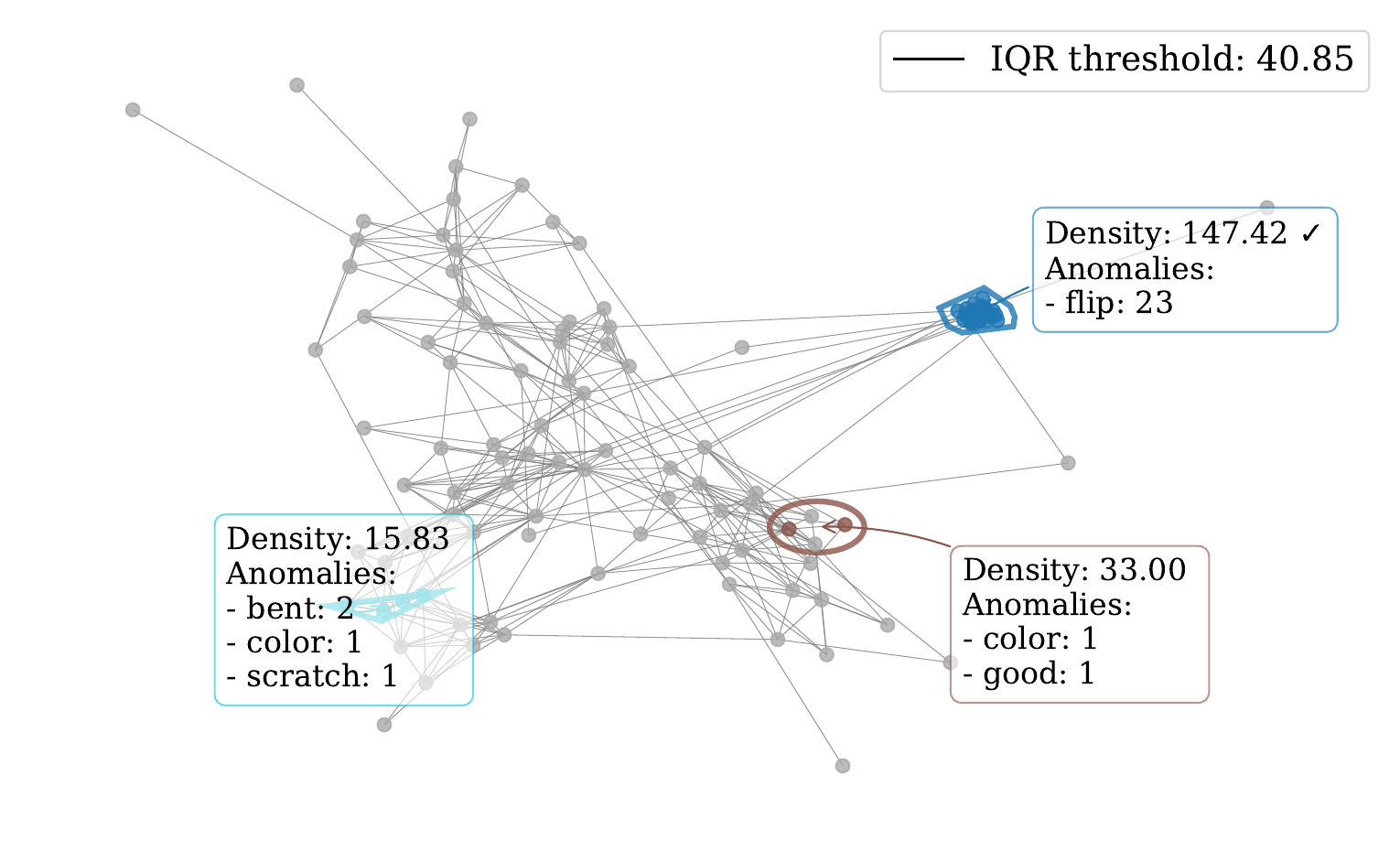}};%        \node[anchor=north east, xshift=-0.24cm, yshift=-1.06cm] at (img.north east) {\footnotesize\greencheck};
      \end{tikzpicture}
      \caption{\texttt{Metal\_nut}}
      \label{fig:metal_nut_communities}
    \end{subfigure}
  \end{minipage}
  \hfill
  \begin{minipage}{6.5cm}
    \centering
    \begin{subfigure}{\textwidth}
      \centering
      \includegraphics[width=\textwidth]{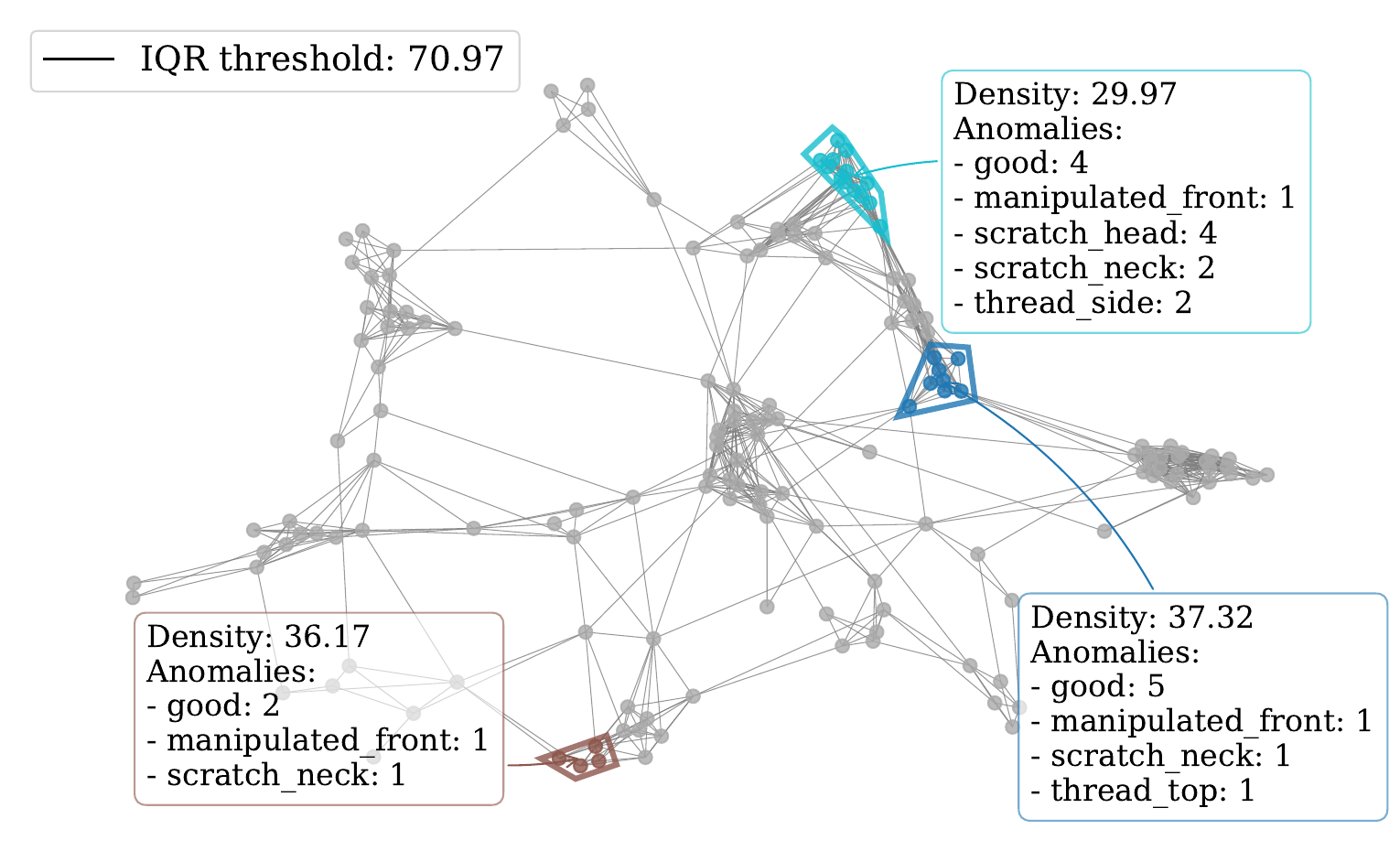}
      \caption{\texttt{Screw}}
      \label{fig:screw_communities}
    \end{subfigure}
  \end{minipage}
  \caption{Anomaly similarity graphs on MVTec AD subclasses showing top three communities by density. (a) \texttt{Metal\_Nut}: Community \#1 contains all 23 flipped metal nuts with exceptionally high density, exceeding the IQR threshold. (b) \texttt{Screw}: Nodes are clustered into distinct communities, but none exhibit exceptionally high density.}
  \label{fig:top_communities}
\end{figure}

\paragraph{Density and outlier detection.}
For each community $M\in\mathcal{M}$, its internal connection strength is quantified by
\begin{equation}
    \label{eq:community-density}
    \rho(M) = \frac{\sum_{i,j\in M,\,i\neq j} w_{ij}}{|M|(|M|-1)},
\end{equation}
which measures how densely the nodes in $M$ are linked through suspicious connections.  
The distribution $\{\rho(M)\}_{M\in\mathcal{M}}$ is then analyzed using Tukey’s inter-quartile range (IQR) rule.  
Writing $Q_1$ and $Q_3$ for the first and third quartiles and $IQR=Q_3-Q_1$, a community is flagged as an \emph{outlier community} if
\begin{equation}
    \label{eq:iqr-rule}
    \rho(M) > Q_3 + k_{\mathrm{IQR}}\!\cdot\! IQR,
\end{equation}
where a large $k_{\mathrm{IQR}}$ (e.g., $k_{\mathrm{IQR}}=4.5$) is used to ensure only extremely dense communities are marked, as visualized in Fig.~\ref{fig:top_communities}. Lower values such as $k_{\mathrm{IQR}}=1.5$ or $k_{\mathrm{IQR}}=3$ can be used in practice for broader sensitivity. The set of outlier communities is denoted by $\mathcal{S}_m \subseteq \mathcal{M}$.

\paragraph{Targeted patch filtering.}
Outlier communities indicate which collections contain repeated anomalies, but not which elements cause them.  
Removing all images in such a community would be excessive, as they may still contain many normal regions.  
Instead, CoDeGraph applies an element-level filtering procedure based on the \emph{dependency ratio}.

Let $\mathcal{B}$ be the current base set, and let $M\in\mathcal{S}_m$ be an outlier community. Consider temporarily removing all collections indexed by $M$ from the base set, obtaining $\mathcal{B}\setminus M$. For every element $\mathbf{z}$, compute its anomaly score \eqref{equation:anomaly_score} with respect to $\mathcal{B}$ and ${\mathcal{B}\setminus M}$. The \emph{dependency ratio} of $\mathbf{z}$ on community $M$ is given by

\begin{equation}
    \label{eq:dependency-ratio}
    r_M(\mathbf{z}) \;=\; \frac{a_{\mathcal{B}\setminus M}(\mathbf{z})}{a_{\mathcal{B}}(\mathbf{z})}.
\end{equation}
If $\mathbf{z}$ is a consistent-anomaly patch whose low score was caused mainly by matches inside $M$, then removing $M$ will make its score jump and $r_M(\mathbf{z})$ will be large. If $\mathbf{z}$ is a normal patch, the score will change little because many alternatives remain in $\mathcal{B}\setminus M$ and $r_M(\mathbf{z})$ stays close to $1$.

To avoid hand-tuning a global threshold, CoDeGraph compares $r_M(\mathbf{z})$ against the empirical distribution of ratios from patches \emph{outside} $M$, which give a good approximation to the normal patches inside $M$. Let $\theta_M$ be, for example, the $99$-th percentile of $\{ r_M(\mathbf{z}) \mid \mathbf{z} \notin M \}$; then the patches to be excluded due to $M$ are
\[
    \mathcal{P}_{\mathrm{ex}}(M)
    \;=\;
    \bigl\{\, \mathbf{z} \in M \;\big|\; r_M(\mathbf{z}) > \theta_M \,\bigr\}.
\]
Collecting over all outlier communities,
\[
    \mathcal{P}_{\mathrm{ex}} \;=\; \bigcup_{M\in\mathcal{S}_m} \mathcal{P}_{\mathrm{ex}}(M),
    \qquad
    \mathcal{B}_{\mathrm{refined}} \;=\; \mathcal{B} \setminus \mathcal{P}_{\mathrm{ex}} .
\]
The refined base set $\mathcal{B}_{\mathrm{refined}}$ preserves normal structures and even normal parts inside consistent-anomaly images, while removing those elements whose scores depended abnormally on a dense, anomaly-driven community. The targeted patch filtering is summarized in Algorithm~\ref{alg:targeted-filtering-thesis}.

\begin{remark}
    The inclusion of a patch \(\mathbf{z}_i^{h} \in \mathcal{P}_{\mathrm{ex}}\) (or its retention in \(\mathcal{B}_{\mathrm{refined}}\)) is determined not by the feature vector \(\mathbf{z}_i^{h}\) itself, but by its spatial position \(h\) in image \(i\), as defined by the ViT patchification process.
\end{remark}

\begin{algorithm}[H]
\caption{Targeted Patch Filtering from Outlier Communities}
\label{alg:targeted-filtering-thesis}
\begin{algorithmic}[1]
\Require Set of outlier communities $\mathcal{S}_m$, base set $\mathcal{B}$
\State $\mathcal{P}_{\mathrm{ex}} \leftarrow \emptyset$
\State Compute $a_{\mathcal{B}}(p)$ for all elements/patches $p$ in $\mathcal{B}$
\For{each community $M \in \mathcal{S}_m$}
    \State $\mathcal{B}_{\text{temp}} \leftarrow \mathcal{B} \setminus M$
    \State $R \leftarrow [\,]$ \Comment{to store ratios from outside $M$}
    \For{each patch $p$ in $\mathcal{B}$}
        \State Compute $a_{\mathcal{B}_{\text{temp}}}(p)$
        \State $r(p) \leftarrow a_{\mathcal{B}_{\text{temp}}}(p) / a_{\mathcal{B}}(p)$
        \If{$p \notin M$}
            \State append $r(p)$ to $R$
        \EndIf
    \EndFor
    \State $\theta \leftarrow$ 99th percentile of $R$
    \State $\mathcal{P}_{\mathrm{ex}} \leftarrow \mathcal{P}_{\mathrm{ex}} \cup \{\, p \in M \mid r(p) > \theta \,\}$
\EndFor
\State \Return $\mathcal{P}_{\mathrm{ex}}$
\end{algorithmic}
\end{algorithm}
\subsection{Final Anomaly Scoring}
\label{subsec:final-anomaly-score}
Up to this point, we have used a simplified notation for the patch features, such as \(\mathbf{z}_i^1\) or \(\mathbf{z}\), for generality without specifying the layer index \(l\) or receptive field size \(r\). In this subsection, we incorporate these details to provide a complete description of the final scoring process, which leverages multi-layer and multi-scale features~\cite{MuSc, patchcore, APRIL-GAN}  from the Vision Transformer (ViT).

\paragraph{Local Neighborhood Aggregation.}
CoDeGraph utilize \emph{Local Neighborhood Aggregation} (LNAMD)~\citep{MuSc,patchcore} to improve the ability to capture anomalies with different size. 
Each transformer layer \(l\) produces patch tokens \(\mathbf{z}_i^{h,l}\in\mathbb{R}^D\) that represent local visual features at spatial position \(h\).  
To capture the scale diversity of anomalies, LNAMD aggregates patch tokens within a local receptive field of size \(r \times r\):
\[
F_i^{l,r} = [\,p^{r}(\mathbf{z}_i^{1,l}), \ldots, p^{r}(\mathbf{z}_i^{N,l})\,],
\]
where \(p^{r}(\cdot)\) denotes an adaptive pooling operator over the receptive field \(r\).  
Small receptive fields emphasize fine-grained texture anomalies, while larger ones capture broad structural deformations. Note that with $p^{r=1}(\cdot)$ is the Identity function.

\paragraph{Multi-layer, multi-scale Mutual Scoring.}
Given the refined base \(\mathcal{B}_{\mathrm{refined}}\), the anomaly score of each patch token \(\mathbf{z}_i^{h,l}\)  
is computed independently for each layer \(l\) and receptive field \(r\) using the Mutual Scoring Mechanism:
\begin{equation}
    a_{\mathcal{B}_{\mathrm{refined}}}^{r}(\mathbf{z}_i^{h,l})
    = \frac{1}{K}\sum_{k=1}^{K} d^{r}\!\left(\mathbf{z}_i^{h,l}\right)_{(k)},
\end{equation}
where
\[
d^{r}(\mathbf{z}_i^{h,l}, C_{(k)}) = 
    \min_{t}\!\bigl\| p^{r}(\mathbf{z}_i^{h,l}) - p^{r}(\mathbf{z}_{(k)}^{t,l}) \bigr\|_2^2
\]
is the squared Euclidean distance to the \(k\)-th most similar collection in \(\mathcal{B}_{\mathrm{refined}}\).  
Since consistent-anomaly positions have already been removed, these computations are free from deceptive matches and can be conducted fully independent across layers and receptive fields.

\paragraph{Aggregation across layers and receptive fields.}
The final anomaly score at position \(h\) in image \(C_i\) is obtained by averaging over all considered layers \(l\) and receptive fields \(r\):
\begin{equation}
    \label{eq:final-anomaly}
    \mathcal{A}_{\mathcal{B}_{\mathrm{refined}}}(\mathbf{z}_i^{h})
    = \mathbb{E}_{l,r}\!\left[a_{\mathcal{B}_{\mathrm{refined}}}^{r}(\mathbf{z}_i^{h,l})\right].
\end{equation}
This cross-layer, cross-scale aggregation fuses complementary information: shallow layers and small receptive fields emphasize fine-grained texture details,  
while deeper layers and large receptive fields capture global structural context.  
The resulting \(\mathcal{A}_{\mathcal{B}_{\mathrm{refined}}}(\mathbf{z}_i^{h})\) provides a unified measure of anomaly likelihood for each spatial position, which is then interpolated to the original image resolution to form the final anomaly map. The collection-level anomaly score is given by max pooling over the anomaly map $\mathcal{A}_{\mathcal{B}_{\mathrm{refined}}}(\mathbf{z}_i^{h})
$.
\section{Experiments}
\label{sec:experiment}
This section presents the empirical evaluation of CoDeGraph. 

\subsection{Experimental Setup}

\paragraph{Datasets.} 
The experimental evaluation was carried out on two widely recognized benchmarks for industrial anomaly classification and segmentation: MVTec AD~\citep{mvtec} and VisA~\citep{zou2022spot}. 
The MVTec AD dataset comprises 15 industrial object categories. 
For the purpose of this study, it is partitioned into two groups according to the consistency characteristics of their anomalies. 
The first group, denoted as MVTec-CA (\texttt{cable}, \texttt{metal\_nut}, \texttt{pill}), contains classes that exhibit strong consistent anomalies, thereby violating the rarity assumption introduced in Assumption~\ref{assump:rarity}. 
The remaining categories form the MVTec-IA group, in which anomalies occur randomly. 
The VisA benchmark provides a complementary evaluation setting with more diverse and subtle defect types, making it suitable for assessing performance under inconsistent-anomaly conditions.

To compensate for the absence of public benchmarks specifically designed to evaluate consistent-anomaly scenarios, two additional datasets were introduced: {MVTec-SynCA} and {ConsistAD}. 
The {MVTec-SynCA} benchmark was constructed by applying controlled perturbations—such as lighting variations and camera displacements—to a single anomalous image within each MVTec AD subclass, thereby generating synthetic consistent anomalies. 
In contrast, {ConsistAD} aggregates consistent-anomaly subsets from MVTec AD, MVTec LOCO~\citep{mvtec-loco}, and MANTA~\citep{manta}, offering a broader empirical basis for cross-domain evaluation. 
Comprehensive dataset descriptions and statistics for both {MVTec-SynCA} and {ConsistAD} are provided in Appendix~\ref{sec:details}.

\paragraph{Implementation Details.}
For fair comparison with previous works, we used the ViT-L/14@336 backbone pre-trained by OpenAI~\citep{CLIP} for feature extraction.  
This ViT version comprises 24 layers organized into four stages of six layers each.  
Patch embeddings were extracted from layers 6, 12, 18, and 24, following the practice in~\citep{APRIL-GAN, MuSc}, while the \texttt{[CLS]} token from the final layer was used for AC optimization via RsCIN~\citep{MuSc}.  
All test images were resized to \(518\times518\) pixels.  
Although CLIP ViT-L/14@336 was used for fair comparison with prior works, additional experiments with different ViT such as DINO~\citep{DINO} and DINOv2~\citep{DINOv2} revealed that DINOv2-L/14 consistently achieved superior segmentation performance (89.9\% pixel-wise AUROC, 69.1\% pixel-wise F1, and 71.9\% pixel-wise AP).  

To aggregate information across multiple semantic levels, the layer-wise distances were averaged to get $d_{\mathrm{agg}}$ for construct the anomaly similarity graph:
\[
  d_{\mathrm{agg}}(\mathbf{z})_{(i)} = \frac{1}{4} \sum_{l \in \{6,12,18,24\}} d^{r=1}(\mathbf{z}^l)_{(i)} .
\]
Consequently, each element-to-collection link $(\mathbf{z}, C_j)$ in the suspicious-link set \(\mathcal{S}_l\) now corresponds to a link from an elemt to a set of four collections. Hence for each selected link in \(\mathcal{S}_l\) raise four connections.
The weighted endurance ratio used \(\alpha = 0.2\) and a reference index \(\omega = 0.3B\), while the target coverage in the coverage-based selection \ref{alg:coverage-selection-thesis} procedure was fixed at \(\tau = 0.95\). For computing the final anomaly scores, the MSM aggregated the lowest 10\% of distances in each mutual similarity vector \(\mathcal{D}_{\mathcal{B}}(\mathbf{z})\), rather than the 30\% fraction adopted in~\citep{MuSc}. In LNAMD, we select the receptive field size \( r \in \{1,3,5\} \).
\begin{remark}
All primary experiments were performed under these fixed hyperparameters without dataset-specific tuning unless specified.
\end{remark}

\paragraph{Evaluation Metrics.} For AC, we reported three metrics: Area Under the Receiver Operating Characteristic curve (AUROC), Average Precision (AP), and F1-score at the optimal threshold (F1). For AS, we reported four metrics: pixel-wise AUROC, pixel-wise F1, pixel-wise AP, and Area Under the Per-Region Overlap Curve (AUPRO). For the evaluation metrics related to consistent anomalies, we selected consistent anomalies as anomalous patches $\mathbf{z}_a$ with anomaly scores $a(\mathbf{z}_a)$ in the full base $\mathcal{B}$ (i.e., MuSc) below the 80th percentile of normal patch scores. Other anomalies were labeled as inconsistent-anomaly patches. This threshold uses ground-truth solely for post-hoc analysis and evaluation metrics, ensuring no impact on the zero-shot nature of CoDeGraph.

\paragraph{Baselines.} We compared CoDeGraph against SOTA zero-shot methods, including text-based approaches: WinCLIP \citep{WinCLIP}, APRIL-GAN \citep{APRIL-GAN}, AnomalyCLIP \citep{AnomalyCLIP}, and batch-based approaches: ACR \citep{ACR} and MuSc \citep{MuSc}. For MuSc, we reported two versions: MuSc 30\% (reported in \citep{MuSc}) and MuSc 10\%, where the percentage indicates the size of the Top-K in Eq.~\eqref{equation:anomaly_score}. Additionally, we compare with representative few-shot methods: Patchcore~\citep{patchcore}, WinCLIP \citep{WinCLIP}, APRIL-GAN\citep{APRIL-GAN} and GraphCore\citep{GraphCore}. When baseline metrics were unavailable, we reproduced results using official implementations.
\subsection{Main Results}

\begin{table}[t]
  \caption{Quantitative comparisons on \textbf{Consistent-Anomaly Datasets}. We compared CoDeGraph with state-of-the-art zero-shot methods. Bold indicates the best performance. All metrics are in \( \% \).}
\centering
\label{tab:consistent_results}
\resizebox{\textwidth}{!}{%
\begin{tabular}{l|l|l|ccc|cccc}
\toprule
\textbf{Dataset} & \textbf{Method} & \textbf{Setting} & \textbf{AUROC-cls} & \textbf{F1-cls} & \textbf{AP-cls} & \textbf{AUROC-seg} & \textbf{F1-seg} & \textbf{AP-seg} & \textbf{PRO-seg} \\
\midrule
\multirow{7}{*}{\thead{\normalfont{MVTec-CA}}}
& AnomalyCLIP & 0-shot  & 81.3 & 87.7 & 91.7 & 81.8 & 29.2 & 24.3 & 64.4 \\
& WinCLIP & 0-shot & 87.6 & 90.8 & 95.4 & 72.6 & 23.2 & - & 46.6 \\
& APRIL-GAN & 0-shot & 79.1 & 88.5 & 93.7 & 71.3 & 26.6 & 22.6 & 43.2 \\
& ACR & 0-shot & 73.3 & 88.1 & 88.1 & 86.9 & 42.7 & 35.5 & 66.0 \\
& MuSc \( 30\% \) & 0-shot & 97.2 & 97.0 & 99.3 & 93.3 & 58.9 & 58.4 & 92.8 \\
& MuSc \( 10\% \)& 0-shot  & 94.1 & 94.9 & 98.7 & 88.3 & 51.6 & 51.2 & 90.1 \\
& CoDeGraph & 0-shot  & \textbf{98.5 (\(\uparrow 1.3\))} & \textbf{97.8(\(\uparrow 0.8\))} & \textbf{99.6(\(\uparrow 0.3\))} & \textbf{98.1(\(\uparrow 4.8\))} & \textbf{73.8(\(\uparrow 14.9\))} & \textbf{77.2(\(\uparrow 18.8\))} & \textbf{95.4(\(\uparrow 2.6\))} \\
\midrule
\multirow{5}{*}{\thead{\normalfont{MVTec}-\\\normalfont{SynCA}}} 
& AnomalyCLIP& 0-shot  & 87.4 & 92.4 & 95.7 & 89.9 & 37.9 & 33.5 & 77.6 \\
& APRIL-GAN & 0-shot & 81.3 & 90.2 & 92.4 & 85.7 & 42.1 & 39.7 & 41.0 \\
& WinCLIP & 0-shot & 89.9 & 93.8 & 96.5 & 81.5 & 26.0 & 18.7 & 59.2 \\
& MuSc \( 10\% \) & 0-shot & 88.8 & 92.3 & 96.8 & 90.7 & 50.8 & 48.3 & 82.9 \\
& CoDeGraph  & 0-shot & \textbf{96.8(\(\uparrow 6.9\))} & \textbf{97.2(\(\uparrow 3.4\))} & \textbf{99.0(\(\uparrow 2.2\))} & \textbf{97.2(\(\uparrow 6.5\))} & \textbf{63.2(\(\uparrow 12.4\))} & \textbf{63.3(\(\uparrow 15.0\))} & \textbf{91.1(\(\uparrow 8.2\))} \\
\midrule
\multirow{5}{*}{ConsistAD} 
& AnomalyCLIP & 0-shot & 75.1 & 73.8 & 76.3 & 73.6 & 31.1 & 25.2 & 53.2 \\
& WinCLIP & 0-shot & 76.6 & 74.5 & 75.9 & 60.8 & 24.7 & 18.8 & 41.8 \\
& APRIL-GAN & 0-shot & 68.7 & 73.9 & 69.2 & 61.1 & 24.3 & 19.9 & 21.7 \\
& MuSc 10\% & 0-shot & 88.9 & 84.8 & 88.9 & 81.7 & 44.6 & 43.6 & 78.9 \\
& CoDeGraph  & 0-shot & \textbf{91.0(\(\uparrow 2.1\))} & \textbf{87.9(\(\uparrow 3.1\))} & \textbf{90.3(\(\uparrow 1.4\))} & \textbf{86.9(\(\uparrow 5.2\))} & \textbf{55.9(\(\uparrow 11.3\))} & \textbf{57.5(\(\uparrow 13.9\))} & \textbf{82.5(\(\uparrow 3.6\))} \\
\midrule
\multirow{5}{*}{\thead{\normalfont{MVTec-CA}}}
& PatchCore & 4-shot & 91.2 & 91.9 & 97.1 & 96.0 & 67.7 & 68.2 & 87.2 \\
& WinCLIP & 4-shot & 94.4 & 93.3 & 97.6 & 95.0 & 63.3 & - & 87.0 \\
& APRIL-GAN & 4-shot & 83.9 & 87.8 & 93.8 & 93.5 & 54.7 & 49.0 & 90.8 \\
& GraphCore & 4-shot & 93.2 & - & - & 97.5 & - & - & - \\
& CoDeGraph  & 0-shot & \textbf{98.5(\(\uparrow 4.1\))} & \textbf{97.8(\(\uparrow 4.5\))} & \textbf{99.6(\(\uparrow 2.0\))} & \textbf{98.1(\(\uparrow 0.6\))} & \textbf{73.8(\(\uparrow 6.1\))} & \textbf{77.2(\(\uparrow 9.0\))} & \textbf{95.4(\(\uparrow 4.6\))} \\
\bottomrule
\end{tabular}
}
\end{table}
\paragraph{Consistent-Anomaly Datasets.}
Table~\ref{tab:consistent_results} summarizes the results on datasets containing consistent anomalies.  
On MVTec-CA, CoDeGraph achieved an image-level AUROC of 98.5\%, outperforming MuSc by 1.3\% and reaching performance comparable to full-shot models\footnote{PatchCore-1 reported a 98.6\% AUROC-cls under full supervision.}.  
In segmentation, CoDeGraph demonstrated substantial improvements, yielding a 14.9\% increase in F1 and an 18.8\% rise in AP relative to the strongest zero-shot baseline.  
These gains emphasize the difficulty conventional zero-shot approaches face when encountering consistent anomalies, where violations of Assumption~\ref{assump:rarity} inflate the normality scores of repeated anomalous patterns.

Further validation on MVTec-SynCA and ConsistAD—datasets explicitly constructed for consistent anomalies—confirmed this trend.  
CoDeGraph consistently surpassed competing methods in both classification and segmentation, with gains exceeding 10\% in F1 and AP.  
This advantage originates from its capacity to identify dense communities within the anomaly similarity graph $\mathcal{G}$, exploiting the Neighbor-Burnout phenomenon to selectively remove consistent anomalies from the base set $\mathcal{B}$.

\paragraph{Inconsistent-Anomaly Datasets.}
On MVTec-IA and Visa, CoDeGraph outperformed the text-based zero-shot methods and achieved results on par with MuSc, as presented in Table~\ref{tab:inconsistent_results}.  
The near-identical performance of the two models reflects the absence of consistent anomalies in these datasets, preventing the formation of dense communities in $\mathcal{G}$. This led to minimal exclusions from $\mathcal{B}$, which had negligible impact on overall performance.  

% This is place of table tab:inconsistent_results
\begin{table}[t]
  \caption{Quantitative comparisons on the \textbf{Inconsistent-Anomaly Dataset}. We compared CoDeGraph with state-of-the-art zero-shot and few-shot methods. Bold indicates the best performance. All metrics are in \( \% \).}
\centering
\label{tab:inconsistent_results}
\resizebox{\textwidth}{!}{%
\begin{tabular}{l|l|l|ccc|cccc}
\toprule
\textbf{Dataset} & \textbf{Method} & \textbf{Setting} & \textbf{AUROC-cls} & \textbf{F1-cls} & \textbf{AP-cls} & \textbf{AUROC-seg} & \textbf{F1-seg} & \textbf{AP-seg} & \textbf{PRO-seg} \\
\midrule
\multirow{7}{*}{\thead{\normalfont{MVTec-IA}}} 
& AnomalyCLIP & 0-shot  & 94.1 & 94.0 & 97.5 & 93.4 & 41.6 & 37.1 & 83.1 \\
& WinCLIP  & 0-shot & 92.8 & 93.4 & 96.8 & 88.2 & 33.8 & - & 69.1 \\
& APRIL-GAN  & 0-shot & 87.9 & 90.9 & 93.5 & 91.7 & 47.5 & 45.4 & 44.3 \\
& ACR  & 0-shot & 88.2 & 92.8 & 94.3 & 93.8 & 44.7 & 40.1 & 74.2 \\
& MuSc \( 30\% \) & 0-shot  & 98.0 & \textbf{97.6} & 99.0 & \textbf{98.2} & 63.7 & 63.8 & 94.0 \\
& MuSc \( 10\% \)  & 0-shot & \textbf{98.3} & 97.3 & \textbf{99.1} & \textbf{98.2} & 64.9 & 65.7 & 94.3 \\
& CoDeGraph (Ours)  & 0-shot & \textbf{98.3} & 97.3 & \textbf{99.1} & \textbf{98.2} & \textbf{65.1} & \textbf{65.8} & \textbf{94.4} \\
\midrule
\multirow{4}{*}{Visa}  & WinCLIP & 0-shot & 78.1 & 79.0 & 81.2 & 79.6 &14.8 & - & 56.8 \\
& APRIL-GAN & 0-shot  & 78.0 & 78.7 & 81.4 & 94.2 & 32.3 & 25.7 & 86.8 \\
& MuSc \( 10\% \) & 0-shot  &\textbf{91.6} & \textbf{89.1} & \textbf{92.2} & \textbf{98.7} & \textbf{48.3} & \textbf{45.4} & \textbf{91.4} \\
& CoDeGraph (Ours) & 0-shot  &  \textbf{91.6} & 89.0 & \textbf{92.2} & \textbf{98.7} & \textbf{48.3} & \textbf{45.4} & \textbf{91.4} \\
\midrule
\multirow{5}{*}{\thead{\normalfont{MVTec-IA}}} 
& PatchCore & 4-shot & 90.3 & 94.7 & 95.3 & 94.1 & 47.5 & 44.3 & 83.0 \\
& WinCLIP & 4-shot & 95.4 & 95.1 & 97.2 & 96.5 & 58.6 & - & 89.5 \\
& APRIL-GAN & 4-shot & 95.0 & 94.0 & 96.9 & 96.5 & 57.4 & 55.9 & 92.1 \\
& GraphCore & 4-shot & 92.8 & - & - & 97.4 & - & - & - \\
& CoDeGraph (Ours) & 0-shot & \textbf{98.3} & \textbf{97.3} & \textbf{99.1} & \textbf{98.2} & \textbf{65.1} & \textbf{65.8} & \textbf{94.4} \\
\bottomrule
\end{tabular}
}
\end{table}

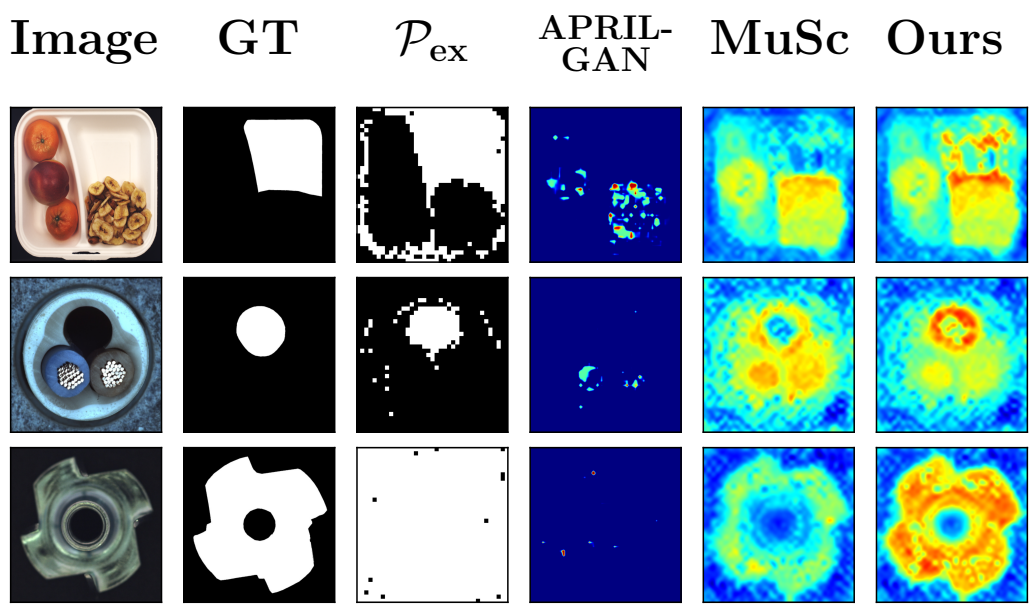
\begin{figure}[p]
\def\gridscale{0.9}
\def\labeloffset{1.5em}
\centering
\begin{adjustbox}{scale=\gridscale}
\begin{tikzpicture}[
    col label/.style={
        font=\bfseries\LARGE,
        anchor=base,
        text depth=0.25ex,
        align=center
    }
]
    \node[inner sep=0] (grid) {\includegraphics[width=6in, height=3in]{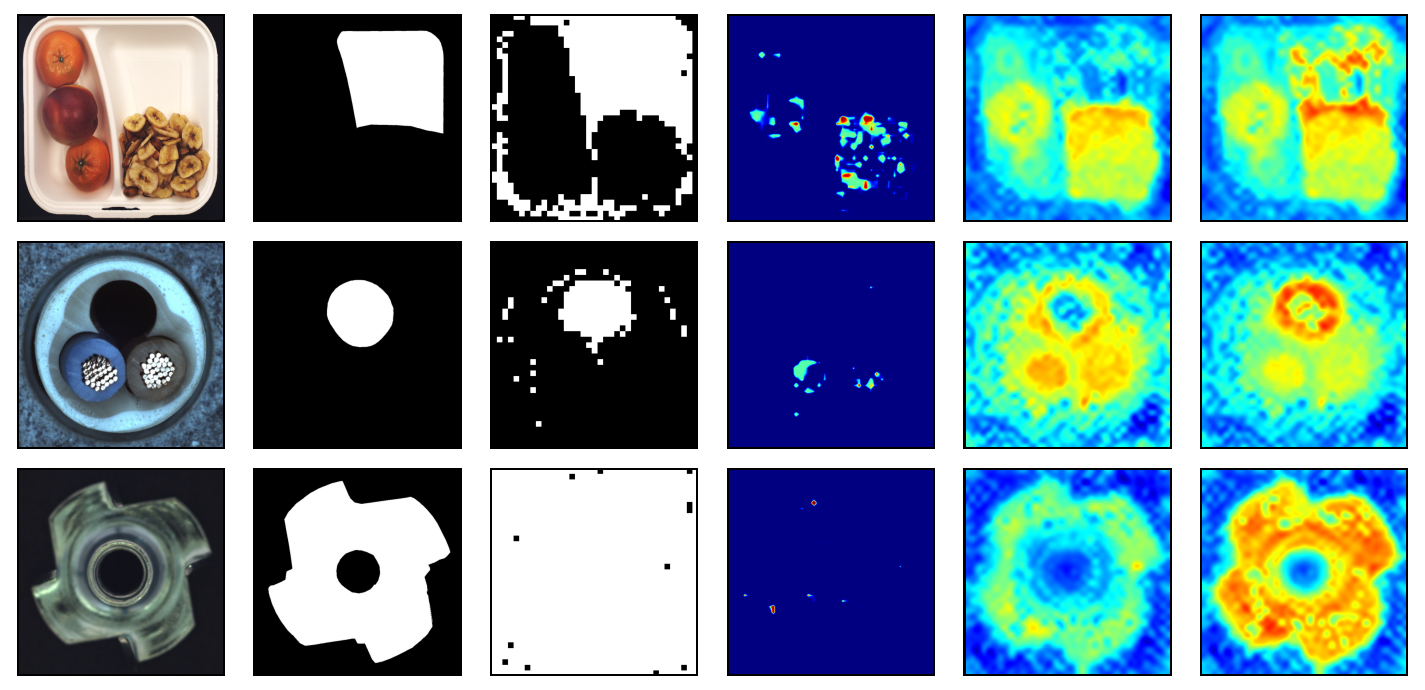}};
    \coordinate (nw) at (grid.north west);
    \coordinate (ne) at (grid.north east);
    
    \foreach \col/\label [count=\j] in {
        0/Image,
        1/GT,
        2/\( \mathcal{P}_{\text{ex}} \),
        3/{\thead{\large APRIL-\\ \large GAN}},
        4/MuSc,
        5/Ours
    } {
        \node[col label] at ($(nw)!{(\col + 0.5)/6}!(ne)$) [yshift=\labeloffset] {\label};
    }
\end{tikzpicture}
\end{adjustbox}
\caption{Objects with consistent anomalies.}
\label{fig:segmentation_comparison_consistent}
\end{figure}

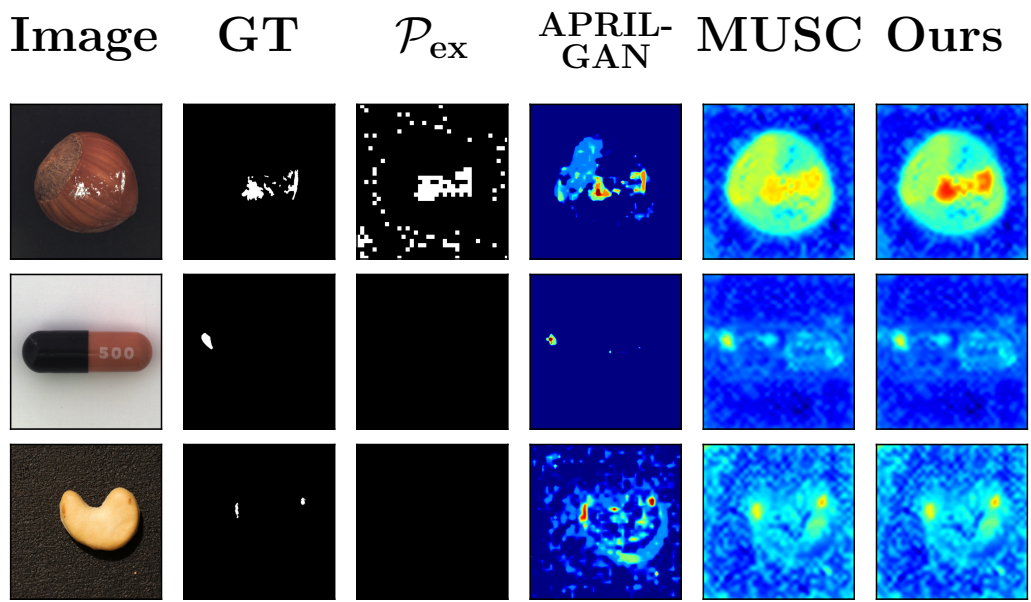
\begin{figure}[t]
\def\gridscale{0.9}
\def\labeloffset{1.5em}
\centering
\begin{adjustbox}{scale=\gridscale}
\begin{tikzpicture}[
    col label/.style={
        font=\bfseries\LARGE,
        anchor=base,
        text depth=0.25ex,
        align=center
    }
]
    \node[inner sep=0] (grid) {\includegraphics[width=6in, height=3in]{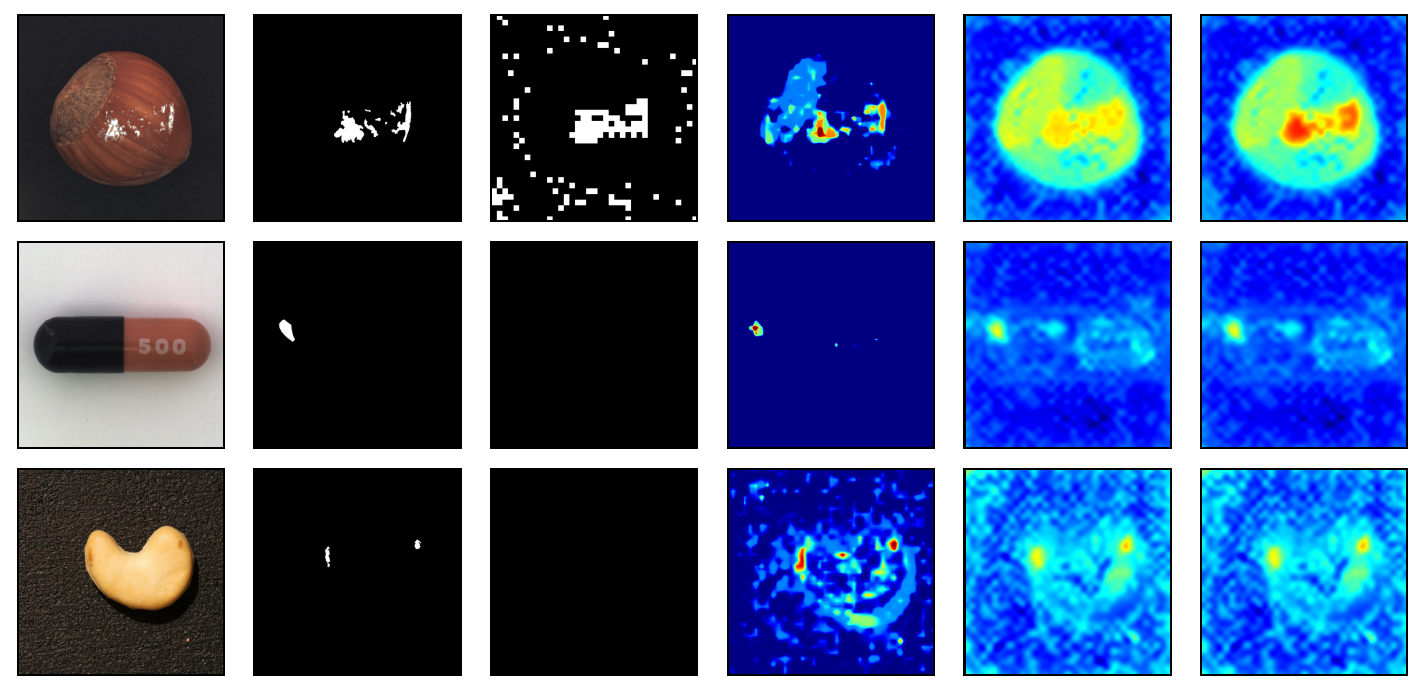}};
    \coordinate (nw) at (grid.north west);
    \coordinate (ne) at (grid.north east);
    
    \foreach \col/\label [count=\j] in {
        0/Image,
        1/GT,
        2/\( \mathcal{P}_{\text{ex}} \),
        3/{\thead{\large APRIL-\\ \large GAN}},
        4/MUSC,
        5/Ours
    } {
        \node[col label] at ($(nw)!{(\col + 0.5)/6}!(ne)$) [yshift=\labeloffset] {\label};
    }
\end{tikzpicture}
\end{adjustbox}
\caption{Objects with inconsistent anomalies.}
\label{fig:segmentation_comparison_inconsistent}
\vspace*{3cm}
\end{figure}

\paragraph{Analysis of $\mathcal{P}_{\mathrm{ex}}$.}
As shown in Figure~\ref{fig:segmentation_comparison_consistent}, the excluded set \(\mathcal{P}_{\mathrm{ex}}\) effectively captured regions associated with consistent anomalies.  
In contrast, on datasets lacking such anomalies, almost no patches $\mathbf{z}$ were excluded from \(\mathcal{B}\), as visualized in Figure~\ref{fig:segmentation_comparison_inconsistent}.  
This observation is supported quantitatively in Table~\ref{tab:pexclude_analysis}: on MVTec-CA, CoDeGraph removed 6.9\% of patches from \(\mathcal{B}\) while successfully identifying 73.9\% of consistent-anomaly patches, and on ConsistAD, 4.8\% of patches were excluded with 55.2\% recall of consistent anomalies.  
For inconsistent-anomaly datasets, the exclusion ratios were negligible—0.3\% on MVTec-IA and 0.05\% on Visa—preserving the normal pool and ensuring robustness.
\begin{table}[h]
\centering
\setlength{\tabcolsep}{6pt}
\caption{Quantitative analysis of the excluded patch set $\mathcal{P}_{\mathrm{ex}}$. 
    The table reports the exclusion rate ($|\mathcal{P}_{\mathrm{ex}}|/|\mathcal{B}|$), consistent-anomaly capture rate by $\mathcal{P}_{\mathrm{ex}}$, and patch-type distribution  
(\textbf{N}ormal / \textbf{C}onsistent / \textbf{I}nconsistent) in $\mathcal{P}_{\mathrm{ex}}$ across datasets. 
All values are expressed in percentages.}
\label{tab:pexclude_analysis}
\small
\begin{tabular}{l|c|c|cc}
\toprule
\textbf{Dataset} & 
\textbf{Excluded Rate} & 
\textbf{Capture Rate} & 
\textbf{Patch Distribution (N/C/I)} \\
\midrule
MVTec-CA   & 6.9  & 73.9 & 39.3 / 31.5 / 29.2 \\
ConsistAD  & 4.8  & 55.2 & 39.8 / 39.6 / 20.6 \\
MVTec-IA   & 0.3  & 0.9  & 92.7 / 1.7 / 5.6   \\
Visa       & 0.05 & 0.0  & 100.0 / 0.0 / 0.0  \\
\bottomrule
\end{tabular}
\end{table}
\paragraph{Qualitative Results.}
Visual comparisons in Figure~\ref{fig:segmentation_comparison_consistent} illustrate CoDeGraph’s superiority on consistent-anomaly datasets.  
Its anomaly maps comprehensively captured the defective regions—such as flipped metal nuts and missing cable components—whereas other zero-shot baselines mostly highlighted local boundaries without covering the full extent of the defect.  
For inconsistent anomalies, CoDeGraph maintained MuSc’s precision in detecting subtle anomalies, while generating fewer false positives than competing methods, as shown in Figure~\ref{fig:segmentation_comparison_inconsistent}.  
% ============================
% Figure 1: Consistent Anomalies
% ============================

% ============================
% Figure 2: Inconsistent Anomalies
% ============================
 % adjust negative or positive spacing
\clearpage

\subsection{Empirical Analysis of Key Components in CoDeGraph}
In this section, we conduct an ablation study on CoDeGraph's core mechanisms to validate its design choices.
\subsubsection*{Effect of Backbone Representations}
\label{subsec:backbone-analysis}
 CoDeGraph is designed to operate on generic vision-transformer feature spaces as long as patch-level tokens can be extracted in a stage-wise manner. To confirm that the proposed graph-based pipeline is independent of specific pre-training objectives or ViT configurations, we evaluated several popular backbones—DINO, DINOv2, and CLIP—using an identical CoDeGraph pipeline. 

Precisely, for all ViT-based backbones, we used the same four-stage division scheme across all architectures: ViT-Large models were divided into 4 stages with 6 layers each, while ViT-Base models used 3 layers per stage. We extract patch tokens from each stage and utilize the final-layer \texttt{[CLS]} token for RsCIN classification optimization, as in the main experiments. All other settings—graph construction, coverage-based link selection, outlier community detection, and targeted patch removal—remained identical to the main experiments to isolate backbone effects.

\begin{table}[h]
\centering
\caption{Performance of CoDeGraph with different ViT backbones on MVTec AD. All metrics are in \%.}
\label{tab:backbone_thesis}
\resizebox{\textwidth}{!}{%
\begin{tabular}{c|c|c|cc|ccc}
\toprule
\textbf{Pre-training} & \textbf{Architecture} & \textbf{Pre-training Data} &
\textbf{AUROC-cls} & \textbf{F1-cls} &
\textbf{AUROC-seg} & \textbf{F1-seg} & \textbf{AP-seg} \\
\midrule
\multirow{2}{*}{DINO~\citep{DINO}}
  & ViT-B-8 & ImageNet-1k & 97.1 & 97.1 & 98.4 & 65.7 & 67.8 \\
  & ViT-B-16 & ImageNet-1k & 96.8 & 96.9 & 98.6 & 66.9 & 69.1 \\
\midrule
\multirow{2}{*}{DINOv2~\citep{DINOv2}}
  & ViT-B-14 & LVD-142M & 97.6 & 97.5 & 98.7 & 67.1 & 69.3 \\
  & ViT-L-14 & LVD-142M & 98.0 & \textbf{97.7} & \textbf{98.9} & \textbf{69.1} & \textbf{71.9} \\
\midrule
\multirow{2}{*}{CLIP~\citep{CLIP}}
  & ViT-B-16 & WIT-400M & 96.2 & 95.8 & 98.1 & 65.9 & 67.0 \\
  & ViT-L-14-336 & WIT-400M & \textbf{98.3} & 97.4 & 98.2 & 66.8 & 68.1 \\
\bottomrule
\end{tabular}
}
\end{table}

Table~\ref{tab:backbone_thesis} shows that all backbones deliver strong anomaly classification and segmentation performance under the same pipeline. Larger-capacity models with higher-resolution pre-training (e.g., DINOv2 ViT-L-14) yield the most consistent improvements, aligning with the expectation that denser, semantically richer features produce better performance on downstream tasks. In conclusion, our CoDeGraph is backbone-agnostic, as it works effectively with any ViT-style backbone exposing patch tokens across layers.

\subsubsection*{Effect of Outlier Community Detection and Targeted Patch Removal.}
We tested IQR vs. top-3 community selection on \( \mathcal{S}_c \) with/without patch filtering. The results are presented in Table~\ref{tab:pf_ablation}. On MVTec-CA, performance remained stable, with pixel-wise F1 decreasing from 73.8\% to 73.5\%. On MVTec-IA, where consistent anomalies are rare, selecting the top-3 communities without patch filtering reduced pixel-wise F1 by 2.9\% and image-wise AUROC by 2.0\%, as it automatically removed a significant number of patches from \(\mathcal{B}\) without selection. Algorithm~\ref{alg:targeted-filtering-thesis} mitigated this negative impact by removing only high-dependency patches.
Sensitivity to Tukey’s parameter \(k_{\mathrm{IQR}}\) was also examined in Table~\ref{tab:tukey_k_ablation}; standard values \(k_{\mathrm{IQR}}=1.5\) and \(3\) yielded nearly identical results to \(k_{\mathrm{IQR}}=4.5\), confirming robustness.  
Together, these experiments demonstrate that community-level outlier detection and targeted patch removal act as complementary safeguards that preserve CoDeGraph’s stability across datasets.
\begin{table}[h]
\centering
\captionsetup{font=small, labelfont=bf}
\caption{Effect of outlier community detection strategies (IQR vs.\ Top-3) with and without targeted patch filtering. All values are in \%.}
\label{tab:pf_ablation}
\setlength{\tabcolsep}{5pt}%
\renewcommand{\arraystretch}{1.15}%
\resizebox{0.9\textwidth}{!}{%
\begin{tabular}{l|l|l|c|c|c|c}
\toprule
\textbf{Dataset} & \textbf{Method} & \thead{\textbf{Patch} \\ \textbf{Filtering}} &
\thead{\textbf{Capture}\\\textbf{Rate}} & \thead{\textbf{Removed} \\ \textbf{Normal Patches}} & \textbf{AUROC-cls} & \textbf{F1-seg} \\ 
\midrule
\multirow{4}{*}{MVTec-CA}
 & IQR   & Enabled \ding{51}  & 73.9 & 2.7  & 98.5 & 73.8 \\
 & IQR   & Disabled \ding{53}  \ & 90.9 & 8.9  & 98.4 & 73.8 \\
 & Top-3 & Enabled \ding{51} & 73.9 & 2.8  & 98.5 & 73.8 \\
 & Top-3 & Disabled \ding{53}& 90.9 & 12.5 & 98.4 & 73.5 \\
\midrule
\multirow{4}{*}{MVTec-IA}
 & IQR   & Enabled \ding{51} & 0.9  & 0.2  & 98.3 & 65.0 \\
 & IQR   & Disabled \ding{53}& 4.9  & 3.2  & 98.2 & 64.9 \\
 & Top-3 & Enabled \ding{51} & 2.0  & 3.4  & 96.8 & 63.2 \\
 & Top-3 & Disabled \ding{53}& 21.0 & 18.7 & 96.3 & 62.1 \\
\bottomrule
\end{tabular}%
}
\end{table}
\begin{table}[h]
\centering
\setlength{\tabcolsep}{6pt}
\small
\caption{Effect of $k_{\textrm{IQR}}$ on outlier detection for MVTec-AD. All metrics are in \%.}
\label{tab:tukey_k_ablation}
\begin{tabular}{l|cc|ccc}
\toprule
\textbf{Value} & \textbf{AUROC-cls} & \textbf{F1-cls} & \textbf{AUROC-seg} & \textbf{F1-seg} & \textbf{AP-seg} \\
\midrule
$k_{\textrm{IQR}} = 1.5$        & 98.0 & 97.1 & 98.1 & 66.3 & 66.6 \\
$k_{\textrm{IQR}} = 3.0$        & 98.3 & 97.4 & 98.2 & 66.4 & 66.8 \\
$k_{\textrm{IQR}} = 4.5$ (default) & 98.3 & 97.4 & 98.2 & 66.8 & 68.1 \\
\bottomrule
\end{tabular}
\end{table}

\subsubsection*{Sensitivity to Reference Rank \(\omega\).}
An effective reference rank \(\omega\) should allow normal patches to locate similar neighbors at $\omega$ while consistent anomalies fail to do so.  
Extreme settings at \(10\% B\) and \(90\% B\) led to performance degradation (Table~\ref{tab:omega_ablation}).  
At \(\omega=10\% B\), consistent anomalies—such as flipped \texttt{metal\_nut} images accounting for 20\% of the test set—could easily match each other, while at \(\omega=90\% N\), normal patches struggled to identify valid counterparts.  
Both extremes violate the neighbor-burnout principle, leading to reduced AC/AS performance.  
Intermediate values between 30\% and 70\% of \(B\) typically provided reliable results for CoDeGraph.  
\begin{table}[h]
\centering
\setlength{\tabcolsep}{6pt}
\caption{Quantitative analysis of reference index \(\omega\) on MVTec-CA. 
The parameter \(\omega\) controls the reference rank used in endurance ratio computation. 
All metrics are expressed in percentages.}
\label{tab:omega_ablation}
\setlength{\tabcolsep}{15pt}
\small
\begin{tabular}{c|c|c}
\toprule
\(\boldsymbol{\omega}\) & 
\textbf{AUROC-cls} & 
\textbf{F1-seg} \\
\midrule
0.1N & 94.1 & 56.1 \\
0.3N & 98.5 & 73.8 \\
0.5N & 98.5 & 73.3 \\
0.7N & 98.5 & 73.3 \\
0.9N & 94.1 & 55.9 \\
\bottomrule
\end{tabular}
\end{table}

\subsubsection*{Role of Coverage-Based Link Selection.}
Coverage-based selection was introduced in Stage~2 (Algorithm~\ref{alg:coverage-selection-thesis}) to ensure that the anomaly similarity graph \(\mathcal{G}=(V,E)\) retains sufficient connections for outlier community detection. In most experiments, the required coverage was achieved with fewer than \(\binom{N}{2}\) links in \(S_l\). However, fixing the number of links proved unreliable, particularly when consistent-anomaly images were dominated by consistent anomalies (e.g., flipped \texttt{metal\_nut}). 

For illustration, if \(q\) identical anomalous images (each with \(M\) patches) are present, then \(\mathcal{G}\) contains only zero-distance links between them when \(|S_l| < M \cdot \binom{q}{2}\), leaving no baseline for IQR-based detection. Empirical tests on \texttt{metal\_nut} confirm this: for \(|S_l| = \binom{N}{2}\), the coverage was 54.7\%, while for \(|S_l| = \binom{N}{2}/2\), the coverage dropped to 33.9\%. In both cases, CoDeGraph failed to identify consistent-anomaly communities as outliers, demonstrating the necessity of coverage-based selection.

\subsubsection*{Choice of Community Detection Algorithm}
Modularity optimization \citep{newman2004finding} is a widely adopted approach that partitions graphs by maximizing the difference between observed and expected edge weights under a degree-based null model. The modularity score is given by:
\[
Q = \sum_{ij}\left(A_{ij} - \frac{k_i k_j}{2m}\right)\delta(\sigma_i, \sigma_j),
\]
where \( A_{ij} \) is the adjacency matrix, \( k_i \) and \( k_j \) are node degrees, \( m \) is the total edge weight, and \( \delta(\sigma_i, \sigma_j) = 1 \) if nodes \( i \) and \( j \) are in the same community.

\begin{figure}[h]
\centering
\includegraphics[width=0.6\textwidth]{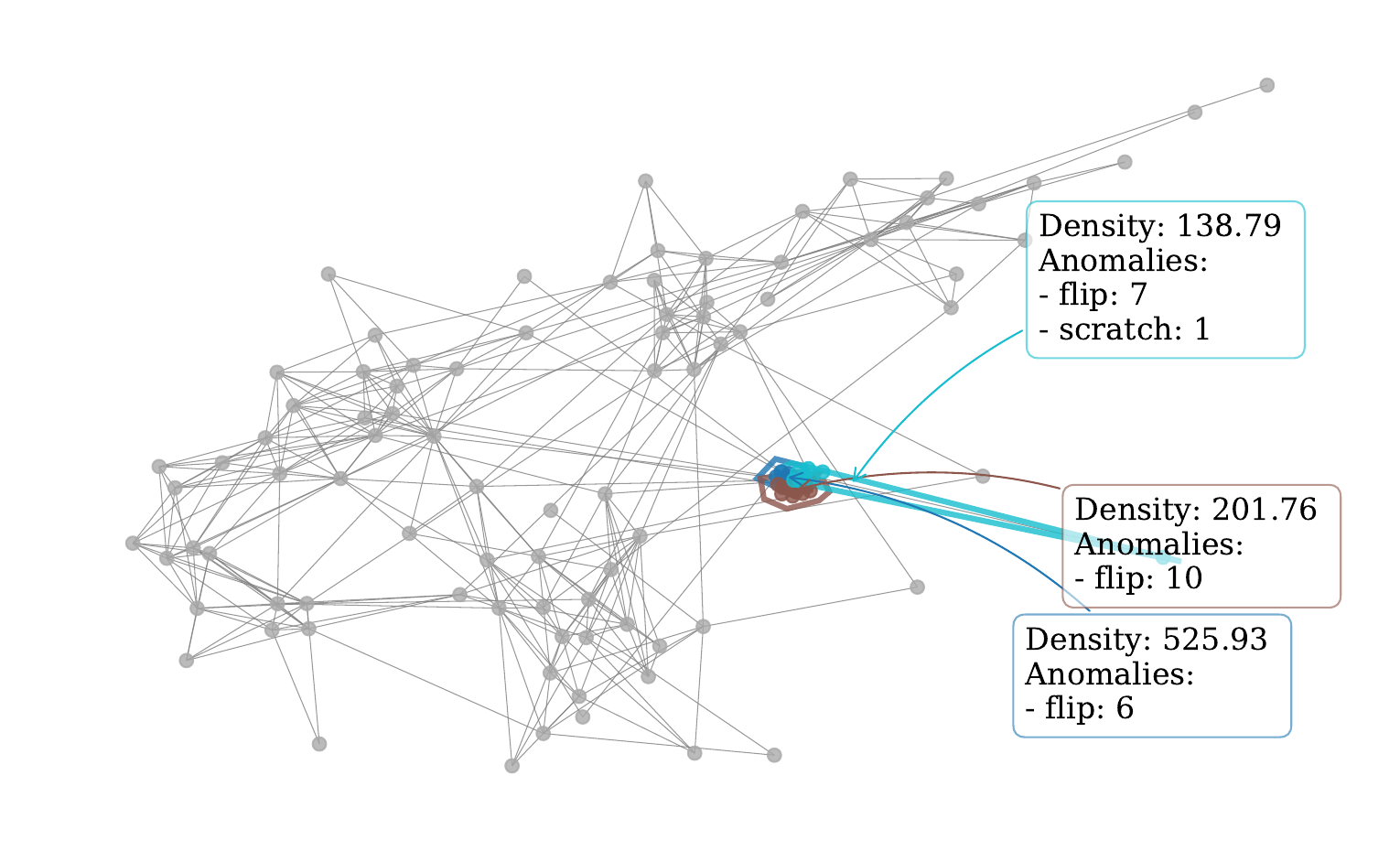}
\caption{Modularity-based community detection fragments a consistent anomaly type (flipped metal nuts) into several smaller communities due to weak intra-group links falling below degree-based expectations. This fragmentation interferes with reliable detection of consistent-anomaly groups.}
\label{fig:modularity_fragmentation}
\end{figure}

However, modularity's dependence on relative node degrees (via expected edge weight term $k_i k_j/2m$) is not compatible with our anomaly similarity graph $\mathcal{G}$. In $\mathcal{G}$, consistent-anomaly communities are expected to exhibit much stronger internal connectivity compared to the rest of the graph. Nonetheless, there still exists some weaker links within these communities. When such weak edges fall below the expected weight threshold, modularity penalizes them, even though the overall community remains densely connected. As a result, strongly cohesive anomaly clusters may be artificially fragmented into smaller subgroups. This fragmentation undermines our IQR-based community scoring, diluting the density of anomaly clusters and hindering the reliable detection of consistent-anomaly patterns. As shown in Figure~\ref{fig:modularity_fragmentation} for the \texttt{metal\_nut} class, the dense community of flip metal nut fragments.

In contrast, CPM evaluates communities based on an absolute resolution parameter \( \gamma \), independent of degrees, making it better suited for preserving intact consistent-anomaly groups. We set \( \gamma \) to the 25th percentile of edge weights, ensuring connections within anomaly communities exceed this threshold and avoiding fragmentation. This leads to more robust detection of outlier communities, enhancing patch filtering and zero-shot segmentation stability.

\subsubsection*{Influence of Weighted Endurance Ratio \(\alpha\).}
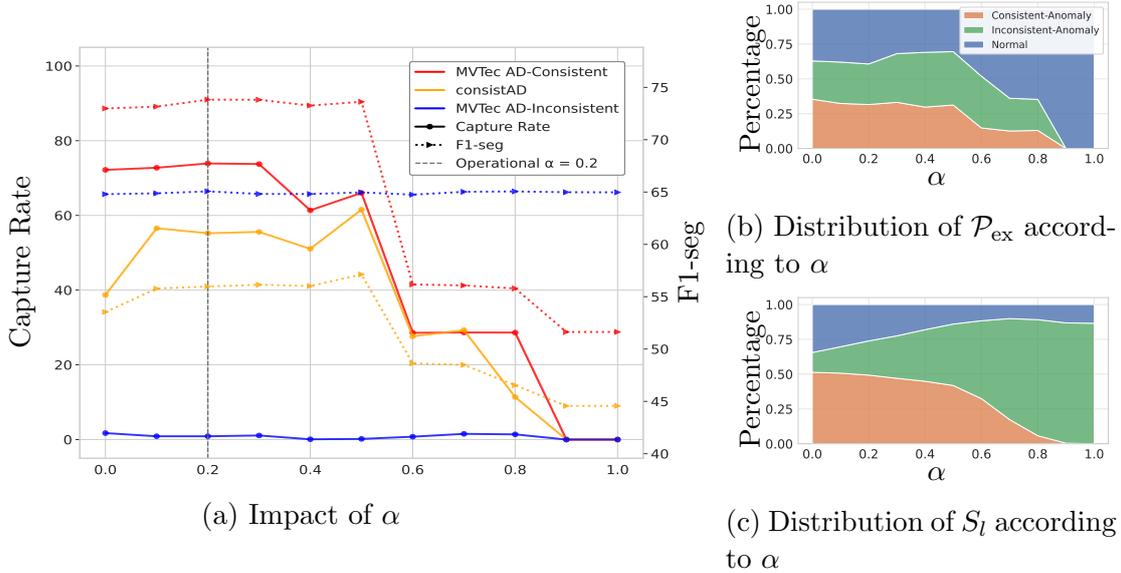
\begin{figure}[h]
  \centering
    \begin{minipage}{0.54\textwidth}  % [t][height][c] = top align, fixed height, center content
  \begin{subfigure}{\textwidth}
    \centering
      \centering
      \begin{tikzpicture}
        \node[inner sep=0pt] (img) {\includegraphics[height=2.3in, width=3.3in]{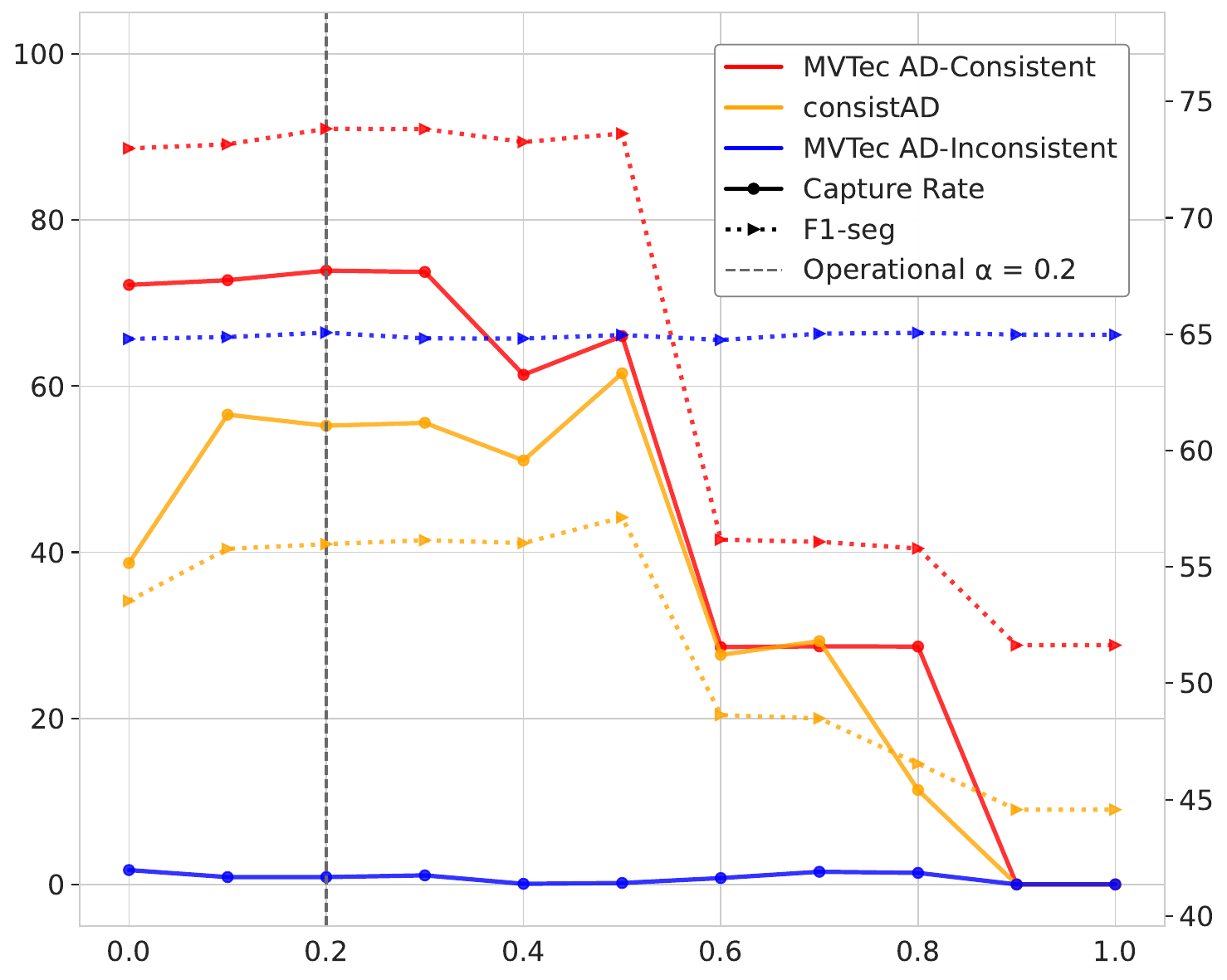}};
        \node[rotate=90, anchor=center] at ([xshift=-0.2cm]img.west) {\small Capture Rate};
        \node[rotate=90, anchor=center] at ([xshift=0.2cm]img.east) {\small F1-seg};
      \end{tikzpicture}
      \caption{Impact of \( \alpha \)}
      \label{fig:endurance_ratio}
  \end{subfigure}
    \end{minipage}
    \hspace{0.2cm}
  \hfill
    \begin{minipage}{0.35\textwidth}  % Same height, but top-align content
  \begin{subfigure}{\textwidth}
    \centering
      \begin{subfigure}{\textwidth}
        \centering
        \begin{tikzpicture}
          \node[inner sep=0pt] (img) {\includegraphics[height=0.9in, width=0.9\textwidth]{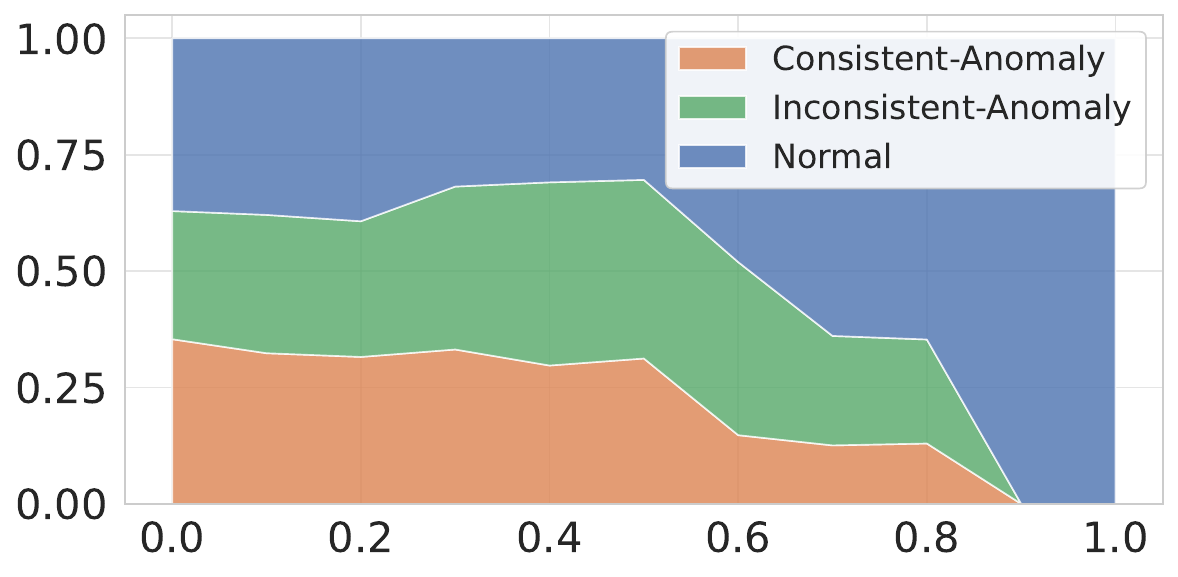}};
\node[rotate=90, anchor=center] at ([xshift=-0.1cm]img.west) {\small Percentage};
          \node[anchor=center] at ([yshift=-0.1cm]img.south) {\small \( \alpha \)};

        \end{tikzpicture}
        \caption{Distribution of $\mathcal{P}_{\text{ex}}$ according to \( \alpha \)}
        \label{fig:excluded_patches}
      \end{subfigure}
      
      \vspace{0.1cm}
      
      \begin{subfigure}{\textwidth}
        \centering
        \begin{tikzpicture}
          \node[inner sep=0pt] (img) {\includegraphics[height=0.9in, width=0.9\textwidth]{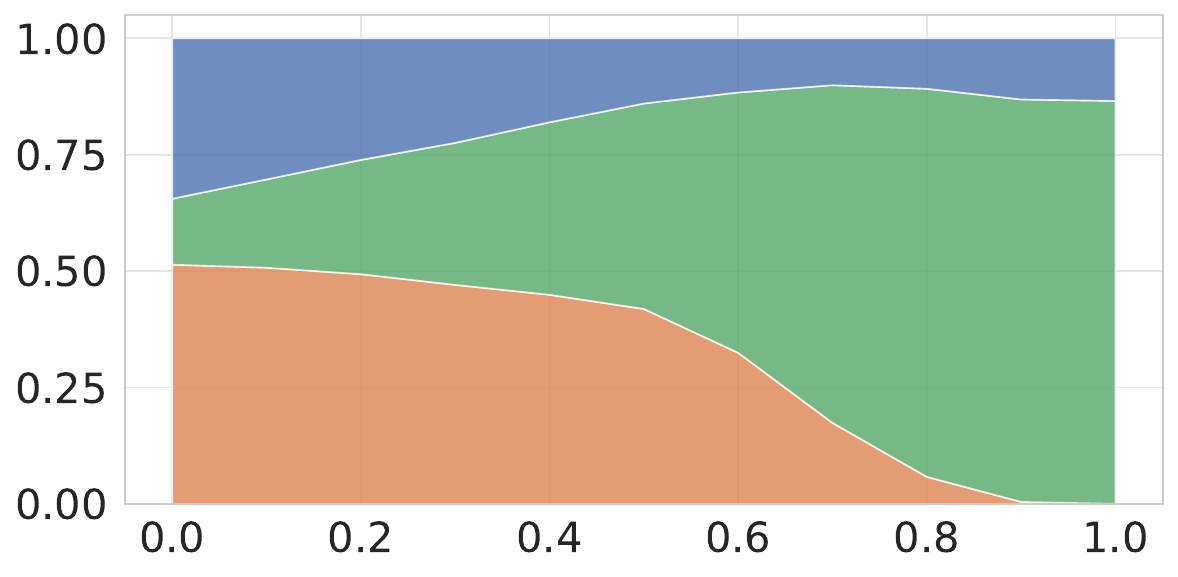}};
          \node[rotate=90, anchor=center] at ([xshift=-0.1cm]img.west) {\small Percentage};
          \node[anchor=center] at ([yshift=-0.1cm]img.south) {\small \( \alpha \)};
        \end{tikzpicture}
        \caption{Distribution of $S_l$ according to \( \alpha \)}
        \label{fig:suspicious_links}
      \end{subfigure}
  \end{subfigure}
    \end{minipage}
  \caption{ Influence of weight endurance ratio \( \alpha \). (Left) Impact of \( \alpha \) on capture rate and F1-seg performance. (Right) Impact of \( \alpha \) on $\mathcal{P}_{\text{ex}}$ and suspicious link $S_{l}$ on MVTec-CA.}
  \label{fig:ablation_endurance_ratio}
\end{figure}

Figure~\ref{fig:ablation_endurance_ratio} shows the relationship between the weighting parameter \(\alpha\), consistent-anomaly capture rate by $\mathcal{P}_{\mathrm{ex}}$, and segmentation performance.  
For \(\alpha > 0.5\), a reduction in capture rate led to decreased pixel-wise F1, whereas for \(\alpha \in [0.0, 0.5]\), both metrics remained stable across datasets.  
On ConsistAD, which exhibits high variability in normal patterns, increasing \(\alpha\) improved both capture rate and segmentation accuracy, highlighting its role in enhancing the separability of consistent-anomaly communities.  
Conversely, for inconsistent-anomaly datasets such as MVTec-IA, performance remained unchanged for all \(\alpha\) values, indicating that the weighting does not distort normal-similarity structures.
\subsection{Limitations and Practical Considerations}
\label{sec:limitations}

This subsection discusses both the practical and conceptual limitations of CoDeGraph, covering computational scalability and its behavior under dataset with multi-modal normal variantions.
\subsubsection*{Computational Efficiency and Memory Footprint}
\label{subsubsec:efficiency}
CoDeGraph introduces negligible additional computation compared to MuSc~\citep{MuSc}, its runtime and GPU memory consumption scale with the number of images per batch \(B\) and the number of patch tokens per image \(N\).  
The pairwise similarity search across all patches remains the dominant bottleneck with quadratic complexity \(O(B^2N^2)\). Furthermore, CoDeGraph additionally requires maintaining mutual similarity indices for constructing the anomaly similarity graph \(\mathcal{G}\). To mitigate these problems, two optimization strategies are adopted in CoDeGraph:
\begin{itemize}
    \item \textbf{Subset Division.}  
    The test set is divided into \(s\) equal-sized subsets (chunks), and each subset is processed independently with identical procedures.  
    This design ensures stable throughput and bounded VRAM usage because only one subset is held in memory at a time.  
    The final anomaly maps are aggregated after all subsets are processed.
    \item \textbf{CLS Token-based Screening.}  
    For each image \(C_i\), cosine similarity is computed between its \texttt{[CLS]} token and those of all base images in \(\mathcal{B}\).  
    Only the top \(\eta \%\) most similar images are retained for patch-level similarity computation, effectively reducing complexity from \(O(B^{2}N^{2})\) to \(O(\eta B^{2}N^{2})\).  
    The hyperparameter \(\eta \in [0,1]\) controls the nearest-neighbor fraction used for the calculation of mutual similarity vector in Eq.\eqref{eq:msv}, which results a vector of length $\lceil \eta B \rceil$.
\end{itemize}

Table~\ref{tab:inference_time} summarizes the inference-time analysis on MVTec-AD.  
Combining subset division (\(s\)) and \texttt{[CLS]}-based screening (\(\eta\)) yields up to a 37\% reduction in runtime with negligible accuracy loss (\(<1.5\%\) in F1-seg).  
However, as confirmed by the extended evaluation in Table~\ref{tab:cls_screening_extended}, excessively small neighbor fractions (\(\eta < 30\%\)) result in insufficient neighbor coverage, limiting the detection of the neighbor-burnout phenomenon and reducing both AUROC-cls and segmentation F1.  
Empirically, \(s = 2\) and \(\eta = 60\%\) provide the best balance between efficiency and accuracy.

\begin{table}[ht]
\centering
\caption{Inference-time ablation on MVTec-AD. \(s\) denotes the number of equal-sized subsets (chunks) processed independently, and \(\eta\) represents the \texttt{[CLS]}-based nearest-neighbor fraction. All experiments were conducted on a single RTX~4070Ti~Super GPU.}
\label{tab:inference_time}
\begin{tabular}{l|c|c|c|c|c}
\toprule
\textbf{Method} & \textbf{$s$} & \textbf{$\eta$} (\%) & \textbf{Time (ms)} & \textbf{AUROC-cls (\%)} & \textbf{F1-seg (\%)} \\
\midrule
MuSc & 1 & 100 & 280.46 & 97.5 & 62.3 \\\addlinespace[2pt]
\hline \addlinespace[2pt]
\multirow{4}{*}{CoDeGraph}
 & 1 & 100 & 281.34 & 98.3 & 66.8 \\
 & 1 & 60 & 211.76 & 98.4 & 66.0 \\
 & 2 & 100 & 192.17 & 98.1 & 66.1 \\
 & 2 & 60 & 158.32 & 98.1 & 65.7 \\
\bottomrule
\end{tabular}
\end{table}

\begin{table}[ht]
\centering
\caption{Extended evaluation of \texttt{[CLS]} token-based screening on MVTec-AD for varying nearest-neighbor fractions \(\eta\). All experiments were conducted on a single RTX~4070Ti~Super GPU.}

\label{tab:cls_screening_extended}
\begin{tabular}{l|c|c|c|c}
\toprule
\textbf{Method} & $\eta$ (\%) & Time (ms) & AUROC-cls (\%) & F1-seg (\%) \\
\midrule
\multirow{4}{*}{CoDeGraph}
 &100 & 281.34 & 98.3 & 66.8 \\
 &80 & 246.33 & 98.2 & 66.5 \\
 &60 & 211.76 & 98.4 & 66.0 \\
 &40 & 178.52 & 98.3 & 65.2 \\
 &20 & 148.62 & 96.7 & 60.5 \\
\bottomrule
\end{tabular}
\end{table}
\paragraph{Memory Considerations.}
The primary memory cost in MuSc arises from three components:
(i) the ViT model weights,  
(ii) the extracted patch features of all images in the base set \(\mathcal{B}\), and  
(iii) the mutual similarity vectors.  
CoDeGraph adds moderate overhead by storing ranking indices required for filtering operations in Algorithm~\ref{alg:coverage-selection-thesis} and Algorithm~\ref{alg:targeted-filtering-thesis}.  
These indices form tensors of shape \([L,\, B,\, N,\, B{-}1]\),  
where \(L\) is the number of ViT layers, \(B\) the number of images per batch (or processing subset), and \(N\) the number of patch tokens per image;  
the \((B{-}1)\) term excludes self-comparisons.

Instead of recomputing distance vectors after refinement,  
CoDeGraph reuses these stored indices and applies a \texttt{torch.isin}-based masking to invalidate distances associated with the excluded patches \(\mathcal{P}_{\mathrm{ex}}\).  
This approach slightly increases VRAM usage but avoids repeated similarity computation, resulting in negligible running time.  
As shown in Table~\ref{tab:inference_time}, the additional indexing cost produces no measurable slowdown.

For memory-limited environments, index tensors can be offloaded to CPU memory while feature computations remain on GPU.  
On a 200-image dataset, this hybrid configuration reduced VRAM from 10.4~GB to 7.4~GB,  
with average inference time increasing only from 281~ms to 291~ms per image on MVTec-AD.  
Even for multi-iteration cases such as MVTec-CA (where Algorithm~\ref{alg:targeted-filtering-thesis} loops over multiple outlier communities in \(\mathcal{S}_c\)),  
runtime increased marginally from 327~ms to 350~ms per image.

\subsubsection*{Challenges with Multi-Modal Normal Objects}
\label{subsubsec:multimodal}

\begin{figure}[ht]
  \centering
  \begin{subfigure}{0.48\textwidth}
    \centering
    \includegraphics[width=\textwidth]{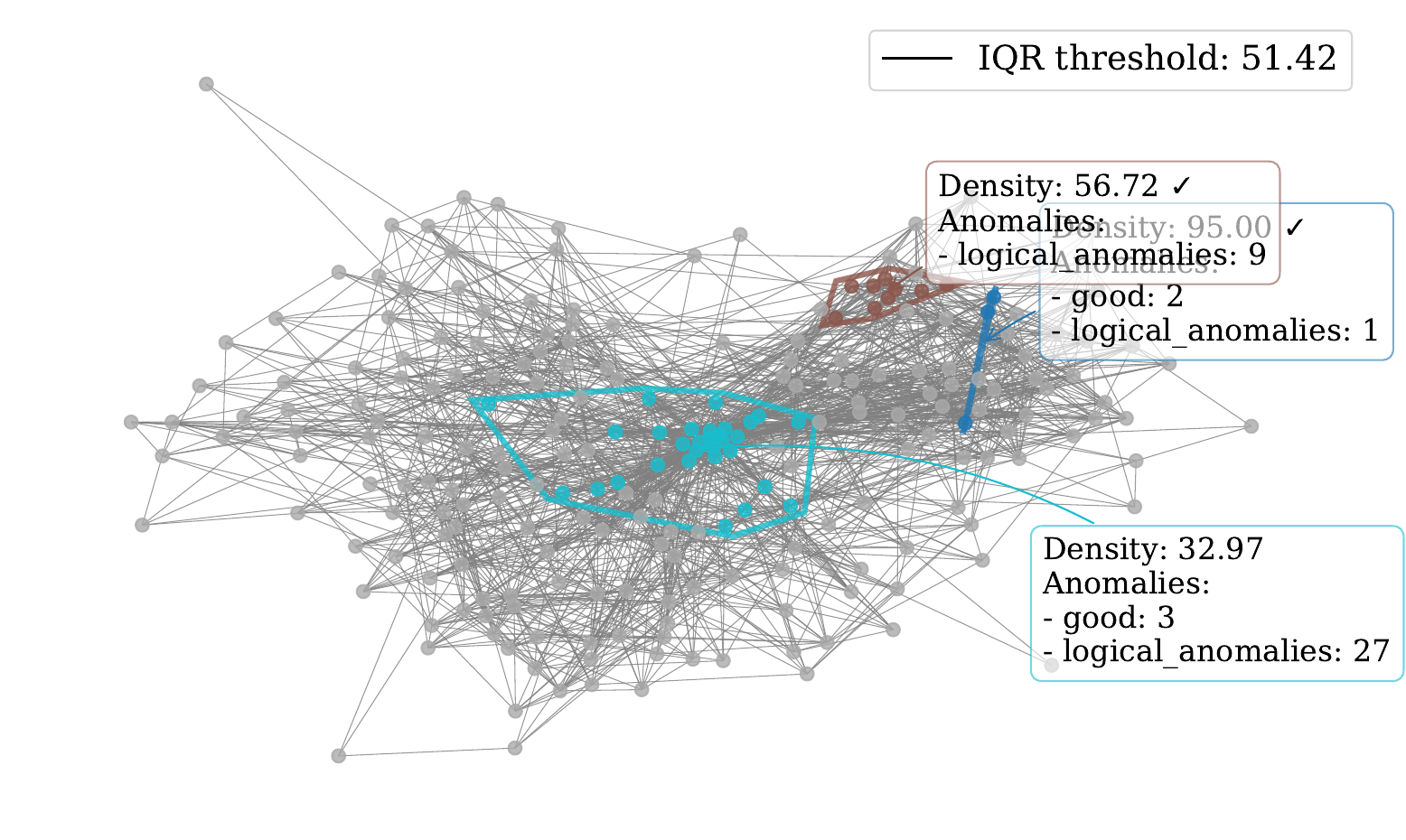}
    \caption{Multi-modal case with \(\omega = 10\%\) of \(N\)}
  \end{subfigure}
  \hfill
  \begin{subfigure}{0.48\textwidth}
    \centering
    \includegraphics[width=\textwidth]{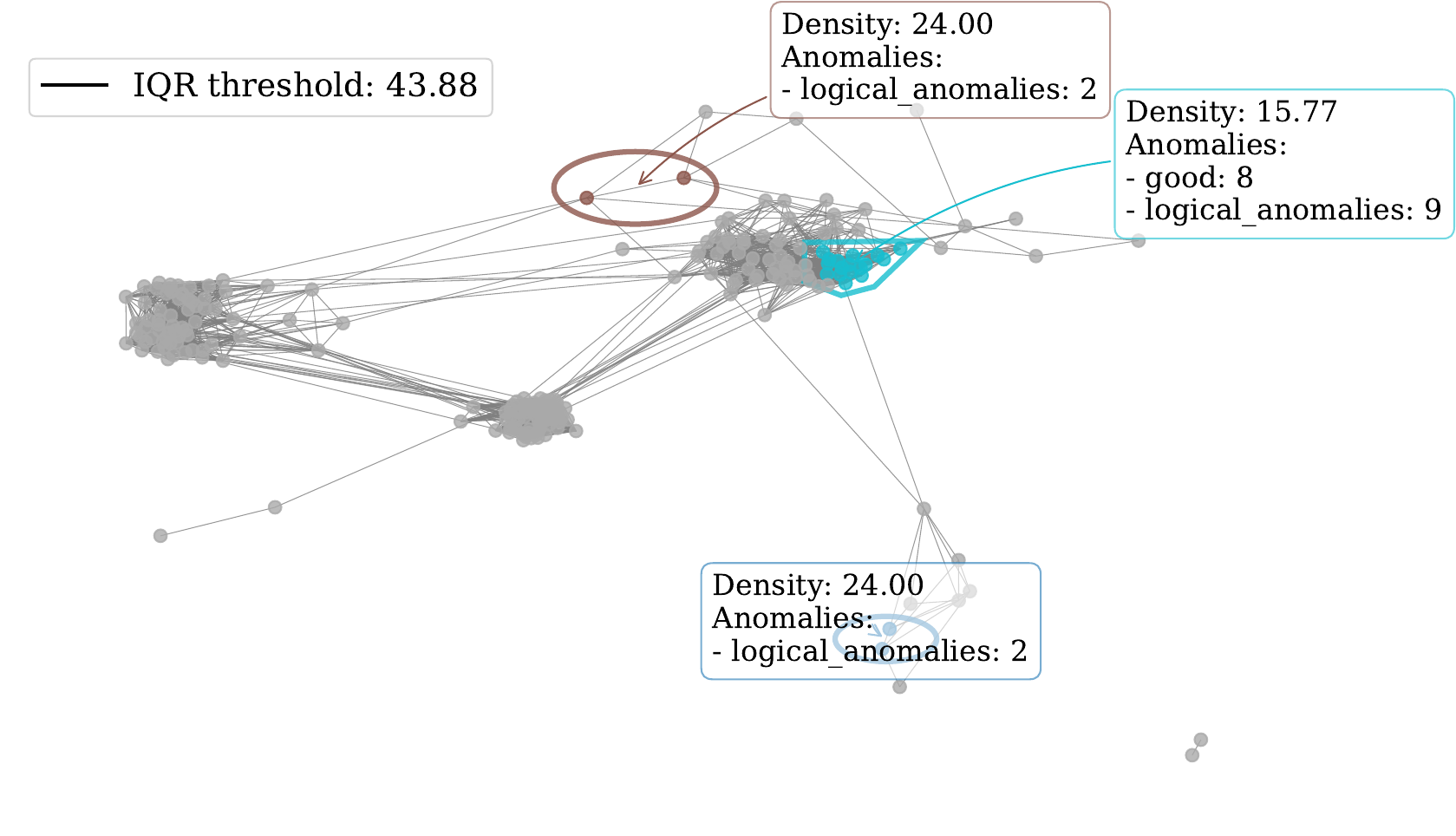}
    \caption{Multi-modal case with \(\omega = 50\%\) of \(N\)}
  \end{subfigure}
  \caption{Visualization of multi-modal juice bottle classification. (a) At \(\omega = 10\%\) of \(N\), connections among anomalous and consistent-anomaly images are clear and detectable by CPM. (b) At \(\omega = 50\%\) of \(N\), variant-based clusters dominate.}
  \label{fig:juice_bottle_multimodal}
\end{figure}

In many industrial datasets, products exhibit multiple normal variants corresponding to distinct manufacturing configurations. For example, the \texttt{juice\_bottle} subset of MVTec-LOCO includes three normal variants---banana, orange, and tomato juice bottles---each forming its own cluster in feature space. When such multi-modal normals exist, the reference index \(\omega\) must satisfy
\[
\omega < \frac{N}{C},
\]
where \(C\) is the number of normal variants. Violating this condition causes the similarity graph to cluster by variant rather than anomaly, potentially misidentifying normal variants as outlier communities, as shown in Figure~\ref{fig:juice_bottle_multimodal}.

\paragraph*{The unsolvable problem.}
A related limitation—expected in multi-modal normal objects, though possible in unimodal cases as well—arises when consistent anomalies outnumber some specific normal variants, violating the Minority assumption (Assumption~\ref{assump:limited-consistent}). When this assumption is violated, dense clusters may represent normal images rather than anomalies, making CoDeGraph vulnerable to misclassification.

\begin{figure}[h]
  \centering
  \includegraphics[width=0.6\textwidth]{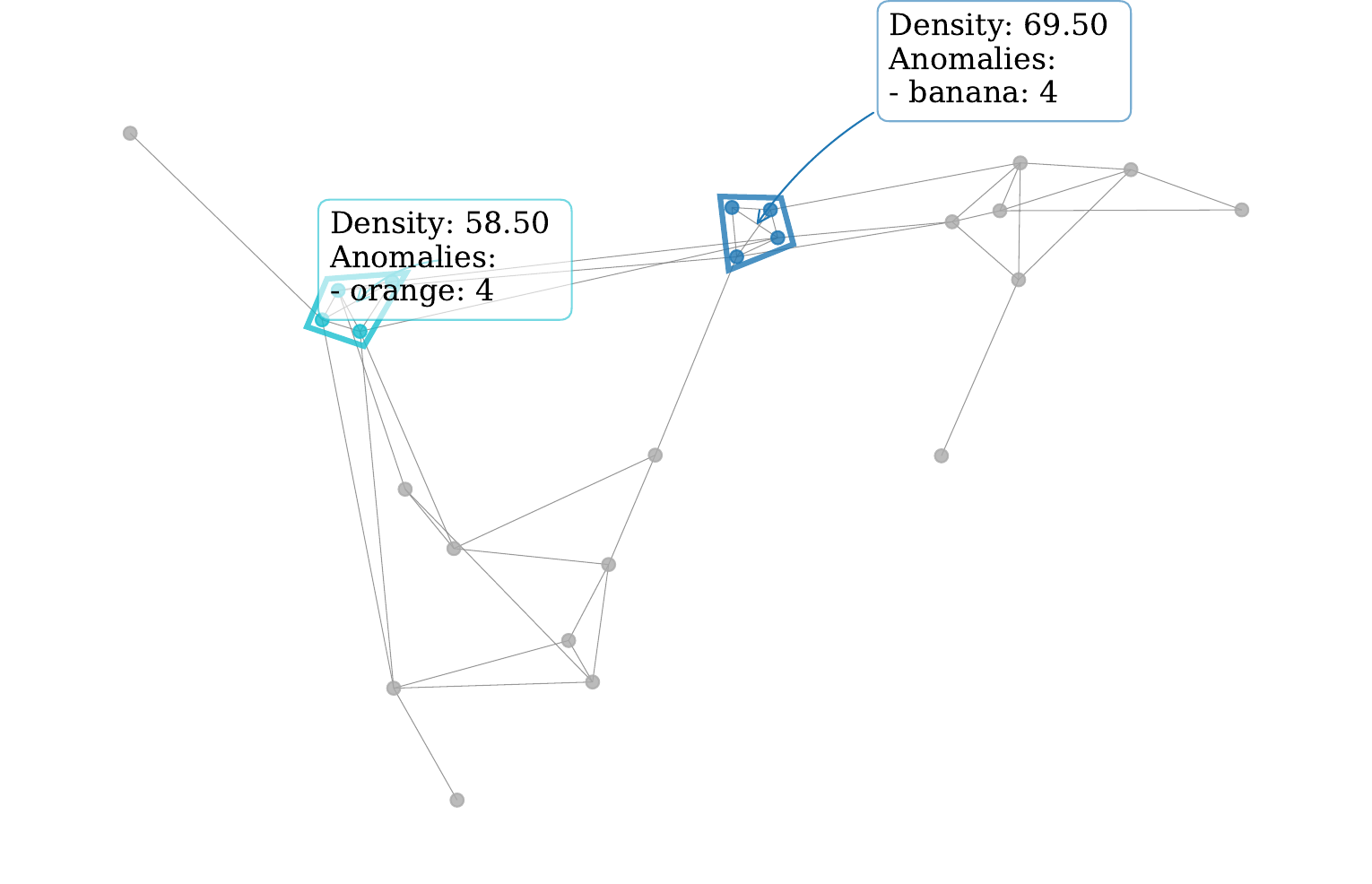}
  \caption{Failure case where dominant consistent anomalies outnumber normal variants, leading to inversion of community labels.}
  \label{fig:failure_mode}
\end{figure}

Due to the absence of multi-modal industrial datasets with dominant consistent anomalies, we designed a controlled experiment using the \texttt{juice\_bottle} subclass from MVTec LOCO. We selected 8 tomato, 4 orange, and 4 banana juice bottle images as normal variants. To simulate dominant anomalies, we introduced 8 consistent-anomaly images (logical anomalies lacking orange labels, e.g., \texttt{logical\_anomalies\_000}, \texttt{003}, \texttt{010}, \texttt{011}, \texttt{072}, \texttt{077}, \texttt{078}, \texttt{079}), outnumbering the orange and banana variants. In the similarity graph (Figure~\ref{fig:failure_mode}), the less frequent orange and banana variants form dense clusters, which can be wrongly identified as anomalous communities.

Although such cases are rare in public datasets (e.g., MVTec-AD, VisA), they reveal CoDeGraph’s dependence on the normal-majority assumption. Future extensions could integrate pre-clustering of normal variants or hierarchical community detection to separate variant-level and anomaly-level structures.
% ------------------------------------------------------
\chapter{Theoretical Explanation for the Similarity Scaling Phenomenon}
\label{chap:theory-scaling}
In this chapter, theoretical evidence for the \emph{Similarity Scaling Phenomenon} is presented. The analysis builds upon and extends the theoretical framework introduced in our published work~\cite{CoDeGraph}.

\medskip

The chapter is organized as follows.  
Section~\ref{sec:theory-overview} outlines the main theoretical motivations and summarizes the core contributions.  
Section~\ref{sec:long-range-independence} develops a probabilistic derivation of the Similarity Scaling Phenomenon under long-range independence using Extreme Value Theory (EVT).  
Section~\ref{sec:geom-intuition} provides geometric intuition by linking local manifold geometry to power-law similarity distributions.  
Subsequent subsections present a toy model based on Poisson-distributed patches and analyze local ball asymptotics under regular variation.  
Finally, Section~\ref{sec:theory-summary} concludes the chapter with a summary of theoretical findings and their implications for zero-shot anomaly detection.

\section{Overview and Main Contributions}
\label{sec:theory-overview}

The \emph{Similarity Scaling} phenomenon describes a universal regularity
in the growth of the {Mutual Similarity Vector}:
\[
  \mathcal{D}_{\mathcal{B}}(\mathbf{z})
  = [d(\mathbf{z})_{(1)}, d(\mathbf{z})_{(2)}, \ldots, d(\mathbf{z})_{(B-1)}],
\]
whose empirical log-spacing
\[
  \tau_i(\mathbf{z})
  = \ln\!\frac{d(\mathbf{z})_{(i+1)}}{d(\mathbf{z})_{(i)}}
\]
decays approximately as power of $i$ across datasets and feature layer.

\vspace{0.4cm}
\noindent
To present our theoretical analysis using Extreme Value Theory (EVT),
we first introduce the notion of similarity.
Let $\mathbf{z}$ be a fixed reference element, and let
\(Z\) denote the random variable representing the distance from
$\mathbf{z}$ to a randomly chosen normal element in $\mathcal{B}$, and we suppose that $Z > 0$ a.s.
The patch-to-patch ($\mathrm{p2p}$) similarity is defined as the reciprocal:
\[
  S = \frac{1}{Z}.
\]
For a random collection \(C\) with \(N\) elements
\(\{\mathbf{z}_1, \ldots, \mathbf{z}_N\}\),
the distance from $\mathbf{z}$ to \(C\) is
\[
  Y = \min_{p \in C} Z_p,
\]
where \(Z_p\) is the distance from $\mathbf{z}$ to element $\mathbf{z}_p \in C$.
The corresponding patch-to-image similarity is
\[
  S_{\mathrm{p2i}}
  = \frac{1}{Y}
  = \max \left(\frac{1}{Z_{1}}, \ldots, \frac{1}{Z_{N}}\right)
  = \max(S_{1}, \ldots, S_{N}),
\]
where \(S_i = 1/Z_i\) are the patch-to-patch similarities between $\mathbf{z}$ to $\mathbf{z}_i$.

\vspace{0.3cm}
\noindent
In our earlier work~\citep{CoDeGraph}, the empirical scaling law
was explained under two simplifying assumptions:
\begin{enumerate}[label=(\alph*)]
  \item The similarity variable \(S\) has a
        \emph{regularly varying tail}:
        \[
          \mathrm{P}(S > s) = s^{-\alpha}L(s),
          \qquad s \to \infty,
        \]
        where $\alpha > 0$ is the tail index and $L(s)$ is a slowly varying
        function.
  \item The random variables \(S_i\) are treated as
        \emph{independent and identically distributed (i.i.d.)}.
\end{enumerate}
Under these assumptions, EVT ensures that the maxima of $\{S_i\}$
converge to a Fréchet distribution, implying that
\(S_{\mathrm{p2i}}\) also exhibits a Fréchet-type tail with the same
index~\(\alpha\).
Consequently, the expected log-spacing of the ordered entries in the mutual similarity vector obeys
\[
  \mathbb{E}[\tau_i] = \frac{1}{\alpha i},
\]
which aligns with empirical scaling curves for normal elements.

\vspace{0.4cm}
\noindent
This chapter aims to generalize the argument by offering a deeper, less empirical explanation of the Similarity Scaling phenomenon. It does so under weaker, more realistic assumptions while providing a geometrical perspective on the phenomenon. The development proceeds along two directions:

\begin{enumerate}[label=(\alph*)]
  \item \textbf{Long-range independence.}
  We relax the i.i.d.\ assumption by adopting Leadbetter’s
  \(D(u_n)\) condition, which assumes long-range independence among similarities.

  \item \textbf{Geometric views.}
  We argue that the heavy-tailed nature of \(S\) arises naturally from the
  geometry of the feature space. Around a reference point $\mathbf{z}$,
  the number of similar elements contained in the ball \(B(\mathbf{z}, r)\) around $\mathbf{z}$ 
  scales with its radius as \(\mu(B(\mathbf{z}, r)) \propto r^{\alpha_0}\).
  Thus, elements that lie on or near a low-dimensional manifold embedded
  in \(\mathbb{R}^D\) produce inverse distances \(S = 1/Z\)
  whose tail decays as \(s^{-\alpha_0}\).

\end{enumerate}

\vspace{0.3cm}
\noindent
Together, these components establish the theoretical chain:
\begin{center}
\begin{tcolorbox}[colback=white,colframe=black!70!white,title=Local Geometry Implies Scale-Invariant Log-Spacing,width=0.9\textwidth]
\textit{
Local geometry
$\;\Rightarrow\;$
Regularly varying similarity tails
$\;\Rightarrow\;$
Fréchet-type extremes under $D(u_n)$
$\;\Rightarrow\;$
Scale-invariant log-spacing $\tau_i \sim 1/i$.
}
\end{tcolorbox}
\end{center}

\section{Similarity Scaling under Long-Range Independence}
\label{sec:long-range-independence}

This section generalizes the theoretical model for the Similarity Scaling Phenomenon by relaxing the strict i.i.d.\ assumption on patch similarities. We begin by assuming a regularly varying tail for the patch-to-patch similarity $S$, as established in Section~\ref{sec:theory-overview}. Under the i.i.d.\ assumption, Extreme Value Theory (EVT) implies that the patch-to-image similarity $S_{\mathrm{p2i}}$ follows a Fréchet distribution with the same tail index $\alpha$. We then introduce Leadbetter's $D(u_n)$ condition to account for long-range independence, which still ensures a Fréchet-type tail for $S_{\mathrm{p2i}}$. This then leads to the power-law decay of the similarity growth rate $\tau_i$.

\subsubsection*{Patch-to-Patch Similarity and Regularly Varying Tail}

Recall from Section~\ref{sec:theory-overview} that $\mathbf{z}$ is a fixed reference element, and $Z$ is the random variable representing the distance from $\mathbf{z}$ to a randomly chosen normal element. The patch-to-patch similarity is defined as
\[
S = \frac{1}{Z}.
\]
\begin{definition}[Slowly Varying Function]
A function $L:(0,\infty)\to(0,\infty)$ is \emph{slowly varying} (at infinity) if, for every $x>0$,
\[
\lim_{t\to\infty}\frac{L(tx)}{L(t)}=1.
\]
\end{definition}
\vspace{0.4cm}
Slowly varying functions grow or decay more slowly than any polynomial, and often appear as correction terms in heavy-tailed distributions.
\noindent
We maintain the core assumption of a regularly varying tail for $S$:
\begin{assumption}[Regularly Varying Tail for $S$] \label{ass:power-law}
The similarity $S$ is regularly varying with index $-\alpha$, i.e.,
\[
\mathrm{P}(S > s) = s^{-\alpha} L(s), \quad \text{as } s \to \infty,
\]
where $\alpha > 0$ is the tail index, and $L(s)$ is a slowly varying function.
\end{assumption}

\begin{figure}[h]
  \centering
  \begin{subfigure}[b]{0.48\textwidth}
    \centering
    \begin{tikzpicture}
      \node[inner sep=0pt] (image1) {\includegraphics[width=0.9\textwidth]{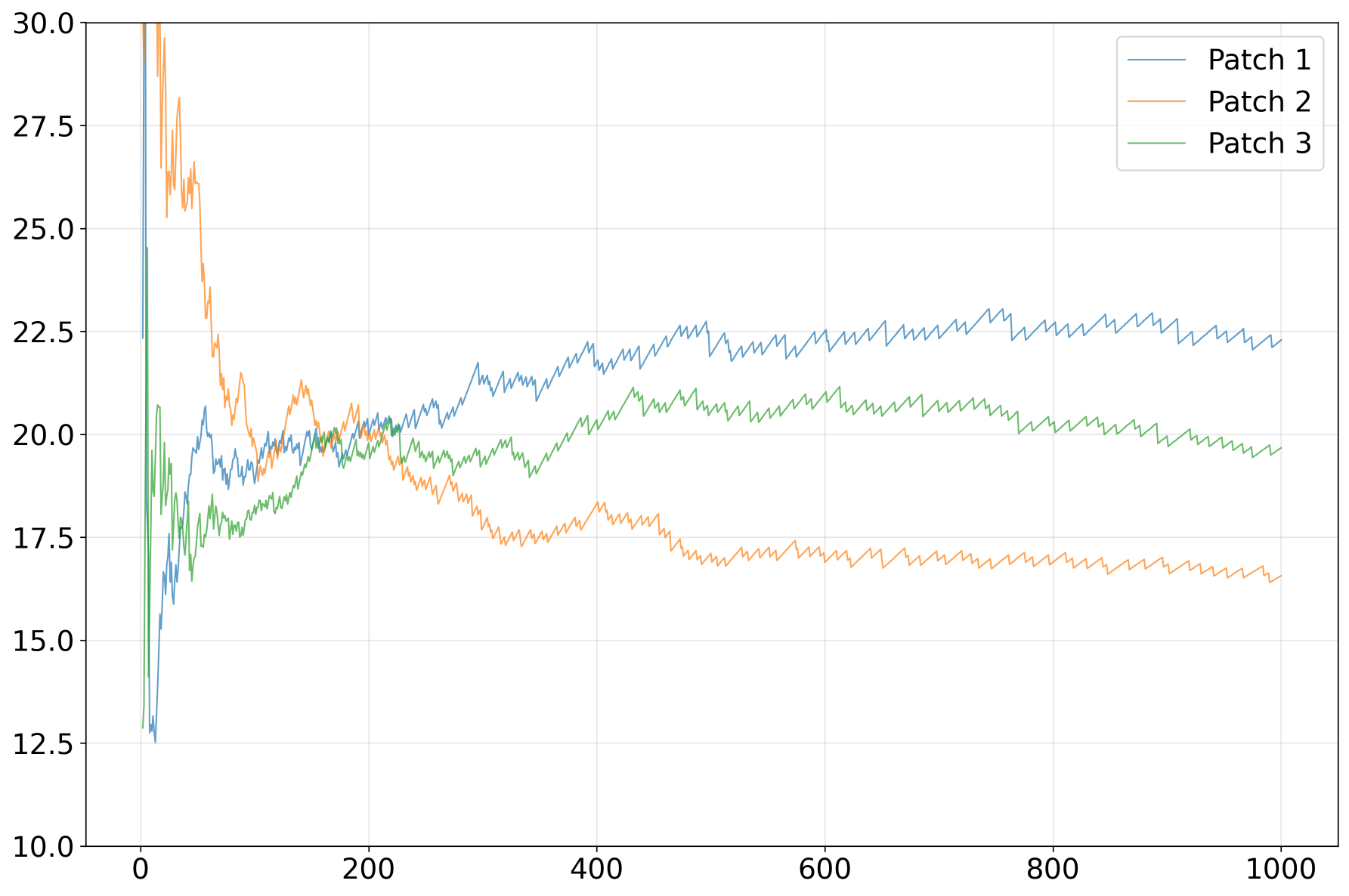}};
      \node[below=0.3cm of image1.south] {\small $k$ (order statistics)};
      \node[rotate=90, anchor=center] at ([xshift=-0.3cm,yshift=0.2cm]image1.west){\small $\hat{\alpha}_k^H$};
    \end{tikzpicture}
    \caption{Hill plots for screw dataset}
    \label{fig:hill-plot-screw}
  \end{subfigure}
  \hfill
  \begin{subfigure}[b]{0.48\textwidth}
    \centering
    \begin{tikzpicture}
      \node[inner sep=0pt] (image2) {\includegraphics[width=0.9\textwidth]{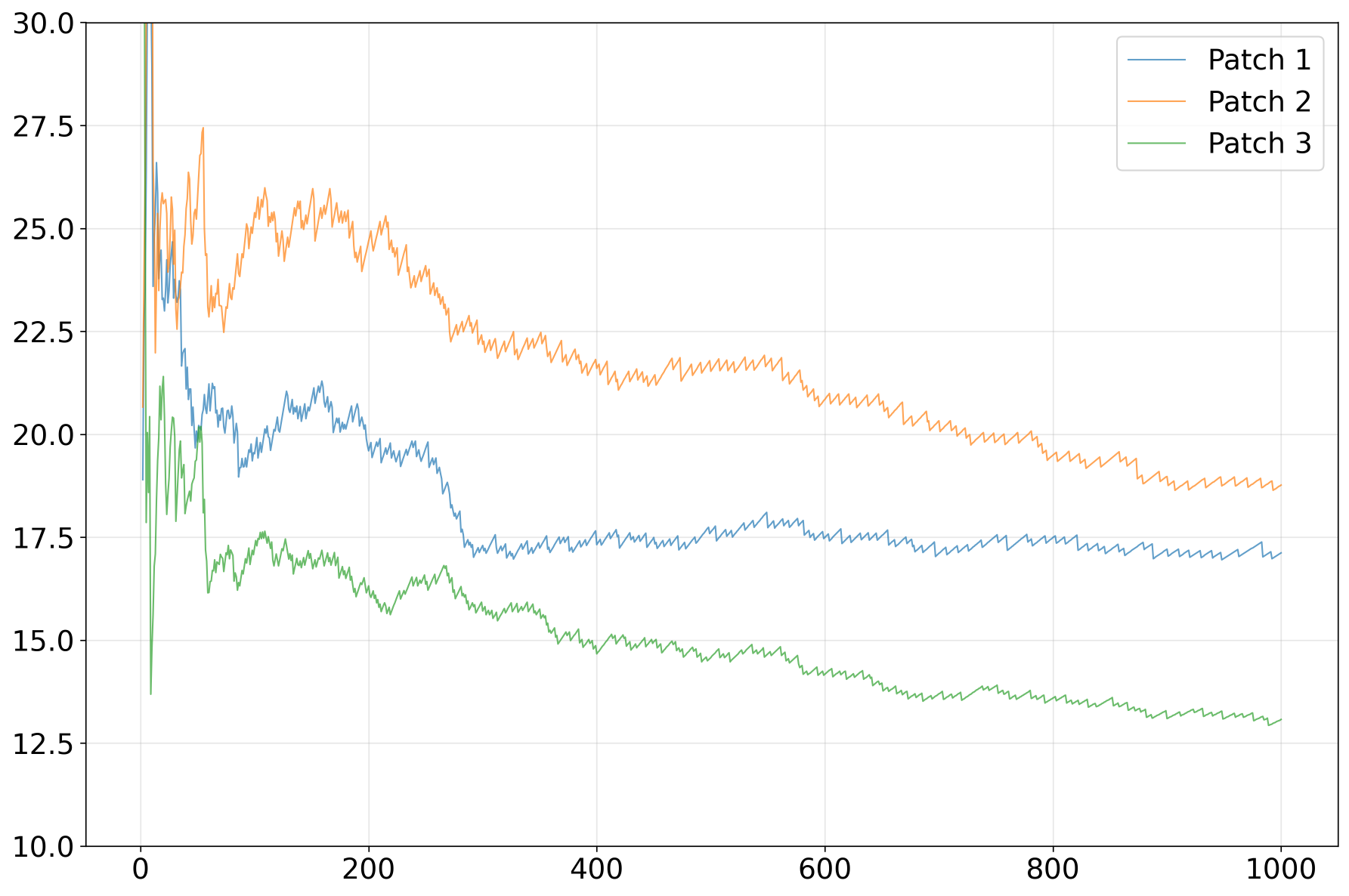}};
      \node[below=0.3cm of image2.south] {\small $k$ (order statistics)};
      \node[rotate=90, anchor=center] at ([xshift=-0.3cm,yshift=0.2cm]image2.west){\small $\hat{\alpha}_k^H$};
    \end{tikzpicture}
    \caption{Hill plots for cable dataset}
    \label{fig:hill-plot-cable}
  \end{subfigure}
  \caption{Hill plots showing the tail index estimation for patch-to-patch similarity distributions in industrial datasets. We randomly selected 3 patches from each dataset and computed distances from these patches to all patches in different images. Each plot shows 3 line plots corresponding to the 3 selected patches. The stable plateau in both plots indicates the presence of power-law tails, supporting our Assumption~\ref{ass:power-law} that the tail of the distribution of \( S \) approximately follows a power law.}
  \label{fig:hill-plot}
\end{figure}
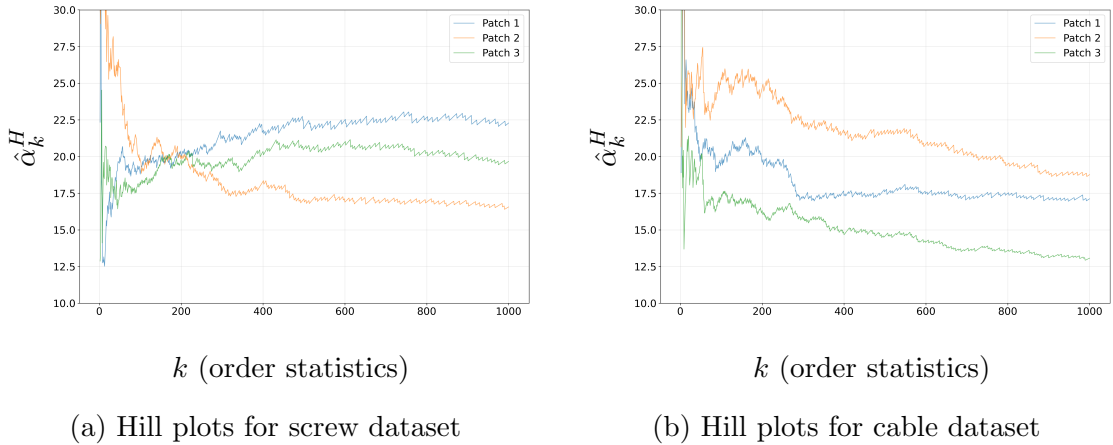

Empirical evidence supporting this assumption is given by Hill plots (Fig.~\ref{fig:hill-plot}), which show stable plateaus for the estimated tail index \(\hat{\alpha}_k^H\) across a wide range of order statistics. The existence of such plateaus provides significant evidence of the asymptotic power-law behavior of \(S\)~\citep{Nair_Wierman_Zwart_2022}.

\subsubsection*{Patch-to-Image Similarity via Extreme Value Theory}
For a collection \(C_j = \{\mathbf{z}_j^1, \ldots, \mathbf{z}_j^N\}\), the distance and the patch-to-image similarity from \(\mathbf{z}\) to \(C_j\) is defined as
\[
Y = \min_{\mathbf{z}_p \in C_j} Z_p,
\qquad
S_{\mathrm{p2i}} = \frac{1}{Y} = \max(S_1, \ldots, S_N).
\]

\vspace{4mm}
Under the i.i.d.\ assumption, the classical Fisher–Tippett–Gnedenko theorem~\citep{fisher1928limiting} states that for random variables with a regularly varying tail of index \(\alpha\), the normalized maximum converges to a Fréchet distribution:
\[
\mathrm{P}\left( \frac{\max(S_1, \ldots, S_N) - b_N}{a_N} > x \right) \to \Phi_\alpha(x) = \exp(-x^{-\alpha}), \quad x > 0,
\]
for suitable normalizing sequences $a_N > 0$ and $b_N$. Consequently,
\[
\mathrm{P}(S_{\mathrm{p2i}} > s) \sim s^{-\alpha}, \quad \text{as } s \to \infty.
\]

\subsubsection*{Leadbetter's $D(u_n)$ Condition}
\label{subsec:Du-intro-new}
The independence assumption is overly simplistic for image data, where neighboring patches are spatially correlated. To account for such dependence, we employ Leadbetter’s \(D(u_n)\) condition~\citep{leadbetter1982extremes, embrechts2013modelling}. First, let us recall the definition of a strictly stationary sequence \citep{embrechts2013modelling, shumway2006time}.
\begin{definition}
    A sequence of random variables $\left(X_n\right)$ is \emph{strictly stationary} if its finite-dimensional distributions are invariant under shifts of time, i.e.,
    \[
        \left(X_{t_1}, \ldots, X_{t_m}\right) \stackrel{d}{=} \left(X_{t_1+h}, \ldots, X_{t_m+h} \right)
    \] 
    for any choice of indices $t_1 < \ldots < t_m$ and integer $h$.
\end{definition}
\vspace{4mm}
We now introduce Leadbetter's \(D(u_n)\) \citep{leadbetter1982extremes} condition in the form presented in \citep{embrechts2013modelling}. Let \(\{S_i\}_{i\ge 1}\) be a strictly stationary sequence,
and let \(\{u_n\}\) be a sequence of threshold levels depending on \(n\).
\begin{condition}[$D(u_n)$]
\label{cond:Du}
For any integers \(p,q\) and \(n\), consider indices
\(1 \le i_1 < \cdots < i_p < j_1 < \cdots < j_q \le n\) such that
\(j_1 - i_p \ge \ell\). Define index sets
\(A_1 = \{i_1,\dots,i_p\}\) and \(A_2 = \{j_1,\dots,j_q\}\).
We say that the sequence \((S_i)\) satisfies condition \(D(u_n)\) if
there exists a sequence \(\alpha_{n,\ell} \to 0\) as \(n\to\infty\) for
some \(\ell = \ell_n = o(n)\) such that
\[
  \Bigl|
    \mathrm{P}\Bigl(
      \max_{i \in A_1 \cup A_2} S_i \le u_n
    \Bigr)
    - \mathrm{P}\Bigl(
        \max_{i \in A_1} S_i \le u_n
      \Bigr)
      \mathrm{P}\Bigl(
        \max_{i \in A_2} S_i \le u_n
      \Bigr)
  \Bigr|
  \le \alpha_{n,\ell}.
\]
\end{condition}
\vspace{4mm}
 The condidtion $D(u_n)$ can be thoughts of a ``long-range" indenpendence in the sequence $(S_i)$. Under Assumption~\ref{ass:power-law} and condition~$D(u_n)$, it follows from Leadbetter’s results~\citep{leadbetter1982extremes} that the limiting distribution of the maximum remains Fréchet, but with a modified tail index:
\[
\mathrm{P}(S_{\mathrm{p2i}} > s) \sim s^{-\alpha_0}, \qquad
\alpha_0 = \theta\alpha, \quad 0 < \theta \le 1.
\]
Here, \(\theta\) is the \emph{extremal index}, measuring the effective reduction in independent extremes due to local dependence.  
Hence, the heavy-tailed nature of \(S_{\mathrm{p2i}}\) is preserved even under restricted long-range dependence—an assumption well-suited to image patches where nearby regions are correlated but distant ones behave asymptotically independently.

\subsubsection*{Normalized Distance and Beta Distribution}

The regularly varying tail of \(S_{\mathrm{p2i}}\) with tail index \(\alpha_0\) implies the following small-value behavior for the distance \(Y = 1/S_{\mathrm{p2i}}\):
\[
\mathrm{P}(Y \le y) = \mathrm{P}(S_{\mathrm{p2i}} \ge 1/y)
\approx C y^{\alpha_0}, \qquad y \to 0.
\]
To model this, we introduce a normalized distance index.

\begin{definition}[Normalized Distance Index]
Let \(s_0\) denote a scale threshold within the power-law region of \(Y\).  
Define the normalized index
\[
Z = \frac{Y}{s_0} \,\Big|\, Y \le s_0.
\]
\end{definition}
The cumulative distribution function of $Z$ is $F_Z(z) = z^{\alpha_0}$ for $z \in [0,1]$, corresponding to a $\mathrm{Beta}(\alpha_0, 1)$ distribution with density $f_Z(z) = \alpha_0 z^{\alpha_0 - 1}$. This leads us to the next assumption.

\begin{assumption}[Beta Distribution] \label{ass:beta-model}
The normalized patch-to-image distance $Z$ follows a $\mathrm{Beta}(\alpha_0, 1)$ distribution.
\end{assumption}

\begin{figure}[p]
\centering
\begin{subfigure}[b]{0.48\textwidth}
    \centering
    \includegraphics[width=2.5in, height=2in]{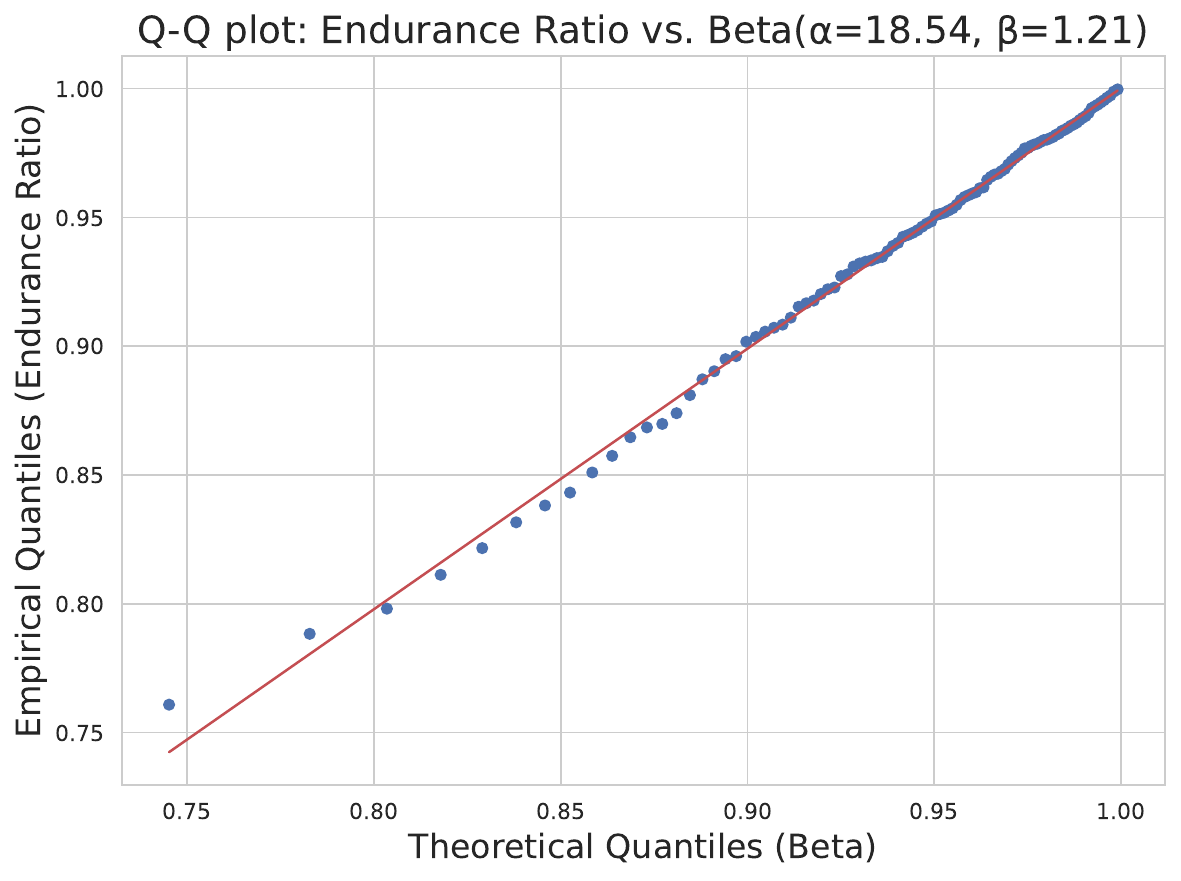}
    \caption{\texttt{Breakfast\_box} (MVTec LOCO)}
    \label{fig:qq_breakfast_box}
\end{subfigure}
\hfill
\begin{subfigure}[b]{0.48\textwidth}
    \centering
    \includegraphics[width=2.5in, height=2in]{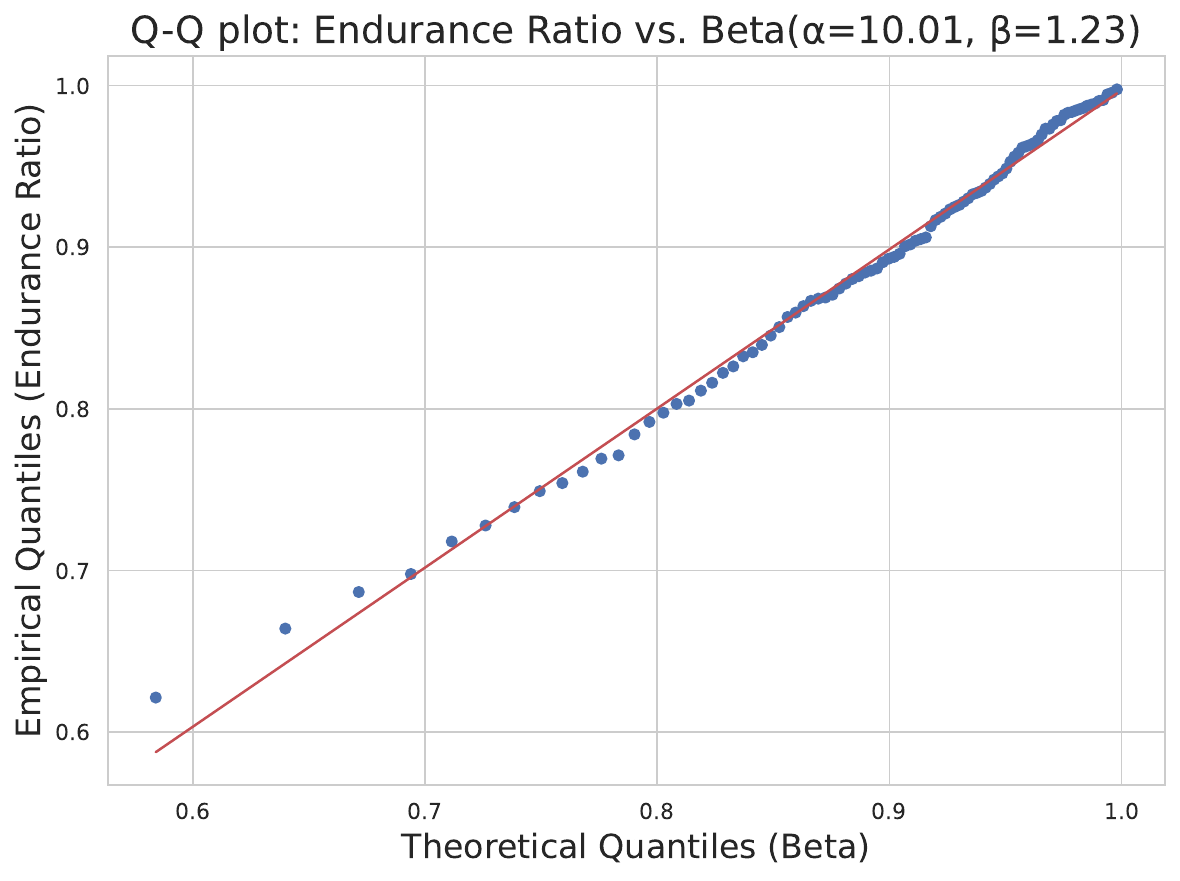}
    \caption{\texttt{Screw} (MVTec AD)}
    \label{fig:qq_screw}
\end{subfigure}

\vspace{0.3cm}

\begin{subfigure}[b]{0.48\textwidth}
    \centering
    \includegraphics[width=2.5in, height=2in]{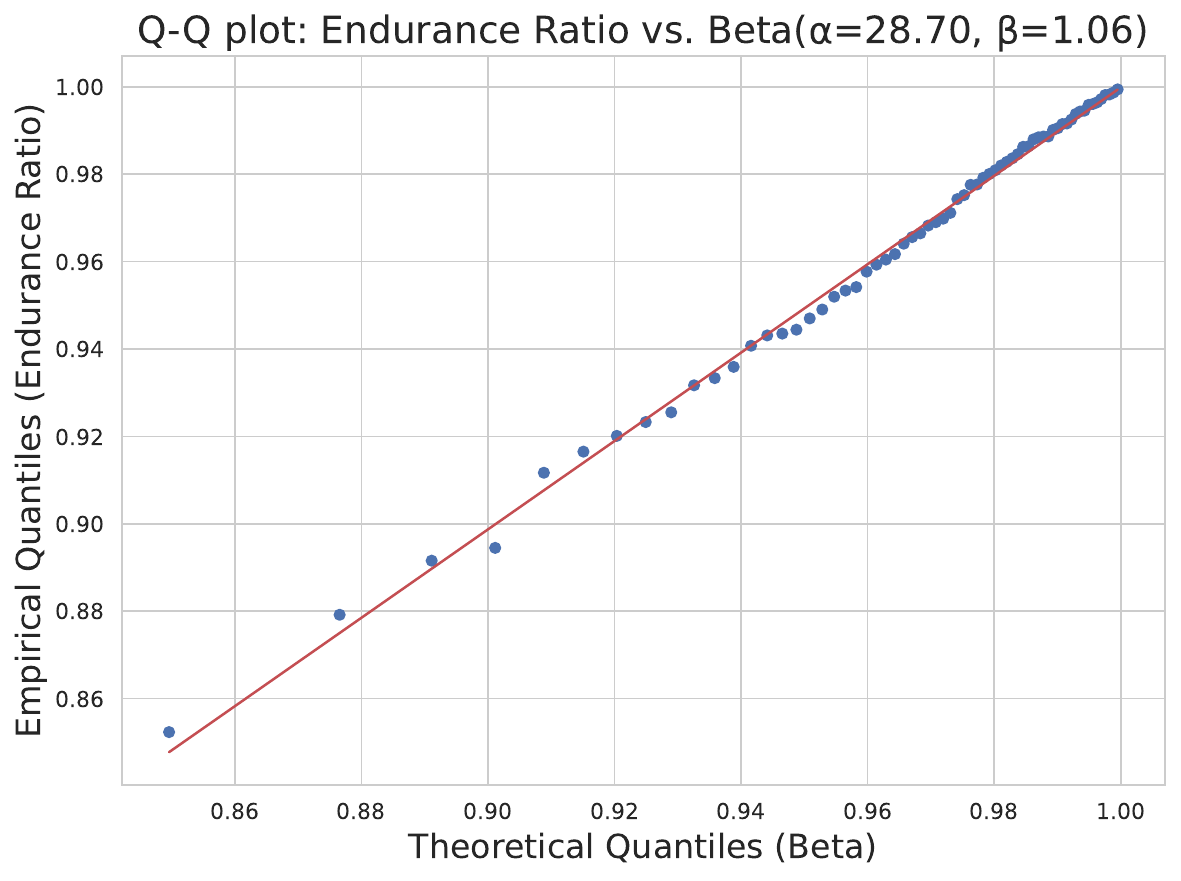}
    \caption{\texttt{Metal\_nut} (MVTec AD)}
    \label{fig:qq_metal_nut}
\end{subfigure}
\hfill
\begin{subfigure}[b]{0.48\textwidth}
    \centering
    \includegraphics[width=2.5in, height=2in]{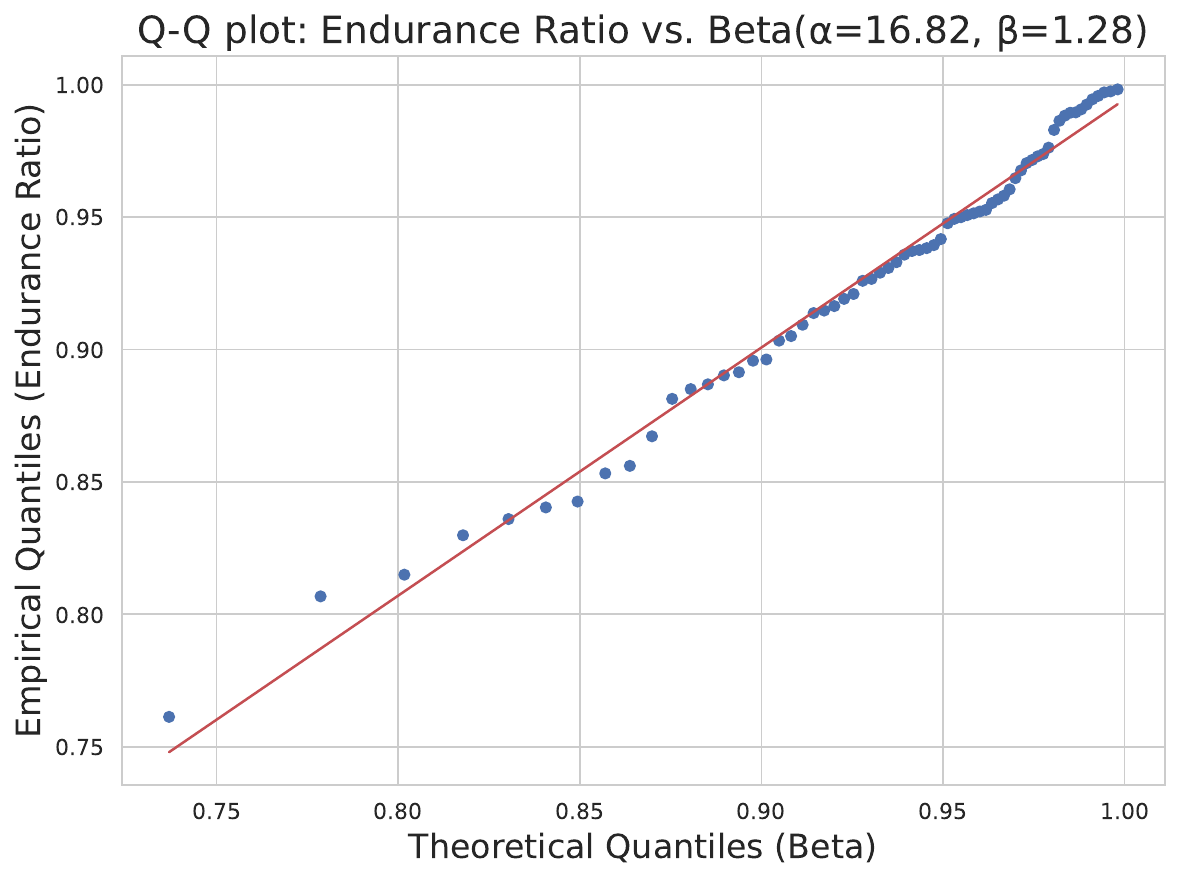}
    \caption{\texttt{Cable} (MVTec AD)}
    \label{fig:qq_cable}
\end{subfigure}
\caption{Q-Q plots validating the Beta distribution assumption for similarity indices across four normal industrial objects, including objects with inconsistent normal patterns (\texttt{screw}, \texttt{breakfast\_box}). For each object, we fixed one randomly selected patch and computed similarity indices from this patch to all patches in different images. Each plot compares the empirical quantiles of $Y_{(i)}/Y_{(\omega)}$ with $\omega=0.3N$ against theoretical quantiles from its fitted Beta distribution. The close alignment along the diagonal demonstrates that our modeling assumption is reasonable.}
\label{fig:qq_plots_validation}
\end{figure}

Empirical validation across several industrial datasets confirms this model.  
We evaluate the endurance ratio \(Y_{(i)}/Y_{(\omega)}\) for \(i < \omega\),
where \(\omega \approx 0.3N\) to make sure that it lies within the power-law region $(0, s_0)$.
Q-Q plots (Fig.~\ref{fig:qq_plots_validation}) show that the empirical
quantiles align closely with the fitted
\(\mathrm{Beta}(\alpha_0,\beta)\) distribution
(\(\beta \approx 1\), \(\alpha_0 \gg \beta\)).

\subsubsection*{Similarity Growth Rate and Log-Spacing}

Let $Y_{(1)} \leq \cdots \leq Y_{(\omega)}$ be the order statistics of $\omega$ i.i.d.\ samples of $Y \mid Y < s_0$, so $Y_{(i)} = s_0 Z_{(i)}$ where $Z_{(i)}$ are order statistics from $\mathrm{Beta}(\alpha_0, 1)$. The similarity growth rate is
\[
\tau_i = \ln \frac{Y_{(i+1)}}{Y_{(i)}}.
\]

The statistical properties of $\tau_i$ follow directly from the results established in our earlier work~\citep{CoDeGraph}.

\begin{theorem}[Log-Spacing of Order Statistics] \label{thm:log-spacing}
The log-spacings $\ln Y_{(i+1)} - \ln Y_{(i)}$ are independent and follow $\mathrm{Exp}(\alpha_0 i)$ for $i = 1, \ldots, \omega-1$.
\end{theorem}

\begin{proof}
See Theorem~3 in \citep{CoDeGraph}, replacing the tail index $\alpha$ with the effective index $\alpha_0$.
\end{proof}

\noindent
This yields Similarity Scaling Law for Mutual Similarity Vector:
\begin{corollary}[Similarity Scaling Law] \label{cor:growth-rate}
Similarity mutual vector exhibit power-law decay in growth rate:
\[
\mathbb{E}[\tau_i] = \frac{1}{\alpha_0 i}, \quad \mathrm{Var}[\tau_i] = \frac{1}{(\alpha_0 i)^2}.
\]
\end{corollary}

\begin{remark}
This framework generalizes the i.i.d.\ case by incorporating dependence via $D(u_n)$, explaining the robust power-law decay $\tau_i \sim 1/i$ under realistic conditions.
\end{remark}
\newpage
\section{Geometric Intuition and Local Scaling of Similarity}
\label{sec:geom-intuition}

A central assumption in our analysis is that the patch-to-patch similarity
\[
  S = \frac{1}{Z},
\]
where $Z$ is the distance from a fixed reference element $\mathbf{z}$ to a
random normal element, has a regularly varying (heavy-tailed) distribution:
\[
  \mathrm{P}(S > s) = s^{-\alpha} L(s), \qquad s \to \infty,
\]
for some $\alpha > 0$ and slowly varying function $L$.
This assumption is empirically supported by Hill plots
(Fig.~\ref{fig:hill-plot}), which exhibit stable plateaus over wide ranges
of order statistics. Nevertheless, a natural question remains:

\begin{center}
\emph{Why should the similarity $S$ behave like a heavy-tailed random variable?}
\end{center}

In this section, we develop geometric intuition for this behavior.
We argue that when normal patches around $\mathbf{z}$ are distributed
in a locally low-dimensional manner, the number of similar patches within
a small radius grows like a power of the radius. This local growth
directly leads to a power-law tail for the similarity.

\subsection{Local manifold geometry and power-law similarity}
\label{subsec:geom-manifold}

Let $\mathbf{z}$ be a fixed reference element in the feature space
$\mathbb{R}^D$, and let $\mu$ denote the distribution of normal patches.
Under the \emph{manifold hypothesis}, normal patches are assumed to concentrate near a lower-dimensional manifold $\mathcal{M} \subset \mathbb{R}^D$ with intrinsic dimension $d \ll D$.  
In a small neighborhood of $\mathbf{z}$ on $\mathcal{M}$, we may think of
similar patches as forming a cloud of points around $\mathbf{z}$.
The key geometric quantity is the growth of the number of points inside
a ball centered at $\mathbf{z}$.

For a radius $r>0$, consider the ball
\[
  B(\mathbf{z}, r)
  = \{\mathbf{x} \in \mathbb{R}^D : \|\mathbf{x} - \mathbf{z}\| \le r\}.
\]
The geometric intuition is that, for sufficiently small $r$, the number
of normal patches lying in $B(\mathbf{z}, r)$ grows proportionally to
the $d$-dimensional volume of $B(\mathbf{z}, r) \cap \mathcal{M}$:
\[
  \#\{\text{patches in } B(\mathbf{z}, r)\}
  \;\propto\; r^{d}.
\]
In probabilistic terms, if $Z$ denotes the distance from $\mathbf{z}$ to
a random normal patch, this heuristic growth translates into
\[
  \mathrm{P}(Z \le r)
  \approx c(\mathbf{z})\, r^{d}, \qquad r \downarrow 0,
\]
for some local constant $c(\mathbf{z})>0$ that reflects the local density
of patches around~$\mathbf{z}$.

The similarity is defined as $S = 1/Z$, thus
\[
  \mathrm{P}(S > s)
  = \mathrm{P}\!\left(Z < \frac{1}{s}\right)
  \approx c(\mathbf{z})\,
          \left(\frac{1}{s}\right)^{d}
  = c(\mathbf{z})\, s^{-d}, \qquad s \to \infty.
\]
Thus, the heavy-tailed behavior of $S$ appears as a direct reflection of
the local geometry of normal patches: a cloud of patches whose count in
$B(\mathbf{z}, r)$ grows like $r^{d}$ leads to a similarity distribution
whose tail decays like $s^{-d}$.
The exponent $d$ can be interpreted as the intrinsic dimension of the
normal patch manifold around $\mathbf{z}$.

\subsection{Toy model: Poisson-distributed patches in a compact domain}
\label{subsec:toy-poisson-model}

To illustrate this geometric intuition, we
consider a simple random model in which similar normal patches are imagined as particles
scattered in a compact domain
$\mathcal{D} \subset \mathbb{R}^{d}$.
A standard model for such random scattering is the \emph{homogeneous
Poisson point process} on $\mathcal{D}$ with intensity $\lambda>0$. Informally, a homogeneous Poisson point process with intensity $\lambda$
is characterized by two properties:
\begin{itemize}
  \item For any bounded Borel set $B \subset \mathcal{D}$, the number of
        points $N(B)$ in $B$ follows a Poisson distribution with mean
        $\lambda\,\mathrm{Vol}(B)$.
  \item For disjoint sets $B_1,\dots,B_k \subset \mathcal{D}$, the random
        variables $N(B_1),\dots,N(B_k)$ are independent.
\end{itemize}
The first property shows that the \emph{expected} number of points inside
a ball $B(\mathbf{z}, r)$ is
\[
  \mathbb{E}[N(B(\mathbf{z}, r))]
  = \lambda\, \mathrm{Vol}(B(\mathbf{z}, r))
  = \lambda \nu_d\, r^{d},
\]
where $\nu_d$ is the volume of the unit ball in $\mathbb{R}^d$.
Hence, for small $r$, the point count in $B(\mathbf{z}, r)$ grows
in expectation like $r^{d}$, exactly matching the geometric picture
described above.

In our simulations, we generate a realization of a homogeneous Poisson
process in $\mathcal{D}$, fix a reference point $\mathbf{z}$ in the
interior, and compute the distances $Z_1, Z_2, \dots$ from $\mathbf{z}$
to all points in the realization.
We then convert these distances to similarities $S_i = 1/Z_i$, sort them
in descending order $S_{(1)} > S_{(2)} > \ldots > S_{n}$, and estimate the tail index using the Hill estimator
\[
  \widehat{\alpha}_k^{\,H}
  = \left(
      \frac{1}{k}\sum_{i=1}^{k} \ln S_{(i)}
      - \ln S_{(k+1)}
    \right)^{-1},
\]
As the number of simulated points increases, the Hill plots
$k \mapsto \widehat{\alpha}_k^{\,H}$ exhibit a plateau close to $d$,
indicating that the empirical tail of the similarity distribution behaves
like a power law with exponent $d$.
This observation is consistent with our geometric intuition and mirrors
the behavior of Hill plots obtained from real patch similarities.
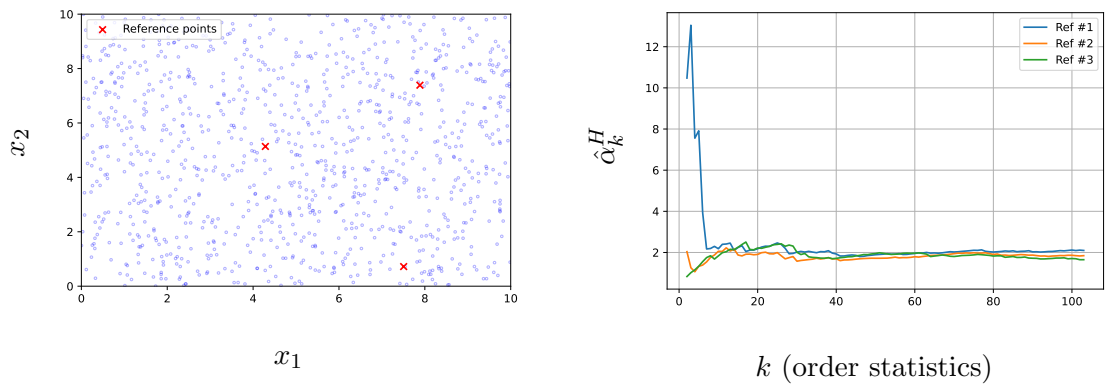
\begin{figure}[t]
  \centering
  % Left: Poisson point simulation
  \begin{subfigure}[b]{0.48\textwidth}
    \centering
    \begin{tikzpicture}
      \node[inner sep=0pt] (poisson) {\includegraphics[width=0.9\textwidth]{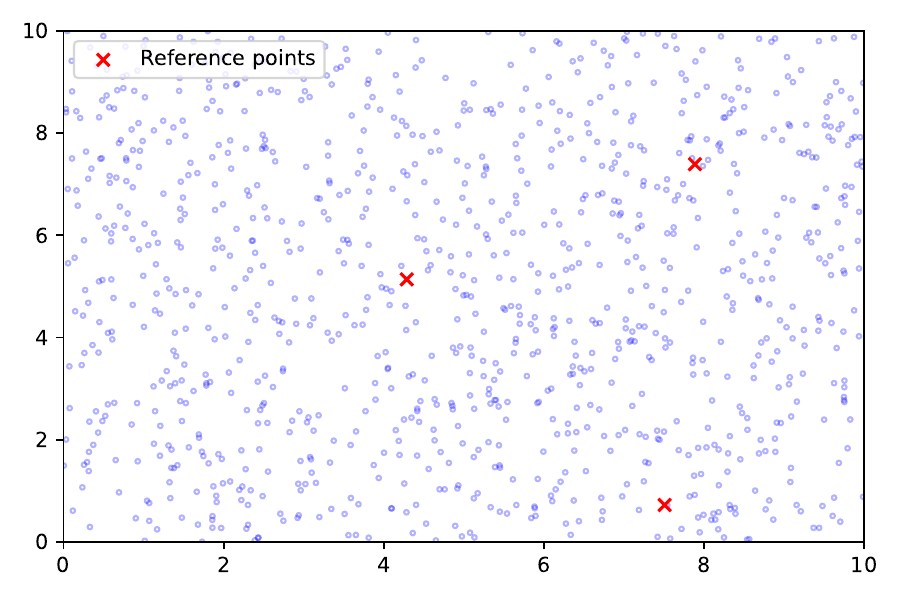}};
      \node[below=0.3cm of poisson.south] {\small $x_1$};
      \node[rotate=90, anchor=center] at ([xshift=-0.35cm,yshift=0.2cm]poisson.west){\small $x_2$};
    \end{tikzpicture}
    \caption{Homogeneous Poisson points.}
    \label{fig:poisson-sim}
    \vspace{2mm}
  \end{subfigure}
  \hfill
  % Right: Hill plot
  \begin{subfigure}[b]{0.48\textwidth}
    \centering
    \begin{tikzpicture}
      \node[inner sep=0pt] (hill) {\includegraphics[width=0.9\textwidth]{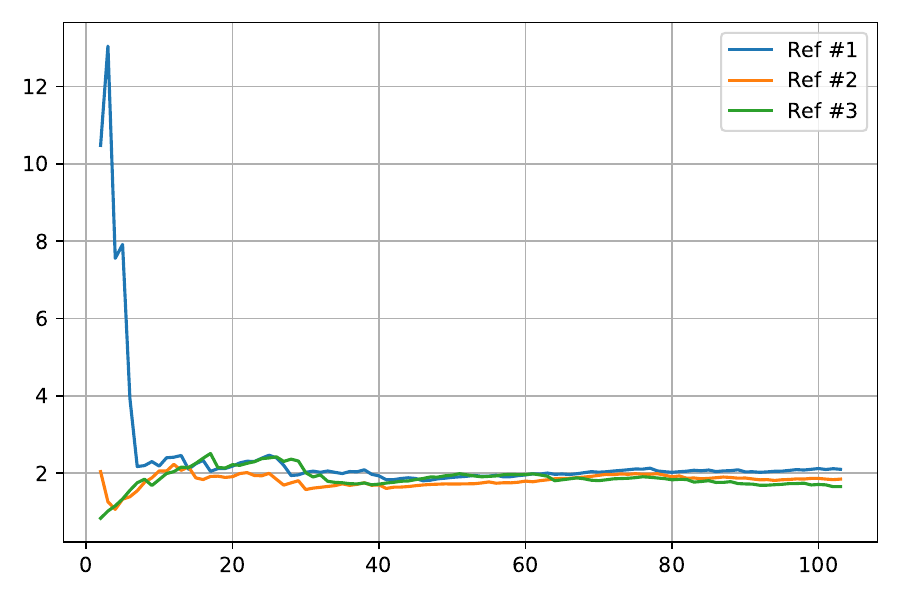}};
      \node[below=0.3cm of hill.south] {\small $k$ (order statistics)};
      \node[rotate=90, anchor=center] at ([xshift=-0.35cm,yshift=0.2cm]hill.west){\small $\hat{\alpha}_k^H$};
    \end{tikzpicture}
    \caption{Hill plots from three reference points.}
    \label{fig:poisson-hill}
  \end{subfigure}
  \caption{Poisson point simulation (about 10 points per unit square in a $10\times 10$ domain) and the corresponding Hill plots of the inverse distances from three reference points.}
  \label{fig:poisson-sim-hill}
\end{figure}
\subsection{Local ball asymptotics and regular variation}
\label{sec:local-density-to-rv-main}

In this section we formalize the geometric intuition we describe above.
We show that, under mild regularity assumptions on the distribution of
normal patches, the mass of small balls around a reference element
$\mathbf{z}$ grows like $r^d$, and that this implies a regularly varying
tail for the similarity $S = 1/Z$.

\begin{theorem}
\label{thm:local-ball-asymptotics}
Let $\mu$ be the distribution of normal patches in $\mathbb{R}^d$, assumed
absolutely continuous with respect to Lebesgue measure, with density
$f \in L^1_{\mathrm{loc}}(\mathbb{R}^d)$.
Let
\[
  B(\mathbf{z}, r)
  = \{x \in \mathbb{R}^d : \|x - \mathbf{z}\| \le r\}
\]
denote the closed ball of radius $r$ around $\mathbf{z}$, and let $\nu_d$ is the volume of unit ball in $\mathbb{R}^d$.
Then for almost every $\mathbf{z} \in \mathbb{R}^d$,
\[
  \lim_{r \downarrow 0}
  \frac{\mu\bigl(B(\mathbf{z}, r)\bigr)}{r^d}
  = f(\mathbf{z})\, v_d.
\]
\end{theorem}

\noindent
The proof is a direct consequence of the Lebesgue Differentiation Theorem~\cite{stein2009real}. In particular, this theorem tells use that the mass of a small ball around $\mathbf{z}$ grows
asymptotically like $r^d$, with proportionality constant
$c(\mathbf{z}) = f(\mathbf{z}) v_d$.

\begin{theorem}[Regular variation of patch-to-patch similarity]
\label{thm:rv-similarity-main}
Let $\mu$ and $f$ be as in
Theorem~\ref{thm:local-ball-asymptotics}, and fix a point $\mathbf{z}$
such that $f(\mathbf{z})>0$ and the conclusion of
Theorem~\ref{thm:local-ball-asymptotics} holds.
Let $X \sim \mu$, define the distance
\[
  Z := \|X - \mathbf{z}\|,
\]
and the patch-to-patch similarity
\[
  S:= \frac{1}{Z}.
\]
Then the tail distribution of $S$ is regularly varying with tail
index $d$, i.e.\ for every $x>0$,
\[
  \lim_{t \to \infty}
  \frac{\mathbb{P}\bigl(S > t x\bigr)}%
       {\mathbb{P}\bigl(S > t\bigr)}
  = x^{-d}.
\]
Equivalently, there exists a slowly varying function $L$ such that
\[
  \mathbb{P}(S > s)
  = s^{-d} L(s), \qquad s \to \infty.
\]
\end{theorem}

\noindent
Thus, under mild local regularity of the normal-patch distribution,
the similarity $S$ is heavy-tailed with tail index $d$,
which coincides with the intrinsic dimension in the geometric intuition
of Section~\ref{sec:geom-intuition}.
Detailed proofs are provided in
Appendix~\ref{app:theorem}.
\subsection*{Remark on the heavy-tail assumption and its scope}
\label{subsec:remark-heavy-tail-scope}

We briefly comment on the choice of a heavy-tailed similarity distribution and on when the Similarity Scaling Phenomenon is expected to appear in practice.

\paragraph{Why assume a heavy-tailed similarity?}
Theorems~\ref{thm:local-ball-asymptotics} and
\ref{thm:rv-similarity-main} show that, for any ``well-behaved" distribution $\mu$ on $\mathbb{R}^d$, and for almost
every reference element $\mathbf{z}$ with $f(\mathbf{z})>0$, the
similarity
\[
  S = \frac{1}{\|X-\mathbf{z}\|}, \qquad X \sim \mu,
\]
has a regularly varying (hence heavy-tailed) tail with exponent~$d$.
In this sense, a heavy-tailed model for $S$ is not an additional ad hoc
assumption: it follows from mild local regularity of the normal-patch
distribution and matches the geometric picture of patches populating a
$d$-dimensional manifold around~$\mathbf{z}$.

\paragraph{Why the scaling may not appear in general datasets.}
The theoretical heavy-tail behavior of $S$ is asymptotic: it describes
$\mathbb{P}(S > s)$ for very large $s$, i.e.\ for extremely similar
patches (extremely small distances to $\mathbf{z}$).
In a finite dataset, we only observe the largest similarity values that
actually occur among the sampled patches.
For the industrial image datasets we consider, normal structures are
highly redundant, so for a typical normal patch $\mathbf{z}$ there exist
many other patches that are very similar to $\mathbf{z}$ in feature
space. As a result, the dataset already samples the high-similarity
regime where the power-law tail of $S$ becomes visible, and the
Similarity Scaling phenomenon can be observed.

In contrast, for a more general dataset with highly diverse normal
patterns, a patch that is considered “normal” relative to $\mathbf{z}$
may lie much farther away in feature space in the sense relevant to
industrial inspection.
Then the extremely similar patches that control the tail of $S$ may be
so rare that they are never observed unless we sample an enormous (in
the limit, infinite) number of images.
In such cases, the asymptotic heavy-tail behavior of $S$ still holds
theoretically under our assumptions, but the finite dataset may never
reach this regime, and the empirical Similarity Scaling Phenomenon may
not appear clearly.
\section{Summary}
\label{sec:theory-summary}

This chapter extended the theoretical explanation of the Similarity
Scaling Phenomenon proposed in our earlier work.
While the original paper relied on two simplifying assumptions—i.i.d.\
similarities and a postulated heavy-tailed distribution—this chapter
established a more general and rigorous framework.

First, we replaced the independence assumption by Leadbetter’s
$D(u_n)$ condition, which allows for long-range dependence between
patch similarities.
Under this weaker assumption, the limiting distribution of the
patch-to-image similarity $S_{\mathrm{p2i}}$ remains Fréchet, but with an
effective tail index $\alpha_0 = \theta\alpha$, showing that the
scale-invariant decay $\mathbb{E}[\tau_i] \sim 1/i$ persists even under
dependence.

Second, we explained the geometric origin of the heavy-tailed similarity
distribution.  
By linking the small-ball mass
$\mu(B(\mathbf{z},r)) \propto r^{d}$ to local manifold geometry, and
formalizing it through the Lebesgue Differentiation Theorem, we proved
that the similarity $S = 1/\|X-\mathbf{z}\|$ must have a regularly
varying tail with tail index $d$.  
Hence, the heavy-tail behavior is not an assumption but a geometric
consequence of locally uniform patch distributions.
\begin{center}
\begin{tcolorbox}[colback=white,colframe=black!70!white,
title=Theoretical Chain Established in This Chapter,
width=0.9\textwidth]
\centering
\textbf{Local manifold geometry} \\[2pt]
$\Downarrow$ \\[2pt]
\textbf{Small-ball mass law} $\;\; \mu(B(\mathbf{z},r)) \propto r^{d}$ \\[2pt]
$\Downarrow$ \\[2pt]
\textbf{Regularly varying similarity tail} $\;\; \mathbb{P}(S > s) \propto s^{-d}$ \\[2pt]
$\Downarrow$ \\[2pt]
\textbf{Fréchet-type extremes under $D(u_n)$ with tail index $\alpha_0=\theta d$} \\[2pt]
$\Downarrow$ \\[2pt]
\textbf{Similarity Scaling Law} $\;\; \mathbb{E}[\tau_i] \sim 1/i$
\end{tcolorbox}
\end{center}
In summary, this chapter transforms the empirical findings of our
original paper into a theoretical framework that connects local
geometry, heavy-tailed similarity, and scale-invariant growth dynamics
under realistic dependence conditions.

\chapter{Extension of Batch-Based Methods to 3D MRI Volumes}
\label{chap:3d-extension}

In the previous chapters, we introduced a batch-based framework for anomaly detection and segmentation on 2D images, exemplified by MuSc and our graph-based CoDeGraph. 
This framework treats each image as a \emph{collection} of element-level feature vectors (patch tokens) and builds reasoning through mutual similarity relations between elements across collections. 
The theoretical principles established earlier—such as the Similarity Scaling Phenomenon and the neighbor-burnout effect—characterize statistical regularities in similarity sequences and are not restricted to any particular input dimensionality.  

The present chapter extends this batch-based formulation from 2D to 3D data domains, demonstrating that the same mathematical machinery applies naturally to volumetric representations. 
We select brain MRI as a representative testbed, as it provides a structured 3D modality with clear spatial organization and clinically meaningful anomalies.

\medskip

The chapter is organized as follows.  
Section~\ref{sec:3d-motivation} describes the motivation and scope of extending batch-based zero-shot anomaly detection to volumetric data.  
Section~\ref{sec:3dpatch-extraction} introduces the proposed method for extracting and aggregating 3D patch tokens from pretrained 2D foundation models.  
Section~\ref{sec:3d-experiments} presents experiments on brain MRI datasets. Finally, Section~\ref{sec:mri3d-discussion} provides discussion and outlines future directions for extending this framework to broader 3D modalities and medical imaging tasks.
% ============================================================
\section{Motivation and Overview}
\label{sec:3d-motivation}
The primary objective of this chapter is to establish a general procedure for adapting any batch-based method to volumetric data, with 3D MRI serving as a proof of concept. 
Rather than redesigning the algorithmic components of MuSc or CoDeGraph, we show that the extension requires only one modality-specific adaptation: the construction of volumetric element tokens. 
Once a 3D MRI volume \(V_i\) is converted into a sequence of 3D-patch tokens,
\[
V_i \;\longmapsto\; \big\{ \mathbf{z}_i^1, \mathbf{z}_i^2, \dots, \mathbf{z}_i^M \big\},
\]
all subsequent batch-based steps—mutual similarity computation, similarity scaling, community detection, and refinement—remain unchanged.

Accordingly, this chapter serves two purposes. 
First, it demonstrates that the batch-based reasoning paradigm is intrinsically dimensionality-agnostic, relying only on the existence of element-level features. 
Second, it proposes a practical, training-free pipeline for constructing such features from volumetric MRI data, using only pretrained 2D foundation models.

\subsubsection{Challenges and Design Principles}
\label{subsec:3d-challenges-design}

Despite rapid progress in volumetric representation learning~\citep{li2024abdomenatlas,wu2410large,blankemeier2024merlin}, current 3D foundation models remain limited in scale and diversity compared to 2D backbones such as DINOv2. 
Training high-capacity 3D transformers requires extensive computational resources and large, well-curated datasets, which are rarely available in medical imaging. 
Consequently, most existing 3D architectures underperform their 2D counterparts and are less transferable for zero-shot inference.  

A common workaround processes MRI volumes slice-by-slice using 2D models and concatenates features along the depth axis~\citep{marzullo2024exploring}. 
While this approach provides a rough volumetric representation, it discards genuine 3D spatial coherence and results in prohibitively large feature tensors. 
For instance, applying DINOv2-L/14 to a \(224^3\) volume across three axes yields roughly 170 million feature values making large-scale similarity computation infeasible.

To ensure compatibility with batch-based algorithms while preserving spatial fidelity, the 3D feature extraction must satisfy the following principles:

\begin{enumerate}[label=(\roman*)]
    \item \textbf{Locality.} Each token encodes information from a spatially cubic region, supporting voxel-level anomaly segmentation.  
    \item \textbf{Spatial completeness.} The representation should capture as much 3D structural information as possible through multi-axis sampling.  
    \item \textbf{Training-free extraction.} The pipeline relies entirely on frozen 2D foundation models, preserving zero-shot capability.  
    \item \textbf{Memory efficiency.} The representation must remain compact enough for cross-volume similarity computation and graph-based processing.  
\end{enumerate}

These principles guide the design of our training-free 3D-patch extraction pipeline presented in Section~\ref{sec:3dpatch-extraction}.
\section{3D-Patch Token Extraction for Batch-Based Zero-Shot}
\label{sec:3dpatch-extraction}

\begin{figure}[t]
    \centering
    \includegraphics[width=1.0\textwidth]{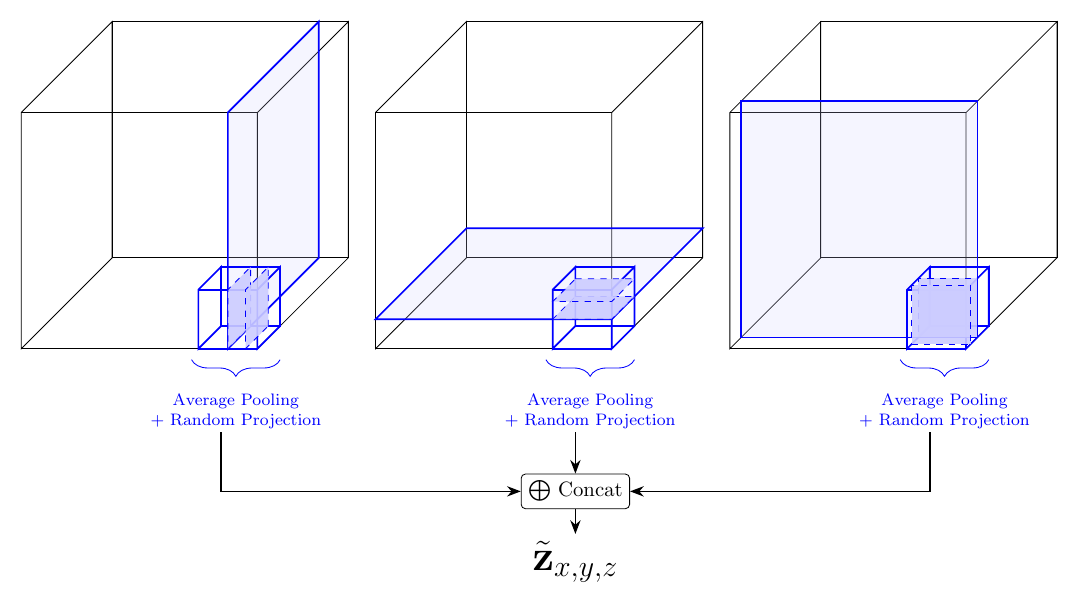}
    \caption{Diagram of the 3D-patch token extraction procedure.}
    \label{fig:extract-3dpatch}
\end{figure}

In this section, we present our method for extracting features from 3D MRI volumes, assuming each volume is a cubic grid of size \(H \times H \times H\) (e.g., \(224^3\)). Drawing inspiration from recent works that leverage 2D foundation models to construct volumetric features~\citep{an2025raptor,kim20243d}, our approach extracts orthogonal cross-sections along the three anatomical axes—axial, coronal, and sagittal—referring to top-down, front-back, and left-right views, respectively. These slices are processed using a frozen 2D model, followed by pooling, random projection, and fusion to form efficient 3D-patch tokens suitable for batch-based anomaly detection. The overall diagram of the extraction procedure is given in Figure~\ref{fig:extract-3dpatch}.

\subsubsection{Axis-wise Feature Extraction and Pooling}
\label{subsec:3dpatch-construction}

For clarity, we describe the extraction along the \(x\)-axis; the process is analogous for the \(y\)- and \(z\)-axes. Each slice \(S^{(\text{axis})}_h\) is processed independently by a frozen 2D foundation model, such as DINOv2~\cite{DINOv2} with patch size \(p\) (typically \(p = 14\)). The output is a grid of patch features (excluding the \texttt{[CLS]} token):
\[
S_h^{(\text{axis})} \;\longmapsto\;
\big\{ \mathbf{f}^{(\text{axis})}_{(h,y,z)} \in \mathbb{R}^{D} : 1 \le y,z \le N_p \big\},
\]
where \(N_p = H/p\) is the number of patches per axis, and \(D\) is the feature dimension. Stacking these features along the depth yields
\[
\mathbf{F}^{(\text{axis})} \in \mathbb{R}^{H \times N_p \times N_p \times D}.
\]
This representation is high-quality but too large for efficient volumetric processing—for example, DINOv2-L/14 on a \(224^3\) volume yields about 58 million floats per axis and roughly 57,000 tokens, exceeding typical GPU memory limits for batch similarity computations.

To enhance efficiency while retaining spatial information essential for anomaly segmentation, we apply \emph{patch-aligned average pooling} along consecutive slices. Slices are grouped into non-overlapping blocks of size \(p\), aligning with the 2D patch size to cover cubic regions of \(p \times p \times p\). For slices \(h = 1,\dots,H\), groups are
\[
\mathcal{G}_x = \{ (x-1)p + 1, \dots, xp \}, \quad x = 1,\dots,N_p.
\]
The 3D-patch token is then
\begin{equation}
\mathbf{z}^{(\text{axis})}_{(x,y,z)} 
= \frac{1}{p} \sum_{h \in \mathcal{G}_x} \mathbf{f}^{(\text{axis})}_{(h,y,z)},
\label{eq:3dpatch}
\end{equation}
followed by \(\ell_2\)-normalization. This reduces the size by a factor of \(p\) along the slicing axis, preserving spatial information. The final result is
\[
\mathbf{Z}^{(\text{axis})} \in \mathbb{R}^{N_p \times N_p \times N_p \times D},
\]
with one feature per cubic patch. After processing all axes, we permute coordinates to align all tensors in a consistent \((x,y,z)\) orientation for fusion.
\subsubsection{Random Projection and Fusion}
Even after pooling, the feature dimension \(D\) (typically 1024 or higher) remains computationally expensive.  
To further reduce computational cost, we apply a random linear projection following the Johnson--Lindenstrauss lemma~\citep{johnson1984extensions}.  
Let \(\mathbf{R} \in \mathbb{R}^{D \times k}\) be a Gaussian random matrix with \(k \ll D\) (e.g., \(k=64\)).  
Each token is projected as
\[
\tilde{\mathbf{z}}^{(\text{axis})}_{(x,y,z)} = 
\mathbf{R}^\top \mathbf{z}^{(\text{axis})}_{(x,y,z)} \in \mathbb{R}^{k}.
\]
We select random projection because it preserves pairwise distances with small distortion bounds and preserving nearest neighbor distances~\cite{indyk2007nearest}, making it well-suited to distance-based methods such as ours.

To integrate complementary context from the three axes, we concatenate projected features from axial, coronal, and sagittal directions at each position \((x,y,z)\):
\[
\tilde{\mathbf{z}}_{(x,y,z)} 
= \big[ \tilde{\mathbf{z}}^{(\text{axial})}_{(x,y,z)},\,
       \tilde{\mathbf{z}}^{(\text{coronal})}_{(x,y,z)},\,
       \tilde{\mathbf{z}}^{(\text{sagittal})}_{(x,y,z)} \big]
       \in \mathbb{R}^{3k}.
\]
Flattening the fused grid yields a volumetric sequence
\[
V_i \;\longmapsto\;
\big\{ \tilde{\mathbf{z}}_i^1, \tilde{\mathbf{z}}_i^2, \dots, \tilde{\mathbf{z}}_i^{M} \big\}, 
\quad M = N_p^3,
\]
where each token corresponds to a cubic region enriched with multi-axis information.
\subsubsection{Applying batch-based methods}
At this point, the 3D representation fits seamlessly into our batch-based framework.  
From the algorithmic perspective, we simply replace the 2D patch tokens with 3D-patch tokens.  
All subsequent procedures—mutual similarity vector computation, anomaly-graph construction, community detection, and base-set refinement—proceed exactly as described for 2D images.  
This direct compatibility arises because both formulations treat their inputs as unordered collections of elements.

\paragraph{Handling Void Voxels.}
MRI volumes often contain large void regions (e.g., zero-valued voxels from skull-stripping), producing identical features that can bias the similarity graph \(\mathcal{G}\) by dominating suspicious links $S_{\ell}$. To address this, we use brain masks (voxels that are greater than zero, easily extracted from MRI) to exclude 3D patches primarily composed of voids before graph construction. This filtering ensures anomaly communities in CoDeGraph reflect meaningful tissue regions.

% ============================================================
\section{Experiments on 3D MRI}
\label{sec:3d-experiments}

% (keep structure as previously agreed; now everything speaks about 3D-patch tokens and 3D anomaly reconstruction)
\subsection{Experimental Setup}
\label{subsec:3d-experiment-setup}

\paragraph{Datasets.}
We evaluated the proposed 3D extension on two brain MRI collections: IXI~\cite{ixi} (healthy controls) and BraTS-METS 2025~\cite{brats} (brain metastases). We used T2-weighted sequence, since the IXI dataset does not provide FLAIR sequence, only T2-weighted sequence. Following the preprocessing protocol in~\citep{brats}, all scans were (i) registered to the SRI-24 template, (ii) skull-stripped using HD-BET~\citep{hd-bet}, and then (iii) cropped to $196\times196\times196$ to remove void background. To match the input size of vision transformers and to minimize zero-valued voxels, the cropped scans were resized to $224\times224\times224$ and standardized using histogram normalization~\citep{nyul2000new}. For the experiments, we randomly selected 100 healthy volumes from IXI and 50 pre-treatment volumes from BraTS-METS 2025 (out of 328 available cases) to obtain a setting with sufficient healthy data to build the batch and a non-trivial number of pathological cases for evaluation.

\paragraph{Implementation Details.}
We primarily used Dinov2-L/14 as the backbone, which provides strong visual features without requiring a text modality. For fair comparison with text-based zero-shot methods that rely on CLIP, we also report results using the CLIP ViT-L/14@224 backbone. 

For both models, we extracted patch features at $6^{\text{th}}, 12^{\text{th}}, 18^{\text{th}}$ and $24^{\text{th}}$ layers. We constructed 3D-patch tokens as described in Section~\ref{sec:3dpatch-extraction}. After patch-aligned pooling, all tokens were projected to a low-dimensional space of size $k = 128$ using a fixed random Gaussian projection shared across volumes. Unless otherwise specified, all other hyperparameters were kept identical to those used in Chapter \ref{chap:algorithm}.  As in the 2D image experiments, the mutual similarity vector for constructing \(\mathcal{G}\) was built by averaging layer-wise mutual similarity vectors.

\paragraph{Evaluation Metrics.}
Performance was evaluated at two levels: patient and voxel.  
At the \textit{patient level}, each 3D MRI volume was assigned a single anomaly score defined as the maximum over all 3D-patch scores within the volume. 
We then reported the Area Under the Receiver Operating Characteristic curve (AUROC) and the maximum F1-score (F1-max) for classification between healthy vs. lesion-containing scans.

At the \textit{voxel level}, we evaluated the ability of the model to localize abnormal regions. 
Anomaly scores were first obtained on the low-resolution token grid of size \(N_p \times N_p \times N_p\) (e.g., \(16^3\) for a \(224^3\) volume with patch size \(p=14\)). 
Ground-truth lesion masks were downsampled to the same grid using max-pooling, marking a token as anomalous if any voxel within its \(14\times14\times14\) cube belonged to a lesion. 
The optimal threshold yielding F1-max on this low-resolution grid was then used to binarize the upsampled predictions (\(224^3\)) for computing the 3D Dice score against the full-resolution ground-truth masks. 
Voxel-wise AUROC was computed directly at full resolution.
For scans with no lesions where the model produced no positive predictions at the chosen threshold, the Dice score was defined as one, preventing penalization of correct all-normal predictions.
\paragraph{Baselines.}

We compared our method with two representative 2D text-based zero-shot approaches, {APRIL-GAN}~\citep{APRIL-GAN} and {AnomalyCLIP}~\citep{AnomalyCLIP}. 
Both methods rely on language-guided vision transformers ViT-L/14@224and were adapted to the 3D MRI setting through multi-view slice-wise processing. 
Each MRI volume was decomposed into 224 slices along the three anatomical planes—axial, sagittal, and coronal—with each slice of size \(224\times224\). 
Every slice was independently processed by the respective 2D model to produce a \(16\times16\) anomaly map. 
For each viewing axis, the slice-level maps were stacked into a volumetric tensor of size \(224\times16\times16\) and downsampled along the depth dimension using max-pooling with kernel and stride equal to the patch size (\(p=14\)), resulting in a \(16\times16\times16\) anomaly map per axis. 
The three volumetric maps obtained from the different axes were then summed element-wise to form the final anomaly map of size \(16\times16\times16\), providing a volumetric representation that integrates complementary spatial evidence from multiple orientations. Voxel-level AUROC and Dice scores were computed using the same thresholding and upsampling procedures as CoDeGraph, while patient-level AUROC and F1-max were obtained by taking the maximum anomaly score across all slices from all three axes. 

For fair and reproducible comparisons, we used the official implementations released by the authors of both APRIL-GAN and AnomalyCLIP. 
We first fine-tuned each model on MVTec AD following the settings reported in the respective papers, and then further fine-tuned them on 2D tumor slices extracted from the remaining pre-treatment BraTS-METS samples, following the same procedure described in Section~\ref{sec:exp-setup} of Chapter~\ref{chap:empirical}. 
This extension enables evaluation of text-based methods quantifies the modality gap between 2D vision–language models and our 3D-patch-based approach.

\subsection{Main Results}
\label{subsec:mri3d-results}
\begin{table}[t]
\centering
\caption{Quantitative comparison on \textbf{3D MRI volumes} (IXI and BraTS-METS 2025). 
Patient-level metrics are computed from the maximum anomaly score per volume. 
Voxel-level AUROC is computed on full-resolution volumes, and Dice is measured at high resolution using the threshold derived from F1-max on the low-resolution grid. 
All values are reported in percentages (\%).}
\label{tab:mri3d_results}
\resizebox{\textwidth}{!}{
\begin{tabular}{l|l|l|ccc|c}
\toprule
\textbf{Method} & 
\textbf{Backbone} & 
\thead{Fine-tuning \\ Dataset} & 
\thead{AUROC \\ (Patient)} & 
\thead{F1-max \\ (Patient)} & 
\thead{AUROC \\ (Voxel)} & 
\thead{Dice \\ (3D)} \\
\midrule
CoDeGraph & DINOv2-L/14 & -- & \textbf{99.96} & \textbf{99.01} & \textbf{98.17} & \textbf{75.43} \\
CoDeGraph & ViT-L/14 & -- & 98.30 & 93.33 & 97.20 & 62.87 \\
MuSc & DINOv2-L/14 & -- & 97.34 & 88.89 & 97.80 & 67.73 \\
MuSc & ViT-L/14 & -- & 98.40 & 92.45 & 97.28 & 61.65 \\
\midrule
AnomalyCLIP & ViT-L/14 & Brain slices & 87.56 & 75.00 & 94.93 & 6.79 \\
AnomalyCLIP & ViT-L/14 & MVTec AD & 64.42 & 56.72 & 93.27 & 3.75 \\
APRIL-GAN & ViT-L/14 & Brain slices & 24.38 & 50.00 & 96.58 & 68.22 \\
APRIL-GAN & ViT-L/14 & MVTec AD & 29.87 & 50.25 & 95.33 & 4.80 \\
\bottomrule
\end{tabular}}
\end{table}
\paragraph{Quantitative Analysis.}
Table~\ref{tab:mri3d_results} presents the quantitative results on volumetric MRI anomaly detection. 
CoDeGraph consistently outperformed both the batch-based baseline (MuSc) and the slice-wise text-based methods under the same volumetric evaluation protocol. 
With the DINOv2-L/14 backbone, CoDeGraph achieved nearly perfect patient-level discrimination (AUROC \(99.96\%\), F1-max \(99.01\%\)) and the highest 3D Dice score (\(75.43\%\)), confirming that its and graph-based refinement and the principles of batch-based methods remains effective in volumetric domains. 

Slice-wise text-based methods exhibited a significant performance gap when transferred from 2D to 3D. 
Although APRIL-GAN and AnomalyCLIP could detect slice-level abnormalities, they failed to maintain volumetric coherence, resulting in low Dice scores. 
Unsuprisingly, the versions fine-tuned on brain slices performed considerably better than those trained on MVTec AD, underscoring the domain and modality gap between industrial textures and medical anatomy. 
This finding aligns with the observations of~\cite{marzullo2024exploring}, which report that adapting 2D vision–language models to volumetric data requires careful handling of spatial and semantic mismatches.

\paragraph{Qualitative Analysis.}

% Qualitative figures
\begin{figure}[t]
    \centering
    \includegraphics[width=0.46\textwidth]{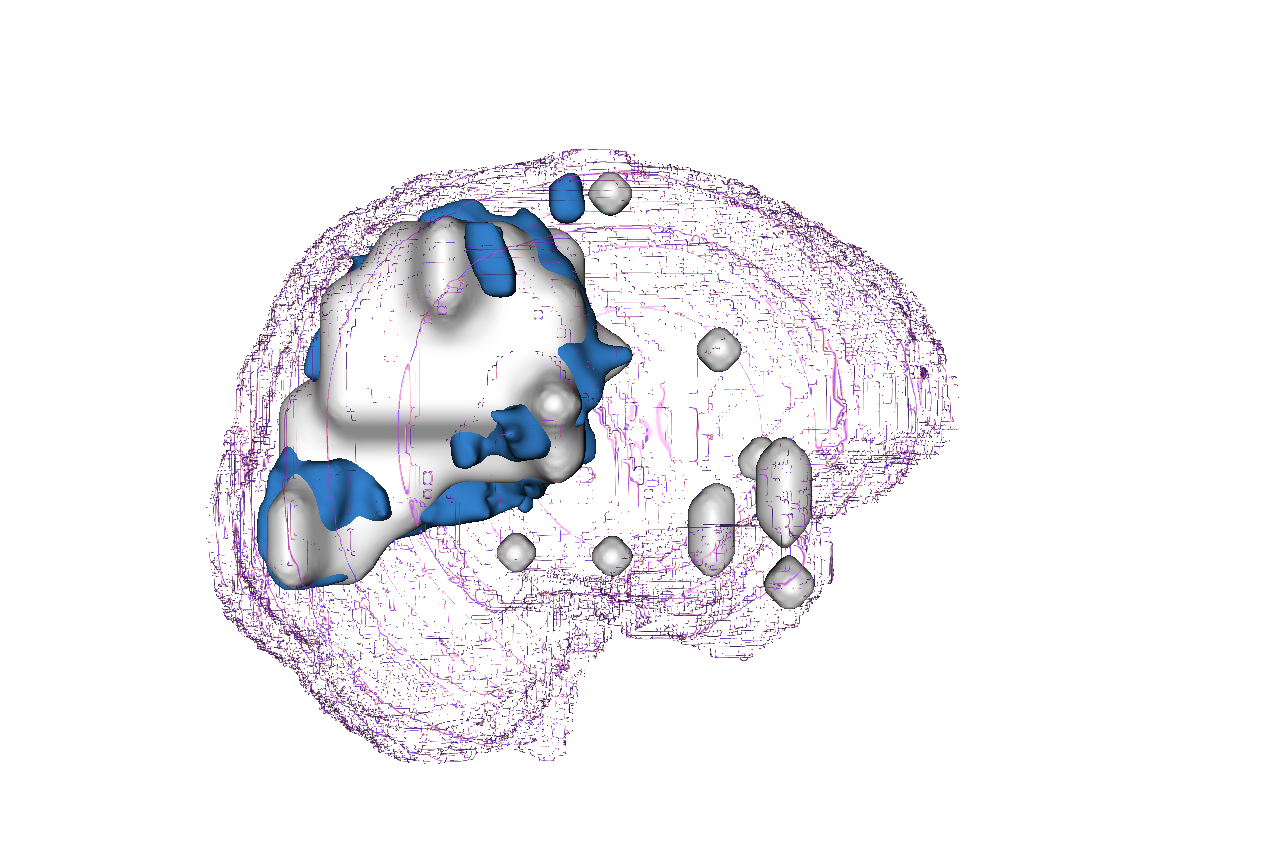}
    \hfill
    \includegraphics[width=0.46\textwidth]{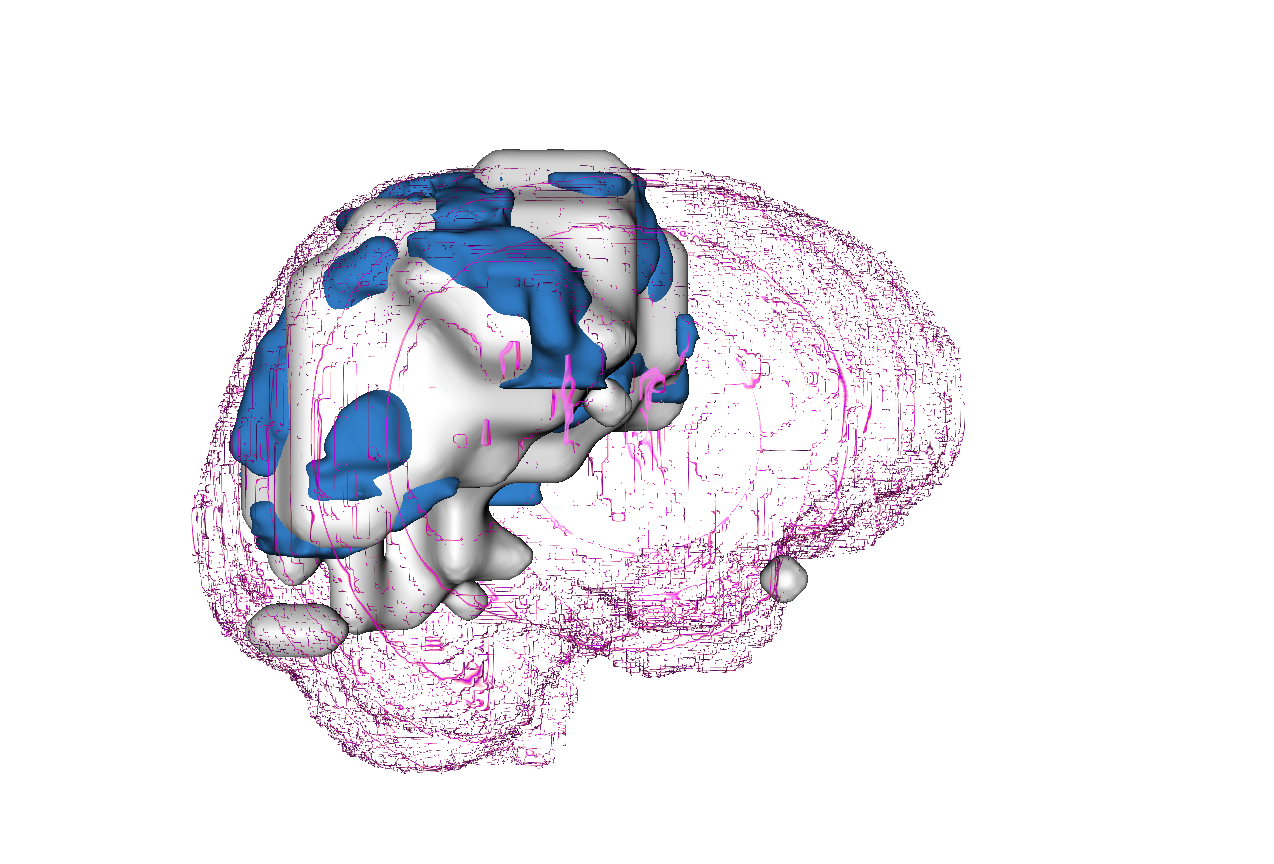}
    \caption{Qualitative 3D segmentation results of CoDeGraph with DINOv2-L/14 on MRI volumes (successful cases). 
    Blue denotes the ground-truth tumor mask and white denotes the CoDeGraph prediction. 
All visualizations were generated in 3D Slicer~\cite{3d-slicer}.}
    \label{fig:3d_mri_good}
\end{figure}

\begin{figure}[t]
    \centering
    \includegraphics[width=0.7\textwidth]{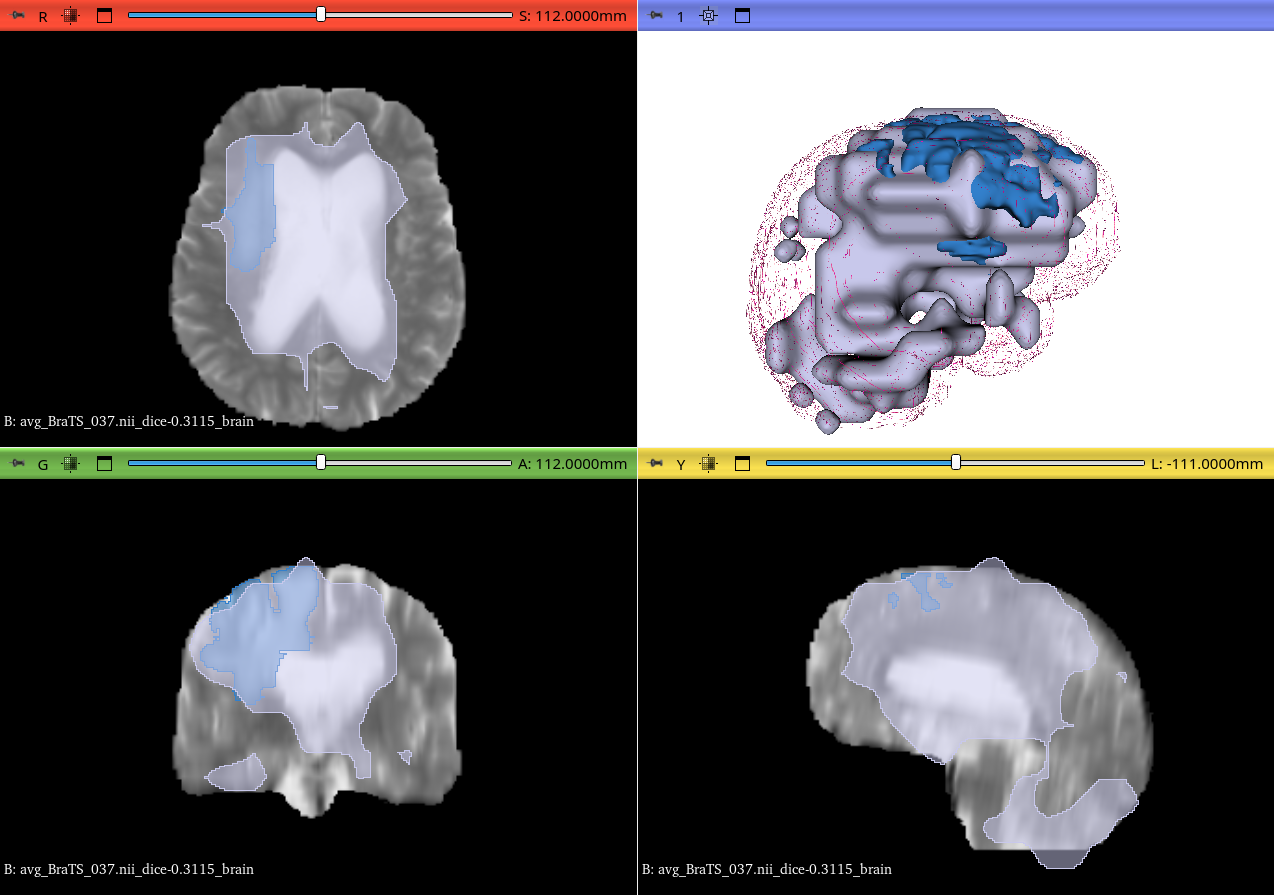}
    \caption{Failure case of CoDeGraph with DINOv2-L/14. 
    The model either misses small tumors or falsely marks normal regions as abnormal. 
    Blue denotes the ground-truth tumor mask and white denotes the CoDeGraph prediction. 
    Visualization produced in 3D Slicer~\cite{3d-slicer}, including three orthogonal projections.}
    \label{fig:3d_mri_fail}
\end{figure}
Visual examples demonstrate that CoDeGraph effectively localizes large tumor regions within 3D MRI volumes, producing spatially coherent and anatomically consistent anomaly maps.
However, CoDeGraph struggles with small or low-contrast tumors, where anomalies occupy only a fraction of a single 3D patch, leading to partial or missed detections. 
False positives occasionally occur in anatomically homogeneous areas, where normal tissues are mistakenly segmented as abnormal. 
These limitations primarily arise from the coarse grid resolution (\(16^3\)) of the 3D-patch tokens. 
Future research aims to improve computational efficiency, enabling higher volumetric resolutions such as \(518^3\), analogous to the \(518^2\) image resolution used in the 2D setting, which would allow finer-grained and more accurate volumetric segmentation.

\paragraph{Anomaly Similarity Graph.}
In the 3D volumetric setting, CoDeGraph (DINOv2) revealed the emergence of two distinct communities of anomalous MRI volumes within the anomaly similarity graph~$\mathcal{G}$, with cluster sizes of 13 and 34. 
These communities correspond to recurring pathological patterns that deviate from the normal population. 
By identifying and removing these anomalous clusters from the base set~$\mathcal{B}$, CoDeGraph effectively refines the reference pool, reducing bias from repeated abnormal samples. 
This refinement leads directly to improved anomaly localization and classification performance of CoDeGraph over MuSc.
\paragraph{Effect of Backbone.}
As in the 2D experiments, DINOv2-L/14 consistently outperformed ViT-L/14 across all metrics, confirming the superiority of self-supervised visual representations for anomaly detection (e.g., CoDeGraph with Dinov2-L/14 backbone achieved a superior 3D Dice score of $75.42\%$). 
The DINOv2 backbone yielded higher accuracy and more coherent volumetric segmentation. 
This observation suggests that for vision-only tasks—where semantic alignment with text is unnecessary—purely image-trained models such as DINOv2 provide stronger and more generalizable features than vision–language models like CLIP ViT-L/14.
\paragraph{Computational Efficiency and Memory Footprint.}
One of the most critical considerations for extending anomaly detection to 3D medical imaging is the computational cost—both in terms of processing time and GPU memory usage. 
Unlike 2D imagery, volumetric data introduces cubic growth in the number of tokens, making runtime efficiency a decisive factor for practical deployment.

\begin{figure}[t]
\centering
\begin{tikzpicture}[scale=2, every node/.style={font=\small}]
    % data
    \def\a{72.9} % token construction
    \def\b{27.1} % MSM + graph
    \def\r{1.3}

    % colors
    \definecolor{pieA}{RGB}{70,130,180} % steelblue
    \definecolor{pieB}{RGB}{220,20,60}  % crimson

    % angles
    \pgfmathsetmacro{\aAngle}{360*\b/100} % 72.9% of circle
    \pgfmathsetmacro{\bAngle}{360 - \aAngle}

    % --- slice A (3D-patch token construction) ---
    \fill[pieA] (0,0) -- (0:\r) arc (0:\aAngle:\r) -- cycle;

    % --- slice B (mutual similarity + graph) ---
    \fill[pieB] (0,0) -- (\aAngle:\r) arc (\aAngle:360:\r) -- cycle;

    % outline
    \draw[thin] (0,0) circle (\r);

    % label for slice A
    \pgfmathsetmacro{\midA}{\aAngle/2}
    \pgfmathsetmacro{\midB}{\aAngle + \bAngle/2}
    \node[white,font=\small] at (\midA:{0.7*\r}) {27.1\%};
    \node[white,font=\small] at (\midB:{0.6*\r}) {72.9\%};
    \coordinate (Aout) at (\midA:{\r+0.35});
    \draw[-{Latex[length=2mm]},thin] (\midA:{\r}) -- (Aout);
    \node[anchor=west] at ($(Aout)+(0.05,0)$) {Mutual similarity \& graph};

    % label for slice B
    \pgfmathsetmacro{\midB}{\aAngle + \bAngle/2}
    \coordinate (Bout) at (\midB:{\r+0.35});
    \draw[-{Latex[length=2mm]},thin] (\midB:{\r}) -- (Bout);
    \node[anchor=east] at ($(Bout)+(-0.05,0)$) {3D-patch token construction};

\end{tikzpicture}

\caption{Runtime breakdown of the 3D MRI pipeline. Most of the time is spent on 3D-patch token construction from the three axes (72.9\%), while mutual similarity computation and CoDeGraph refinement account for the remaining 27.1\%. Total time: 620.29\,s for 150 volumes (4.13\,s/volume).}
\label{fig:mri3d_runtime_pie_tikz}
\end{figure}
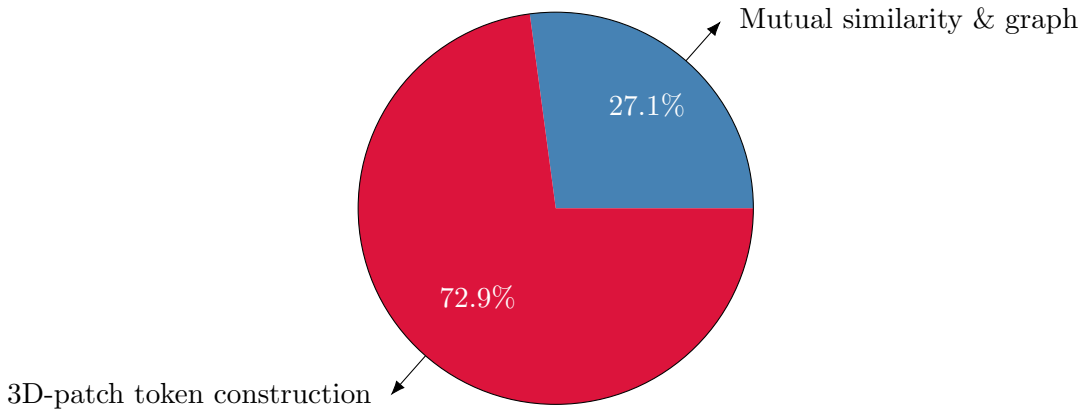

The complete 3D CoDeGraph pipeline—including multi-axis patch-token extraction, mutual similarity computation, and graph-based refinement—was executed on a single RTX~4070~Ti~Super GPU. 
Processing all 150 MRI volumes required a total of \(620.29\)~seconds, corresponding to approximately \(\mathbf{4.13}\)~seconds per MRI, which is acceptable for 3D processing. 
As shown in Figure~\ref{fig:mri3d_runtime_pie_tikz}, 3D patch-token extraction dominated the runtime, accounting for about \(72.9\%\) (\(452.43\)~seconds) of the total, while mutual similarity and graph-related operations contributed \(27.1\%\) (\(167.86\)~seconds). 
Throughout the process, maximized GPU memory consumption was \(6.49\)~GB. These results demonstrate that the proposed 3D extension of CoDeGraph remains computationally tractable even for volumetric data. 
% ============================================================
\section{Discussion and Future Directions}
\label{sec:mri3d-discussion}
The experiments demonstrate that batch-based zero-shot anomaly detection methods are a viable and scalable solution for 3D volumetric data, particularly in patient-level screening. 
Without any fine-tuning or supervision, CoDeGraph effectively distinguishes normal and pathological MRI scans while localizing large tumor regions with high reliability. 
The entire process runs within a few seconds per volume and requires no annotated data, making it a practical candidate for integration into medical imaging workflows—especially in scenarios where both unlabeled normal and abnormal samples are abundant. Among the different stages, segmentation remains the most computationally demanding step, yet the overall pipeline still enables rapid filtering of normal versus abnormal scans and fast identification of suspicious regions that may contain lesions or tumors.  
This aligns with the growing demand for triage-oriented tools that support radiologists by reducing manual screening time.

However, the framework inherits certain assumptions from its theoretical foundation. 
In particular, the Doppelgänger assumption presumes that normal patches across volumes share consistent structural similarity, which might not hold for many anatomical structures such as vasculars. 

Future research will focus on optimizing the computational pipeline to handle higher volumetric resolutions—potentially extending input size to \(518^3\) voxels, analogous to the \(518^2\) resolution used in 2D experiments.  
Beyond the brain domain, further investigations could explore applications to other medical modalities such as liver CT, cardiac MRI and retinal OCT volumes, where structural regularity and consistency assumptions are partially preserved.  

% ============================================================
\chapter{Bridging Batch-Based and Text-Based Zero-Shot Methods}
\label{chap:bridge-batch-text}

This chapter takes intial steps toward combining two complementary paradigms in zero-shot anomaly detection: batch-based methods and text-based vision–language models

\medskip

The chapter is organized as follows.  
Section~\ref{sec:motivation-bridge} describes the motivation and provides an overview of this preliminary bridging framework.  
Section~\ref{sec:gmm-pseudomask} presents the generation of pseudo-masks from batch-based anomaly scores using statistical modeling and adaptive thresholding.  
Section~\ref{sec:exp-bridge} reports experimental results, including the setup and limitations of this proof-of-concept approach.
\section{Motivation and Overview}
\label{sec:motivation-bridge}

Although they arise from distinct philosophies, both batch-based and text-based zero-shot methods pursue the same ultimate goal: detecting anomalies without human supervision. 
Batch-based methods exploit the overwhelming prevalence of normal patterns in a batch, achieving high robustness and accuracy through cross-sample consistency. 
However, they require large batches, substantial memory, and expensive computation. 
In contrast, text-based methods rely on vision–language alignment and semantic prompts to perform anomaly detection on individual images. 
They are computationally efficient, interpretable, and easily extensible across domains, yet their pixel-level localization accuracy remains limited without explicit fine-tuning using annotated datasets.

This chapter serves as a \emph{proof-of-concept experiment} that explores whether pseudo anomaly masks generated by a robust batch-based model—such as CoDeGraph—can substitute ground-truth masks to train or fine-tune a text-based model. 
If successful, this bridging approach would suggest that strong structural knowledge from unsupervised, batch-based inference can compensate for the lack of labeled data in text-based models. 
It thus raises a broader question: \emph{Can high-quality pseudo-masks, when combined with a robust text-based model, approach the performance achieved using true pixel-level supervision?}

Such a strategy has two primary implications. 
First, it enables the transfer of dense structural information from unsupervised visual similarity to semantically aligned language-vision representations, effectively transforming unsupervised outputs into pseudo-supervised signals. 
Second, it opens a new perspective on the \emph{robustness} of text-based models—how far they can generalize when trained with imperfect, yet statistically consistent, supervision.

% ============================================================
\section{Pseudo-Mask Generation from Batch-Based Models}
\label{sec:gmm-pseudomask}
\begin{figure}[t]
  \centering
  \includegraphics[width=0.7\textwidth]{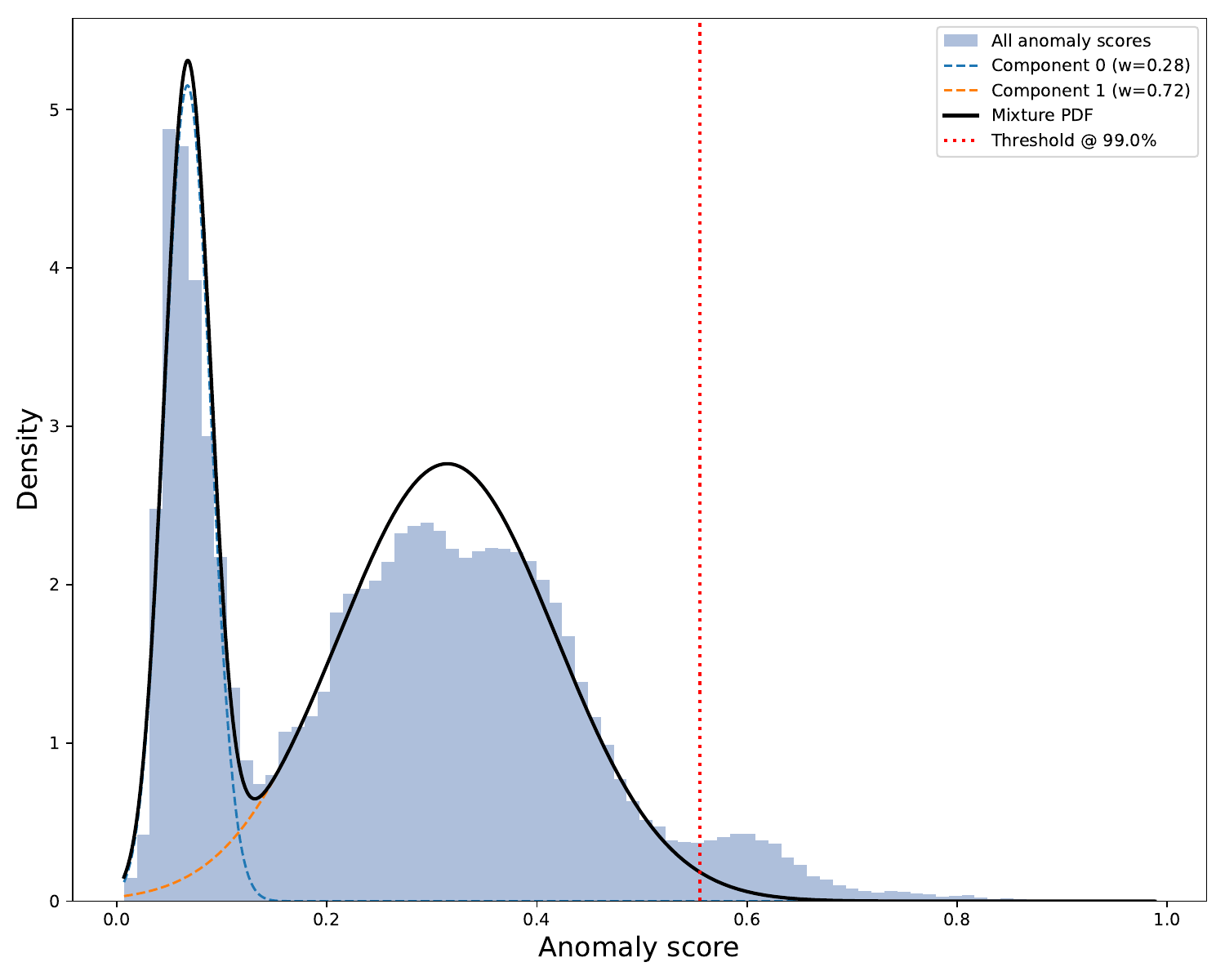}
  \caption{
  Histogram of pixel-wise anomaly scores for the \emph{bottle} category in MVTec AD, together with a fitted two-component Gaussian mixture model (GMM). 
  The first mode (left) corresponds to background pixels, while the second mode (middle) captures object pixels. 
  The heavy right tail reflects anomalous regions on the bottle surface, which deviate from both background and object modes and are used to define the pseudo-mask threshold in Section~\ref{sec:gmm-pseudomask}.}
  \label{fig:bottle_score_hist}
\end{figure}
\subsection{Statistical Modeling of Batch-Based Scores and Adaptive Thresholding}
The purpose of this stage is to generate a binary \emph{pseudo anomaly mask} from continuous anomaly maps produced by batch-based methods. 
To do so, we require a decision threshold \(\tau\) as defined in Eq.~\ref{for:anomaly_threshold}, which separates normal and anomalous pixels without access to any labeled validation data.

Empirically, the anomaly scores obtained from batch-based frameworks such as CoDeGraph or MuSc exhibit approximately Gaussian behavior, as shown in Fig~\ref{fig:bottle_score_hist}.
This arises naturally from the averaging of multiple top-\(k\) similarity distances, which—by the Central Limit Theorem—tends toward a Gaussian distribution for both normal regions. 
Hence, the overall distribution of pixel scores can be effectively modeled as a mixture of one or two Gaussian components, corresponding to background and anomaly domains.

Our method is intentionally simple: we fit a Gaussian Mixture Model (GMM) on a truncated portion of the anomaly scores, and then use the fitted parameters to derive an adaptive threshold for binarization. 
The truncation step removes the extreme tail of the distribution to ensure that the mixture fit is not distorted by a small fraction of true anomalies. 
Since the proportion of anomalous pixels is typically very low (smaller than \(3\%\) in MVTec AD and \(1\%\) in Visa), we can safely approximate this upper truncation point using a fixed quantile \(q_{\mathrm{fit}}\) (e.g., \(0.95\)).

Formally, given the continuous anomaly map \(A_i\) for image \(i\), the truncated fitting set is
\[
\mathcal{S}_{\mathrm{fit}} = \{\, s \in A_i \mid s \le F^{-1}(q_{\mathrm{fit}})\,\},
\]
where \(F^{-1}\) denotes the empirical quantile function. 
A GMM with one or two components is then fitted to \(\mathcal{S}_{\mathrm{fit}}\):
\[
p(s) = \sum_{j=1}^{K} \pi_j \, \mathcal{N}(s \mid \mu_j, \sigma_j^2), \quad K \in \{1, 2\},
\]
where \(K\) is automatically determined using the Bayesian Information Criterion (BIC). 
From the fitted mixture, a candidate threshold for each component is computed as
\[
t_j = \mu_j + \sigma_j\,\Phi^{-1}(q_{\mathrm{comp}}),
\]
where \(\Phi^{-1}\) is the inverse CDF of the standard normal distribution and \(q_{\mathrm{comp}}\) (typically \(0.99\)) controls the tail sensitivity. 
The final adaptive cutoff is then chosen as
\[
\tau^\ast = \max_{j} t_j,
\]
corresponding to the most anomalous component in the mixture.

Finally, the continuous anomaly map is converted into a binary pseudo-mask:
\[
\hat{M}_i(x) =
\begin{cases}
1, & \text{if } A_i(x) \ge \tau^\ast, \\
0, & \text{otherwise.}
\end{cases}
\]
This process transforms the unsupervised batch-based anomaly scores into pseudo annotations, ready for training or fine-tuning text-based models. 
\subsection{Fine-Tuning Text-Based Models with Pseudo-Masks}
\label{subsec:fine-tune-text}

Beyond obtaining reliable pseudo-masks from batch-based models, an effective strategy also requires a robust text-based framework capable of learning from imperfect supervision. 
Since pseudo-masks inevitably contain inaccuracies, the selected text-based models must tolerate label noise while still exploiting strong vision–language alignment.

For this study, we adopt two lightweight and practical text-based methods: APRIL-GAN and AnomalyCLIP. 
Both follow the binary classification formulation of WinCLIP~\citep{WinCLIP}, in which each patch is classified as ``normal'' or ``anomalous’’ according to its alignment with text prompts.

\paragraph{APRIL-GAN.}
APRIL-GAN constructs text embeddings using 85 CLIP prompt templates describing normal and abnormal conditions. 
It fine-tunes a small linear projection that maps visual patch tokens into the joint vision–language embedding space. 
Training uses pixel-level Dice loss and Focal loss while the CLIP backbone remains frozen.  
When pseudo-masks replace ground-truth annotations, the training pipeline remains unchanged.

\paragraph{AnomalyCLIP.}
AnomalyCLIP keeps the CLIP vision encoder frozen and instead fine-tunes learnable text tokens. 
It employs an object-agnostic prompting strategy:
\[
g_n = [V_1][V_2]\dots[V_E][\text{object}], \qquad
g_a = [W_1][W_2]\dots[W_E][\text{damaged}][\text{object}],
\]
instead of the class-specific form:
\[
g_n = [V_1][V_2]\dots[V_E][\text{cls}], \qquad
g_a = [W_1][W_2]\dots[W_E][\text{damaged}][\text{cls}].
\]
This removes dependence on predefined object names and forces the finetuning learn generic anomaly semantics.  

\medskip
In both methods, pseudo-masks act as pixel-level binary supervision, allowing text-based models to acquire spatial precision from batch-based inference while preserving their semantic adaptability. 
This yields a practical and annotation-free pathway to adapt text-based models for anomaly segmentation.

\section{Experiments}
\label{sec:exp-bridge}
\subsection{Experimental Setup}

\paragraph{Datasets}
We used the test split of the MVTec AD dataset~\cite{mvtec} as the source of training images for generating pseudo-masks.  
All pseudo-masks in this chapter were produced from the anomaly scores of CoDeGraph, as described in Chapter~\ref{chap:algorithm}.  
The fine-tuned text-based models were then evaluated on two groups of target datasets.  

The industrial group consisted of Visa~\cite{visa}, BTAD~\cite{mishra21-vt-adl}, and MPDD~\cite{9631567}, covering texture- and object-level defects encountered in real manufacturing environments.  
The medical group included CVC-ClinicDB~\cite{BERNAL201599}, CVC-ColonDB~\cite{7294676}, Kvasir~\cite{10.1007/978-3-030-37734-2_37}, and Thyroid~\cite{9434087}, representing a variety of endoscopic and ultrasound modalities. 

In addition, we conducted an in-domain experiment on MVTec AD by randomly splitting each category into 50\% for training and 50\% for testing.  
This experiment was designed to study how pseudo-mask supervision influences the domain generalization behaviour of text-based models when trained and evaluated under the same dataset distribution.

\paragraph{Implementation Details}
All experiments used the ViT-L/14@336 CLIP encoder, with all images resized to \(336 \times 336\).  
Pseudo-masks were generated using the truncated GMM procedure described in Section~\ref{sec:gmm-pseudomask}, with truncation quantile \(q_{\mathrm{fit}} = 0.95\) and component threshold \(q_{\mathrm{comp}} = 0.99\).  
APRIL-GAN and AnomalyCLIP were fine-tuned using their public source implementations without architectural modifications.  
For both APRIL-GAN and AnomalyCLIP, only the segmentation branch was fine-tuned while the anomaly classification head remained frozen to preserve its original zero-shot behaviour.  
Both pseudo-mask and ground-truth supervision were applied under identical training settings to enable a fair comparison.

\paragraph{Evaluation Metrics}
Since this chapter focused on pixel-level anomaly segmentation, we reported four segmentation metrics: pixel-AUROC, pixel-F1, pixel-AP, and pixel-AUPRO (PRO-seg).  
These metrics quantified discriminability, region-wise localization quality, and the overall accuracy of the predicted anomaly maps after fine-tuning with either pseudo- or ground-truth masks.

\begin{center}
\begin{table}[h]
\centering
\caption{Comparison between pseudo-mask and ground-truth fine-tuning on industrial, medical, and MVTec (50\%/50\%) datasets. 
$\Delta$ denotes the performance difference (Pseudo $-$ GT).}
\begin{tabular}{l|cccc}
\toprule
\textbf{Setting} 
& \textbf{AUROC-seg} 
& \textbf{F1-seg} 
& \textbf{AP-seg} 
& \textbf{PRO-seg} \\
\midrule

\multicolumn{5}{c}{\textbf{AnomalyCLIP}} \\
\midrule
Industrial (Pseudo) & 95.6 & 35.2 & 30.8 & 81.7 \\
Industrial (GT)     & 95.4 & 35.3 & 30.4 & 79.8 \\
$\Delta$            & +0.2 & $-$0.1 & +0.4 & +1.9 \\
\midrule
Medical   (Pseudo)  & 85.2 & 47.8 & 44.7 & 60.9 \\
Medical   (GT)      & 84.0 & 45.0 & 40.8 & 59.1 \\
$\Delta$            & +1.2 & +2.8 & +3.9 & +1.8 \\
\midrule
MVTec 50\% (Pseudo) & 94.1 & 46.3 & 43.2 & 88.2 \\
MVTec 50\% (GT)     & 95.0 & 48.1 & 45.7 & 88.4 \\
$\Delta$            & $-$0.9 & $-$1.8 & $-$2.5 & $-$0.2 \\
\midrule

\multicolumn{5}{c}{\textbf{APRIL-GAN}} \\
\midrule
Industrial (Pseudo) & 92.4 & 75.4 & 24.8 & 30.0 \\
Industrial (GT)     & 91.9 & 71.3 & 24.0 & 30.1 \\
$\Delta$            & +0.5 & +4.1 & +0.8 & $-$0.1 \\
\midrule
Medical   (Pseudo)  & 76.4 & 54.4 & 30.0 & 36.5 \\
Medical   (GT)      & 80.3 & 55.3 & 33.5 & 41.3 \\
$\Delta$            & $-$3.9 & $-$0.9 & $-$3.5 & $-$4.8 \\
\midrule
MVTec 50\% (Pseudo) & 97.8 & 64.4 & 64.6 & 91.6 \\
MVTec 50\% (GT)     & 99.1 & 76.7 & 81.5 & 83.7 \\
$\Delta$            & $-$1.3 & $-$12.3 & $-$16.9 & +7.9 \\
\bottomrule
\end{tabular}
\label{tab:pseudo}
\end{table}
\end{center}
\subsection{Results Analysis}

The results in Table~\ref{tab:pseudo} reveal clear differences in how the two text-based models respond to pseudo-mask supervision.

AnomalyCLIP showed strong robustness to pseudo-masks.  
In both industrial and medical domains, pseudo supervision matched or exceeded ground-truth performance, with especially large gains in the medical datasets.  
This suggests that the object-agnostic prompting and text-side adaptation of AnomalyCLIP naturally absorb the softer, structurally consistent pseudo-masks produced by CoDeGraph.  

APRIL-GAN was more sensitive to mask quality.  
Pseudo-masks performed well for industrial datasets but degraded performance on medical and in-domain MVTec experiments.  
The large gap between pseudo and ground-truth training in the 50\%/50\% MVTec split emphasizes this sensitivity.

Overall, these findings demonstrate that batch-based pseudo-masks can train text-based models.  
AnomalyCLIP, in particular, proved highly robust and consistently benefited from pseudo supervision, while APRIL-GAN required cleaner training masks to maintain performance.  
This confirms that the proposed pseudo-mask generation pipeline is practical and that AnomalyCLIP is especially well-suited for bridging batch-based and text-based zero-shot methods.

\subsection{Limitations}
The experiments in this chapter demonstrate that bridging batch-based and text-based methods through pseudo-mask supervision is a promising direction, particularly given the strong robustness exhibited by AnomalyCLIP.  
However, the scope of the investigation remains limited.  
A more comprehensive study would require evaluating additional text-based architectures, exploring alternative pseudo-mask generation schemes, and conducting systematic cross-domain analyses beyond the datasets considered here.  
Due to time and hardware constraints, the author was unable to perform these more extensive experiments.  
As such, the results presented in this chapter should be viewed as an initial proof-of-concept rather than a complete characterization of the bridging strategy’s potential.
\appendix
\chapter{Appendices}
\section{Appendix: Proofs}
\label{app:theorem}
\begin{proof}[Proof of Theorem~\ref{thm:local-ball-asymptotics}]
Let $f \in L^1_{\mathrm{loc}}(\mathbb{R}^d)$ be the density of $\mu$ and
$B(\mathbf{z},r)=\{x:\|x-\mathbf{z}\|\le r\}$.
By the Lebesgue Differentiation Theorem, for almost every
$\mathbf{z}\in\mathbb{R}^d$,
\[
  \lim_{r\downarrow 0}
  \frac{1}{\mathrm{Vol}\bigl(B(\mathbf{z},r)\bigr)}
  \int_{B(\mathbf{z},r)} f(x)\,dx
  \;=\; f(\mathbf{z}).
\]
Since $\mu\bigl(B(\mathbf{z},r)\bigr)=\int_{B(\mathbf{z},r)} f(x)\,dx$
and $\mathrm{Vol}\bigl(B(\mathbf{z},r)\bigr)=v_d\,r^d$, we obtain
\[
  \lim_{r\downarrow 0}
  \frac{\mu\bigl(B(\mathbf{z},r)\bigr)}{r^d}
  \;=\;
  \lim_{r\downarrow 0}
  \frac{\mu\bigl(B(\mathbf{z},r)\bigr)}{\mathrm{Vol}\bigl(B(\mathbf{z},r)\bigr)}
  \cdot
  \frac{\mathrm{Vol}\bigl(B(\mathbf{z},r)\bigr)}{r^d}
  \;=\; f(\mathbf{z})\,v_d.
\]
This holds for almost every $\mathbf{z}$, which proves the theorem.
\end{proof}

\medskip

\begin{proof}[Proof of Theorem~\ref{thm:rv-similarity-main}]
Fix $\mathbf{z}$ such that $f(\mathbf{z})>0$ and
\[
  \lim_{r\downarrow 0} \frac{\mu\bigl(B(\mathbf{z},r)\bigr)}{r^d}
  \;=\; f(\mathbf{z})\,v_d \;=:\; c(\mathbf{z}) \in (0,\infty).
\]
Let $X\sim \mu$, $Z=\|X-\mathbf{z}\|$, and $S=1/Z$.
Write $F_Z(r)=\mathbb{P}(Z\le r)=\mu\bigl(B(\mathbf{z},r)\bigr)$.
By the above limit, for every $\varepsilon\in(0,1)$ there exists
$\delta>0$ such that for all $0<r<\delta$,
\begin{equation}
\label{eq:local-bounds}
  (1-\varepsilon)\,c(\mathbf{z})\,r^d
  \;\le\;
  F_Z(r)
  \;\le\;
  (1+\varepsilon)\,c(\mathbf{z})\,r^d.
\end{equation}
Since $\mathbb{P}(S>s)=\mathbb{P}(Z<1/s)=F_Z(1/s)$, applying
\eqref{eq:local-bounds} with $r=1/s$ (valid for all $s>1/\delta$) yields
\begin{equation}
\label{eq:tail-bounds}
  (1-\varepsilon)\,c(\mathbf{z})\,s^{-d}
  \;\le\;
  \mathbb{P}(S>s)
  \;\le\;
  (1+\varepsilon)\,c(\mathbf{z})\,s^{-d},
  \qquad s>1/\delta.
\end{equation}
To prove regular variation, fix $x>0$. For all $t>\max\{1/\delta,\,1/(x\delta)\}$
we can apply \eqref{eq:tail-bounds} at $s=t$ and $s=tx$ to get
\[
  \frac{(1-\varepsilon)c(\mathbf{z})(tx)^{-d}}{(1+\varepsilon)c(\mathbf{z})t^{-d}}
  \;\le\;
  \frac{\mathbb{P}(S>tx)}{\mathbb{P}(S>t)}
  \;\le\;
  \frac{(1+\varepsilon)c(\mathbf{z})(tx)^{-d}}{(1-\varepsilon)c(\mathbf{z})t^{-d}}.
\]
After simplification,
\[
  \frac{1-\varepsilon}{1+\varepsilon}\,x^{-d}
  \;\le\;
  \frac{\mathbb{P}(S>tx)}{\mathbb{P}(S>t)}
  \;\le\;
  \frac{1+\varepsilon}{1-\varepsilon}\,x^{-d}.
\]
Letting $\varepsilon\downarrow 0$ shows
\[
  \lim_{t\to\infty}
  \frac{\mathbb{P}(S>tx)}{\mathbb{P}(S>t)}
  \;=\; x^{-d},
\]
i.e., the tail of $S$ is regularly varying with index $d$.

For the equivalent representation, define
\[
  L(s) \;:=\; s^{d}\,\mathbb{P}(S>s), \qquad s>1/\delta.
\]
From \eqref{eq:tail-bounds}, $L(s) \to c(\mathbf{z})$ as $s\to\infty$.
Hence $L$ is slowly varying (its ratio $L(tx)/L(t)\to 1$ for all $x>0$),
and
\[
  \mathbb{P}(S>s) \;=\; s^{-d}\,L(s), \qquad s\to\infty,
\]
with $L(s)\to c(\mathbf{z})=f(\mathbf{z})\,v_d$. This completes the proof.
\end{proof}
\section{Implementation Details}
\label{sec:details}
\subsection*{Construction of the MVTec-SynCA Dataset}

To create MVTec-SynCA, for each object subclass, we randomly selected a representative anomalous image and applied a series of geometric and photometric transformations to simulate real-world imaging variations. The applied transformations included:

\begin{itemize}
    \item Randomly applied rotations of $\pm 15^{\circ}$.
    \item Randomly applied translations of $\pm 2.5\%$ of image dimensions in both x and y directions.
    \item Randomly applied brightness, contrast and saturation adjustments of $\pm 10\%$ to 80\% of generated images.
    \item Applied Gaussian noise ($\sigma = 7.5$) and salt-and-pepper noise (1\% of total pixels) to 20\% of the generated images.
\end{itemize}

All transformations were applied to both the anomalous images and their corresponding ground truth masks using reflection padding to maintain boundary integrity. For each subclass, the number of synthetic anomalies added to each subclass is approximately 15\% of the total images. For the consistent-anomaly subclass \texttt{metal\_nut}, we kept the original dataset unchanged as the number of consistent-anomaly images already exceeds 20\% of the total images.

\subsection*{Construction of the ConsistAD Dataset}

The ConsistAD dataset was designed as a comprehensive benchmark for evaluating consistent anomaly detection by aggregating test images from three established anomaly detection datasets: MVTec AD, MVTec LOCO, and MANTA. It totally comprises 9 subclasses: \texttt{cable}, \texttt{metal\_nut}, and \texttt{pill} from MVTec AD; \texttt{breakfast\_box} and \texttt{pushpins} from MVTec LOCO; and \texttt{capsule}, \texttt{coated\_tablet}, \texttt{coffee\_bean}, and \texttt{red\_tablet} from MANTA. The dataset construction process for each source is detailed below:

\begin{itemize}
  \item \textbf{MVTec AD and MVTec LOCO}: We directly used the original test images from the specified subclasses without modifications.
  \item \textbf{MANTA}: The MANTA dataset consists of multi-view images, where each image concatenates five views of an object captured from different viewpoints, tailored for multi-view anomaly detection. To adapt these for our purpose, we extracted single-view images as follows:
  \begin{itemize}
    \item For normal images, we selected the middle view.
    \item For anomalous imagss, we chose the view with the largest anomaly mask area, as some views may not display anomalies.
    \item From each subclass, we randomly sampled 100 normal images and 50 anomalous images. To emphasize consistent anomalies, which often exhibit large anomaly masks in MANTA, we ensured that 40\% of the anomalous images (20 images) were those with the largest anomaly masks, while the remaining 40 were randomly selected from the other anomalous views.
  \end{itemize}
\end{itemize}
\bibliography{ref}
\bibliographystyle{plainnat}
\end{document}